\documentclass[lettersize,journal]{IEEEtran}
\usepackage{amsmath,amsfonts}
\usepackage{algorithmic}
\usepackage{algorithm}
\usepackage{array}
\usepackage{textcomp}
\usepackage{stfloats}
\usepackage{url}
\usepackage{verbatim}
\usepackage{graphicx}
\usepackage{cite}
\usepackage{blindtext}
\usepackage{hyperref}
\usepackage{multirow}
\usepackage{pifont}% http://ctan.org/pkg/pifont
\newcommand{\cmark}{\ding{51}}%
\newcommand{\xmark}{\ding{55}}%
\usepackage{footnote}
\usepackage{threeparttable}
\usepackage{booktabs}

\usepackage[font=small]{caption} % Standardize caption formatting
\usepackage{subcaption} % Add this line in the preamble if not already included
\usepackage{orcidlink}

\usepackage{etoolbox}
\usepackage{standalone}
\usepackage{tikz}
\usetikzlibrary{mindmap,backgrounds}
\usepackage[inkscapeexe=inkscape,inkscapeopt=-z]{svg}
\usepackage{booktabs}
\usepackage[commandnameprefix=ifneeded,todonotes={textsize=scriptsize}]{changes}
\usepackage{pgfplotstable}
\usetikzlibrary{arrows.meta,math}
\pgfplotsset{compat=1.18} 
\usepackage{siunitx}
\begin{document}

\title{Joint Perception and Prediction for \\Autonomous Driving: A Survey}

\author{Lucas Dal'Col\orcidlink{0000-0001-6121-6743}, Miguel Oliveira\orcidlink{0000-0002-9288-5058}, and Vítor Santos\orcidlink{0000-0003-1283-7388},~\IEEEmembership{Member,~IEEE}
        % <-this % stops a space
\thanks{Manuscript received... This work has been supported by FCT - Foundation for Science and Technology, in the context of Ph.D. scholarship 2023.02251.BD and under unit 00127-IEETA. \textit{(Corresponding author: Lucas Dal'Col)}.}% <-this % stops a space
\thanks{The authors are with the Department of Mechanical Engineering (DEM), the Intelligent System Associate Laboratory (LASI), and the Institute of Electronics and Informatics Engineering of Aveiro (IEETA), of the University of Aveiro (UA), 3810-193 Aveiro, Portugal (e-mail: lucasrdalcol@ua.pt; mriem@ua.pt; vitor@ua.pt).}}

% The paper headers
\markboth{\fontsize{7.5}{8}\selectfont \MakeLowercase{\MakeUppercase{T}his work has been submitted to the \MakeUppercase{IEEE} for possible publication. \MakeUppercase{C}opyright may be transferred without notice, after which this version may no longer be accessible.}}{}
% {Shell \MakeLowercase{\textit{et al.}}: A Sample Article Using IEEEtran.cls for IEEE Journals}

% \IEEEpubid{0000--0000/00\$00.00~\copyright~2021 IEEE}
% Remember, if you use this you must call \IEEEpubidadjcol in the second
% column for its text to clear the IEEEpubid mark.

\maketitle

\begin{abstract} % Maximum of 250 words

Perception and prediction modules are critical components of autonomous driving systems, enabling vehicles to navigate safely through complex environments. The perception module is responsible for perceiving the environment, including static and dynamic objects, while the prediction module is responsible for predicting the future behavior of these objects.
These modules are typically divided into three tasks: object detection, object tracking, and motion prediction. Traditionally, these tasks are developed and optimized independently, with outputs passed sequentially from one to the next. However, this approach has significant limitations: computational resources are not shared across tasks, the lack of joint optimization can amplify errors as they propagate throughout the pipeline, and uncertainty is rarely propagated between modules, resulting in significant information loss. 
To address these challenges, the joint perception and prediction paradigm has emerged, integrating perception and prediction into a unified model through multi-task learning. This strategy not only overcomes the limitations of previous methods, but also enables the three tasks to have direct access to raw sensor data, allowing richer and more nuanced environmental interpretations. 
This paper presents the first comprehensive survey of joint perception and prediction for autonomous driving. We propose a taxonomy that categorizes approaches based on input representation, scene context modeling, and output representation, highlighting their contributions and limitations. Additionally, we present a qualitative analysis and quantitative comparison of existing methods. Finally, we discuss future research directions based on identified gaps in the state-of-the-art.

\end{abstract}


%%%%%% SECTIONS
\section{Introduction} \label{sec:intro}

Autonomous driving (AD) is an exciting technology that holds the promise of revolutionizing transportation, providing a safe, comfortable and efficient driving experience \cite{Yurtsever2020}. AD systems are inherently complex, integrating diverse components such as sensors, processing units, and advanced algorithms. These systems must address a wide variety of challenges, including sensor inaccuracies, hardware reliability, real-time decision-making, adverse weather conditions, and dynamic traffic scenarios. The primary objective of an AD system is to process sensory inputs and generate vehicle control commands, such as steering angles and accelerator or brake inputs. 

AD systems are usually developed using one of two primary approaches: modular \cite{Yurtsever2020, Guo2020, Levinson2011, Chen2023, Wang2023} or end-to-end \cite{Kiran2022, Coelho2022, Chen2022RL, Coelho2024RLfOLD, Coelho2024RLAD}. The modular approach decomposes the overall AD system into sequential, easier-to-solve subproblems, while the end-to-end approach formulates the driving task as a single learning process, directly transforming sensor data into control commands. In the modular approach, the core components include perception, prediction, planning, and control. Among these, the perception and prediction modules play a critical role, as accurately perceiving the environment and predicting the future behavior of dynamic agents are essential for safe navigation through traffic. These modules must therefore be highly accurate, robust, and capable of operating in real time \cite{Feng2021, Ghorai2022, Karle2022}. Accurate perception and prediction significantly ease the tasks of downstream modules, such as motion planning \cite{Gonzalez2016} and control \cite{Paden2016}, while minimizing the risk of catastrophic failures due to error propagation \cite{Qian2022}.

The perception and prediction modules typically involve three key tasks: object detection, object tracking, and motion prediction \cite{Liang2020}. Object detection identifies objects of interest in the environment, such as vehicles, pedestrians, bicycles, and static obstacles. Accurate detection is crucial for understanding the surrounding scene and serves as the foundation for subsequent tasks. Building on detection, object tracking monitors these objects over time to establish their trajectories. It ensures temporal consistency, enabling the system to distinguish between stationary and moving objects. Finally, motion prediction is designed to predict the future movement of the tracked objects to anticipate potential collisions or conflicts. 

Traditionally, these tasks are designed and optimized independently and executed sequentially, with the output of one task feeding into the next. While this simplifies the design and implementation of individual tasks, it comes with notable shortcomings. First, computation is not shared across the tasks, resulting in inefficiencies and higher resource demands. For example, independently learning detection and tracking may result in redundant processing of scene features, thereby missing opportunities to optimize computational efficiency. Second, the lack of joint optimization means that errors in one module can propagate downstream. For instance, a false positive in object detection may lead to unnecessary tracking and motion predictions. Third, uncertainty is rarely propagated across modules, leading to information loss \cite{Zhang2020}.

Recently, joint perception and prediction approaches have emerged to address these challenges by integrating perception and prediction tasks into a unified learning-based framework. These methods leverage multi-task learning, allowing a single model to simultaneously handle the perception and prediction problems \cite{Luo2018, Liang2020, Zeng2019}. This integration offers several benefits, including shared computation across tasks, which significantly enhances efficiency. This a critical factor for real-time AD systems where high latency can be fatal. Furthermore, motion prediction and object tracking tasks can directly access raw sensor data instead of relying on the output of the object detection process, enabling more nuanced interpretations of the environment. This shared knowledge also strengthens the object detection task itself by accumulating contextual information over time. In this way, joint perception and prediction approaches reduce the detection of false negatives when dealing with occluded and far away objects, and the detection of false positives by accumulating evidence over time \cite{Luo2018}.

In light of the growing importance of this emerging field, this paper presents the \textbf{first comprehensive survey on joint perception and prediction for autonomous driving}. The contributions of this paper are as follows:

\begin{itemize}
    \item To present a survey of the state-of-the-art in joint perception and prediction for autonomous driving;
    \item To propose a taxonomy to classify the joint perception and prediction approaches;
    \item To provide a qualitative analysis and a quantitative comparison of existing methods;
    \item To identify research gaps and potential future research directions to advance the state-of-the-art.
\end{itemize}

The remainder of this paper is structured as follows: \autoref{sec:jpnp} introduces the taxonomy, categorizing approaches based on input representation, scene context modeling, and output representation. \autoref{sec:evaluation} provides a qualitative analysis and a quantitative comparison of existing methods. \autoref{sec:future_research_directions} discusses future research directions, highlighting research gaps to be addressed in joint perception and prediction. Finally, \autoref{sec:conclusion} summarizes the key findings and conclusions of this survey.

% \begin{figure*}[ht]
%     \centering
%     \input{figures/JPPmindmap}
%     \caption{Alternative diagram of taxonomy (Just an idea. May be discarded! :-) vsantos}
%     \label{fig:taxonomymindmap}
% \end{figure*}

\begin{figure*}[!ht]
    \centering
    \includegraphics[width=1\linewidth]{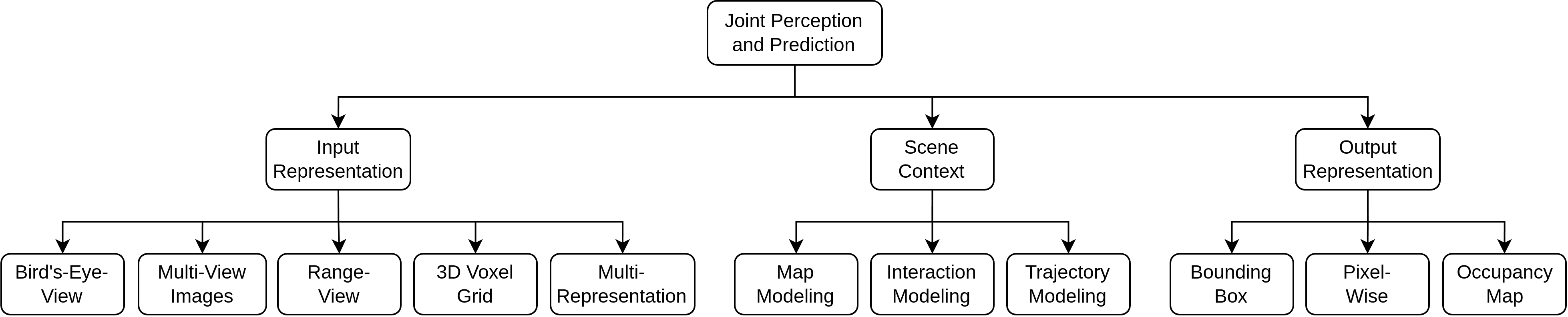}
    % \includesvg[width=1\textwidth,pretex=\sffamily\tiny\hypersetup{pdfborder={0 0 0}}]{figures/taxonomy_survey_paper.svg}
    \caption{The proposed taxonomy of joint perception and prediction for autonomous driving.}
    \label{fig:taxonomy_survey_paper}
\end{figure*}

\section{Joint Perception and Prediction} \label{sec:jpnp}

Joint perception and prediction methods are designed to simultaneously detect, track, and predict the motion of multiple agents within a scene. This integrated approach enables a deeper understanding of the environment by leveraging the synergies between perception and prediction tasks \cite{Luo2018}. These methods utilize diverse input and output representations, various sensor modalities, and different ways of modeling the scene context to optimize overall performance. 
% vsantos
\autoref{fig:taxonomy_survey_paper} summarizes the taxonomy and its various levels. %added a break here vsantos

In this section, based on a thorough analysis of the state of the art, we categorize the joint perception and prediction approaches into three key areas: input representation, scene context, and output representation. The \textbf{input representation} category is further divided into bird's-eye-view, multi-view images, range-view, 3D voxel grid, and multi-representation. The \textbf{scene context} is broken down into map modeling, interaction modeling, and trajectory modeling. Lastly, the \textbf{output representation} is classified into bounding box, pixel-wise, and occupancy map, highlighting different strategies to represent predicted outcomes. It is important to note that these taxonomy levels are not mutually exclusive; therefore, methods can simultaneously span multiple levels. This taxonomy highlights the range of strategies and methodologies employed in recent research to achieve joint perception and prediction in autonomous driving.

\subsection{Input Representation} \label{subsec:input_representation}
% Try to talk about the input representation and how they are modeled, how they encode the information.
% TODO: Say in this introduction that we will focus our discussions of the methods on how they create their input representation, and how they encode and process the information. DONE

\begin{figure*}[t]
    \centering
    \includegraphics[width=0.95\linewidth]{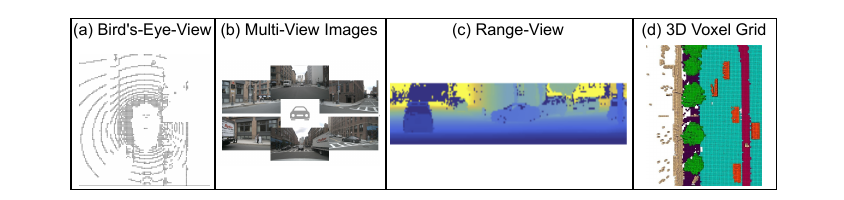}
    % \includesvg[width=0.95\linewidth]{figures/input_representation.svg}    
%% if you prefer sans-serif fonts inside the image, use the option pretex=\sffamily in includesvg  %vsantos
    \caption{Illustration of the \textbf{input representations} used in joint perception and prediction for autonomous driving: (a) \textbf{bird's-eye-view}, (b) \textbf{multi-view images}, (c) \textbf{range-view}, and (d) \textbf{3D voxel grid}. Multi-representation is not depicted in this figure, as it simply involves using two or more of these representations. Figure created based on \cite{Wu2020, Zhang2022, Meyer2021, Tong2023, Tian2023}.}
    \label{fig:input_representation_illustration}
\end{figure*}

The choice of input representation is crucial for joint perception and prediction approaches as it determines how information from the environment is captured, processed, and utilized. This information is gathered from a variety of sensors, such as cameras and LiDARs, and can be transformed into different representations to improve performance, reduce computational complexity, and enhance other aspects of the model. This section describes and compares the input representations commonly explored in recent research, including, \textbf{bird's-eye-view}, \textbf{multi-view images}, \textbf{range-view}, \textbf{3D voxel grid}, and \textbf{multi-representation}. \autoref{fig:input_representation_illustration} illustrates these representations. We discuss how these representations are constructed, as well as how different approaches propose to encode and process this data. Additionally, we also discuss how fusing these representations can improve the understanding of complex driving scenarios, contributing to more robust joint perception and prediction models.

\subsubsection{Bird's-Eye-View} \label{subsubsec:BEV}

% First talk about the BEV definition. 
% Talk about the advantages of BEV, and link these advantages to the fact the BEV representation is the most used representation in joint perception and prediction.
% Talk about why multi-view images do not enter in BEV as input representation.
% Describe the approaches. Decide whether we talk the sequence of papers within agent-level, occupancy-level and motion-level, or we talk about more generally.
% Try to focus on how they construct the BEV representation, however give the contributions of each paper, to strong the assumption of each category of the taxonomy is a independent section.
% Make a summary of the BEV representation and the general focus of the approaches. 
% Discuss open challenges and possible solutions for them.

% TODO: Write about how the BEV representation is created from the 3D point cloud data. DONE
% TODO: Put more references in the introduction of the section. DONE
% TODO: review this section to focus more on the encoding process of the approaches, regarding their input representation. DONE

Bird's-eye-view (BEV) representation is a type of voxelization that consists of a top-down view of the environment by transforming 3D point cloud data, typically captured by LiDAR sensors, into a 2D grid-based map. BEV provides strong prior information about object shapes and facilitates the fusion of data across multiple frames, making it ideal for joint perception and prediction methods \cite{Khalil2021}. For this reason, BEV is the most commonly used input representation for these methods.

FaF \cite{Luo2018} was a pioneering work that introduced the concept of joint perception and prediction. FaF used a BEV grid, representing the 3D environment in a 4D tensor (x, y, height, time). Two fusion strategies were proposed: early fusion, which aggregates temporal information at the input level using 1D convolutions before extracting features with 2D convolutions, and late fusion, which incrementally combines temporal data with 2D and 3D convolutions. FaF established the groundwork for joint perception and prediction within a single end-to-end neural network, allowing the propagation of uncertainty and improving holistic reasoning. IntentNet \cite{Casas2018} furthered this approach by stacking height and time dimensions into the channel dimension, optimizing computational efficiency with 2D convolutions. Subsequent works \cite{Zeng2019, Liang2020, Sadat2020, Casas2021, Djuric2021, Luo2021, Khurana2022} followed IntentNet and encoded the BEV representation in a similar way. Several approaches subsequently focused on enhancing the extraction of spatial and temporal features. The authors of \cite{Casas2020SpAGNN, Casas2020ILVM, Casas2020, Phillips2021, Cui2021} applied the Rotated Region of Interest Align (RRoI align) \cite{Ma2018} to extract per-actor features. MotionNet \cite{Wu2020} explored when and how to aggregate temporal features, in order to better capture both local and global contexts. To achieve that, they proposed a Spatio-Temporal Pyramid Network (STPN) that relies solely on 2D and pseudo-1D convolutions. MotionNet set an efficient baseline, later adopted by works such as \cite{Wang2022, Li2023, Wang2024semi, Wang2024self} for feature extraction. LidNet \cite{Khalil2022} built upon MotionNet with enhancements such as residual convolutional blocks and replacing strided convolutions with average pooling for spatial reduction. SDP-Net \cite{Zhang2021} introduced a BEV flow map that dynamically estimates motion and aligns features across frames, allowing more effective motion estimation and feature aggregation. SDAPNet \cite{Ye2021} fused multi-scale feature maps from a 2D Convolutional Neural Network (CNN) using a Multi-to-Single Fusion (MoSF) mechanism. ImplicitO \cite{Agro2023} used a 2D CNN feature extractor combined with a Feature Pyramid Network (FPN) \cite{Lin2017} to process multi-resolution feature planes. 

Departing from traditional occupancy BEV grids, FS-GRU \cite{Chen2022}, ContrastMotion \cite{Jia2023} and STINet \cite{Zhang2020} used PointPillars \cite{Lang2019} to encode features within vertical pillars of the point cloud, forming 2D BEV pseudo-images further processed by 2D CNNs. To capture temporal dynamics, FS-GRU used Convolutional Gated Recurrent Unit (ConvGRU) for shared feature extraction between frames, ContrastMotion proposed Gated Multi-Frame Fusion (GMF) for complementary features from adjacent frames, and STINet developed a Temporal Region Proposal Network (T-RPN) to generate future object proposals using current and past bounding boxes. Meanwhile, FutureDet \cite{Peri2022} and DeTra \cite{Casas2024} employed VoxelNet \cite{Zhou2018} instead of PointPillars to extract voxel features from point cloud sweeps, which are also further processed by 2D CNNs. Additionally, DeTra also integrated a multiscale deformable attention to fuse multi-level feature maps. To model long-range spatial and temporal interactions, STAN \cite{Wei2022} introduced a spatiotemporal transformer network with dedicated temporal and spatial attention modules, diverging from earlier approaches that predominantly used CNNs or RNNs.

The evolution of BEV-based joint perception and prediction approaches has brought significant advancements, but limitations remain. The voxelization process inherent in BEV representations can lead to the loss of fine-grained details from the original 3D point cloud, which may degrade the accuracy of perception and prediction tasks. Additionally, BEV grids can become computationally expensive as their resolution increases, posing challenges for real-time applications.

\subsubsection{Multi-View Images} \label{subsubsec:multi-view_images}

% TODO: Put more references in the introduction of the section. DONE
% TODO: review this section to focus more on the encoding process of the approaches, regarding their input representation.

Multi-view images are captured by multiple cameras positioned around a vehicle, providing comprehensive 360-degree coverage of its surroundings. Processing consecutive frames from these cameras enables joint perception and prediction of the behavior of multiple agents in complex driving scenarios. Approaches using this input representation have evolved rapidly, with most studies lifting camera features into a BEV representation. It is noteworthy that BEV formation from multi-view images is part of the network’s learning process and not the input itself, while in the methods described in \autoref{subsubsec:BEV} BEV is the primary input to the neural network. These camera-based methods have shown the potential to rival the performance of LiDAR-based approaches, offering benefits such as lower cost and higher resolution \cite{Hu2021}.

FIERY \cite{Hu2021} was the first approach to achieve joint perception and prediction from multi-view images. It used a convolutional encoder to extract features from each camera and predicted discrete depth probabilities. These depth estimates, combined with camera intrinsics and extrinsics, allowed the model to lift the 2D images into 3D space. The 3D features were then pooled along the vertical axis to create the BEV feature map for each time frame. To align these features over time, FIERY employed ego-motion data and a Spatial Transformer \cite{Jaderberg2015}, followed by a 3D convolutional network to capture spatio-temporal dynamics. Building upon the foundation established by FIERY, subsequent works focused on enhancing the BEV transformation process and improving modeling efficiency. BEVerse \cite{Zhang2022} introduced a SwinTransformer backbone for more effective feature extraction from 2D images, while PowerBEV \cite{Li2023PowerBEV} used 2D convolutions, collapsing time and feature dimensions, to improve computational efficiency. ST-P3 \cite{Hu2022ST-P3} proposed an egocentric aligned accumulation strategy, ensuring better spatial alignment across frames. StretchBEV \cite{Akan2022} incorporated temporal dynamics using a recurrent neural network (RNN) with stochastic residual updates, enabling diverse long-term predictions. More recent approaches, PIP \cite{Jiang2022} and UniAD \cite{Hu2023} simultaneously learned static map features and dynamic agent motion features through interactions between queries and the environment. TBP-Former \cite{Fang2023} unified the BEV construction by transforming image features and synchronizing multiple time frames in a single step using a cross-view attention mechanism. Additionally, it introduced a pyramid transformer to better capture spatial-temporal features, outperforming traditional RNNs or 3D convolutions. 

Shifting away from explicit BEV grids, ViP3D \cite{Gu2023} used agent-centric 3D queries to aggregate spatial features and track agents dynamically over time, implicitly achieving a top-down spatial context. VAD \cite{Jiang2023} argued that autonomous driving could be achieved using a fully vectorized representation instead of dense BEV grids, achieving high computational efficiency. Through vectorized map and motion representations, VAD demonstrated that agent and map queries could effectively learn and represent the scene. 

In summary, the evolution of multi-view image approaches has focused on enhancing the multi-view images transformation to a top-down view representation, temporal modeling, and computational efficiency. Alternative methods, such as the vectorized representation in VAD and the implicit top-down spatial context in ViP3D, offer promising directions for future research. However, approaches that rely solely on multi-view images face notable limitations. Depth estimation from monocular images across multiple cameras can be inaccurate, as cameras do not directly capture depth information. This process is highly dependent on accurate camera calibration to ensure proper alignment in 3D space. Aligning consecutive frames can be challenging due to motion blur, mismatched frame rates, or fast-changing scenes. Finally, camera-based methods struggle in poor visibility conditions such as fog, rain or low light, where the quality of visual input is significantly reduced.

% Balancing accuracy and computational efficiency in these frameworks will be crucial.

\subsubsection{Range-View} \label{subsubsec:RV}

Range-view (RV) representation is a native format for LiDAR data, where the 3D point measurements from a LiDAR sweep are projected onto a 2D panoramic range image. Each sweep captures measurements from a full \ang{360} rotation, resulting in a dense representation of the environment. In an RV image, each pixel corresponds to a LiDAR point, with its position determined by the azimuth and elevation angles of the sensor. When multiple points project to the same pixel, the point with the smallest range is retained. Compared to BEV representation, which renders the 3D point cloud into a 2D grid, RV maintains the original maximum range and resolution of the sensor data. This allows fine-grained details to be captured, such as identifying which parts of the scene are visible to the sensor and which parts are occluded. Furthermore, RV retains the native structure of the data without the information loss associated with voxelization, enabling the detection of smaller and distant objects more effectively than BEV methods \cite{Meyer2021}.

Several approaches have utilized RV representation for joint perception and prediction, proposing innovative methods to fuse multiple sweeps from consecutive time frames and extract meaningful features. LaserFlow \cite{Meyer2021} was the first to use an RV-only representation, introducing a multi-sweep fusion architecture to address information loss due to viewpoint changes. It accomplishes this by extracting features independently from each sweep in its original view using 2D convolutions and then transforming these features to a common viewpoint through ego-motion compensation. Building on this concept, RV-FuseNet \cite{Laddha2021} proposed an incremental fusion approach, which sequentially fuses sweeps to minimize information loss, particularly in scenarios with significant ego-motion or object movement. Meanwhile, SPFNet \cite{Weng2020} used RV to forecast future point cloud sweeps, thereby avoiding the need for object-level labels. They employed a shared 2D CNN encoder to extract features from each RV image, followed by a Long Short-Term Memory (LSTM) network to capture temporal dynamics, treating it as a sequence-to-sequence problem.

In summary, RV representation enables a detailed and efficient way to utilize LiDAR data, supporting end-to-end joint perception and prediction approaches. Its ability to maintain the native structure of sensor data makes it a powerful alternative to BEV. However, RV still faces significant challenges, particularly with aligning features across multiple sweeps due to changes in perspective. Fusing consecutive time frames of this representation makes distortions arise due to the shift in the center of spherical projections \cite{Meyer2021, Fadadu2022}. These challenges limit the exploration of RV compared to BEV, despite its advantages.
% Addressing these issues requires robust fusion strategies, such as deep learning-based feature alignment and hybrid approaches that combine RV and BEV representations to leverage the strengths of both.

\begin{table*}[t]  % Use table* to span both columns
\renewcommand{\arraystretch}{1.4}
\centering
\caption{Summary of the joint perception and prediction approaches according to the \textbf{input representation} level of the taxonomy. Works are sorted in ascending chronological order.}
\label{tab:summary_input_representation}
\resizebox{\textwidth}{!}{%
\begin{tabular}{m{2.6cm}m{4.5cm}m{8.5cm}}  % Adjust widths to fit content
% \toprule
\textbf{Input Representation} & \textbf{Characteristics} & \textbf{Works} \\ \toprule
Bird's-Eye-View               & Top-down 2D grid map from 3D LiDAR point cloud & FaF \cite{Luo2018}, IntentNet \cite{Casas2018}, NMP \cite{Zeng2019}, SpAGNN \cite{Casas2020SpAGNN}, PnPNet \cite{Liang2020}, MotionNet \cite{Wu2020}, STINet \cite{Zhang2020}, ILVM \cite{Casas2020ILVM}, PPP \cite{Sadat2020}, Casas et al. \cite{Casas2020}, SDP-Net \cite{Zhang2021}, Phillips et al. \cite{Phillips2021}, MP3 \cite{Casas2021}, MultiXNet \cite{Djuric2021}, SDAPNet \cite{Ye2021}, SA-GNN \cite{Luo2021}, LookOut \cite{Cui2021}, STAN \cite{Wei2022}, BE-STI \cite{Wang2022}, FutureDet \cite{Peri2022}, FS-GRU \cite{Chen2022}, Khurana et al. \cite{Khurana2022}, LidNet \cite{Khalil2022}, WeakMotionNet \cite{Li2023}, ImplicitO \cite{Agro2023}, ContrastMotion \cite{Jia2023}, MSRM \cite{Wang2024semi}, Wang et al. \cite{Wang2024self}, and DeTra \cite{Casas2024} \\ \midrule
Multi-View Images             & 360-degree coverage from multiple vehicle-mounted RGB cameras & FIERY \cite{Hu2021}, BEVerse \cite{Zhang2022}, ST-P3 \cite{Hu2022}, StretchBEV \cite{Akan2022}, PIP \cite{Jiang2022}, TBP-Former \cite{Fang2023}, UniAD \cite{Hu2023}, ViP3D \cite{Gu2023}, PowerBEV \cite{Li2023PowerBEV}, and VAD \cite{Jiang2023} \\ \midrule
Range-View                    & Panoramic range image from LiDAR measurements & SPF2 \cite{Weng2020}, LaserFlow \cite{Meyer2021}, and RV-FuseNet \cite{Laddha2021} \\ \midrule
3D Voxel Grid                 & 3D LiDAR point cloud divided into uniform voxel grids & Proxy-4DOF \cite{Khurana2023}, and Occ4cast \cite{Liu2023} \\ \midrule
% HD Map                        & High-resolution maps with detailed road information & IntentNet \cite{Casas2018}, NMP \cite{Zeng2019}, PnPNet \cite{Liang2020}, ILVM \cite{Casas2020ILVM}, PPP \cite{Sadat2020}, SpAGNN \cite{Casas2020SpAGNN}, InteractionTransformer \cite{Li2020}, Casas et al. \cite{Casas2020}, LiRaNet \cite{Shah2020}, Phillips et al. \cite{Phillips2021}, LookOut \cite{Cui2021}, MultiXNet \cite{Djuric2021}, SA-GNN \cite{Luo2021}, MVFuseNet \cite{Laddha2021MVFuseNet}, Fadadu et al. \cite{Fadadu2022}, FS-GRU \cite{Chen2022}, ViP3D \cite{Gu2023}, ImplicitO \cite{Agro2023}, DeTra \cite{Casas2024} \\ \midrule
Multi-Representation   & Combined data from various sensor modalities and representations & FISHING Net \cite{Hendy2020}, InteractionTransformer \cite{Li2020}, LiRaNet \cite{Shah2020}, SSPML \cite{Luo2021Pillar}, MVFuseNet \cite{Laddha2021MVFuseNet}, Khalil et al. \cite{Khalil2021}, LiCaNext \cite{Khalil2021LiCaNext}, Fadadu et al. \cite{Fadadu2022}, LiCaNet \cite{Khalil2022LiCaNet}, FusionAD \cite{Ye2023}, and Fang et al. \cite{Fang2024} \\ \bottomrule
\end{tabular}%
}
\end{table*}

\subsubsection{3D Voxel Grid} \label{subsubsec:3D_voxel_grid}

3D voxel grids are a volumetric representation of the environment, typically derived from LiDAR point cloud data, where the space is divided into a uniform grid of small cubic cells, known as voxels. Each voxel stores information about whether the space is occupied or not. Compared to RV and BEV representation, 3D voxel grid provides a more comprehensive understanding of the 3D scene. While RV suffers from distortions due to spherical projections, and BEV voxelizes the 3D scene to a top-down 2D plane, losing important vertical information, 3D voxel grids better preserve the 3D geometry.

% Methods leveraging 3D voxel grid representations have primarily focused on scene completion and occupancy forecasting. 
Khurana et al. \cite{Khurana2023} used consecutive LiDAR point cloud sweeps to create a voxel grid with spatial and temporal dimensions. By merging the vertical and temporal dimensions into a single channel, they were able to apply 2D convolutions to the data while still capturing the 4D spatial-temporal occupancy. Occ4cast \cite{Liu2023} introduced the Occupancy Completion and Forecasting (OCF) task, combining scene completion and forecasting within a single framework. To demonstrate the feasibility of their approach, they explored different baseline architectures, such as 3D convolutions and Convolutional Long Short-Term Memory (ConvLSTM) \cite{Shi2015}, for modeling spatial-temporal correlations.
% \todo[inline]{As far as I understood, 3D point clouds are never used as a direct input representation, but are first converted into Range Images or 3D Voxel Grids, is that correct?}

Despite its potential, 3D voxel grid representation is not widely adopted in joint perception and prediction, with only a few approaches exploring it. The main challenge lies in the high computational cost of processing large voxel grids with temporal dimensions, which becomes even more significant when using high-resolution grids. Since joint perception and prediction approaches often require multiple consecutive frames of sensor data to forecast the trajectories of dynamic objects, they typically rely on simplified representations such as BEV and RV to reduce computational overhead. For the same reason, in joint perception and prediction approaches, raw point clouds are never used directly without a pre-processing step to convert them into BEV, RV, or 3D voxel grids.
\subsubsection{Multi-Representation} \label{subsubsec:multi-representation}

Multi-representation fusion in autonomous driving integrates data from various sensor modalities and representations, such as BEV point cloud, RV point cloud, and camera images. By combining different formats, these approaches exploit the complementary strengths of each representation while compensating for their individual limitations. For instance, BEV preserves the physical size of objects and spatial relationships, RV retains detailed occlusion information and RGB images provide dense semantic features \cite{Khalil2022}. This fusion enables a more robust and comprehensive understanding of the driving scene.

The approaches discussed in this section employ multi-representation fusion in various ways, primarily differing in the types of representations they use and how they integrate the data. Some methods \cite{Li2020, Ye2023, Luo2021Pillar, Fang2024} fuse camera images with BEV representations. For example, the authors in \cite{Li2020} employed a dual-stream architecture where a 2D CNN processes multi-sweep BEV LiDAR, while a pretrained ResNet-18 \cite{He2016} extracts features from front-view images. The features are then fused through a continuous fusion layer in the BEV space, resulting in a dense and unified representation. Similarly, FusionAD \cite{Ye2023} processes multi-view camera images and BEV LiDAR data separately with backbone networks, then combines them using a transformer-based architecture with multiple attention mechanisms: points cross-attention for the LiDAR BEV features, image cross-attention for image features, and temporal self-attention for historical BEV features. The authors of \cite{Luo2021Pillar} and \cite{Fang2024} incorporate optical flow from camera images to complement LiDAR BEV representation, thereby enhancing motion prediction capabilities. 

Other approaches (\cite{Khalil2021, Laddha2021MVFuseNet}) fuse LiDAR point clouds using both BEV and RV representations. For instance, the authors of \cite{Khalil2021} employed separate CNN branches for historical BEV point cloud sweeps and the current RV sweep. The RV features are then processed by a U-Net \cite{Ronneberger2015} and projected onto the BEV space for further fusion with the BEV features. MVFuseNet \cite{Laddha2021MVFuseNet} goes a step further by performing multi-view temporal fusion over multiple LiDAR sweeps, sequentially processing RV features and projecting them to BEV from the oldest to the most recent sweep. LiCaNet \cite{Khalil2022LiCaNet} and its successor, LiCaNext \cite{Khalil2021LiCaNext}, enhance this by incorporating camera images into the fusion process, with LiCaNext adding residual images to capture temporal dynamics. The authors of \cite{Fadadu2022} utilized cameras, LiDAR BEV, and RV in a manner similar to LiCaNet and LiCaNext.

Radar data has also been explored in multi-representation fusion, although it has not been investigated as a standalone input for joint perception and prediction approaches as the representations discussed in the previous sections. LiRaNet \cite{Shah2020} integrates radar and LiDAR data by applying spatial-temporal processing to radar features with graph-based convolutions and multi-layer perceptrons (MLPs) for temporal fusion. These features are then fused into a shared BEV representation for further processing. FISHING Net \cite{Hendy2020} fuses consecutive frames of multi-view camera images, radar BEV, and LiDAR BEV representations using separate convolutional encoder-decoder networks, aggregating the outputs in the BEV space via average or priority pooling methods. 

Multi-representation fusion leverages the complementary strengths of various representations, significantly enhancing system capabilities, but still presents challenges. The increased computational complexity of processing multiple representations can lead to higher latency, posing a concern for real-time applications. Additionally, aligning features from different modalities, such as cameras, LiDAR, and radar, is difficult due to variations in resolution, noise characteristics, and field of view. Radar, in particular, is less widely used because it offers 2-3 orders of magnitude fewer points and has a much lower angular resolution compared to LiDAR \cite{Shah2020}. 
% Addressing these challenges requires the development of more efficient fusion architectures. Promising solutions include the use of adaptive attention mechanisms, such as deformable attention, which can dynamically focus on relevant regions and integrate features across modalities. 

\subsubsection*{Summary} \label{subsubsec:summary_input_representation}

\autoref{tab:summary_input_representation} summarizes the joint perception and prediction approaches categorized by their input representation level in the taxonomy. BEV representations are by far the most commonly used, appearing in 29 works from a set of 55. Additionally, there has been considerable research into multi-view images and multi-representation approaches.

\begin{figure*}[t]
    \centering
    \includegraphics[width=1\textwidth]{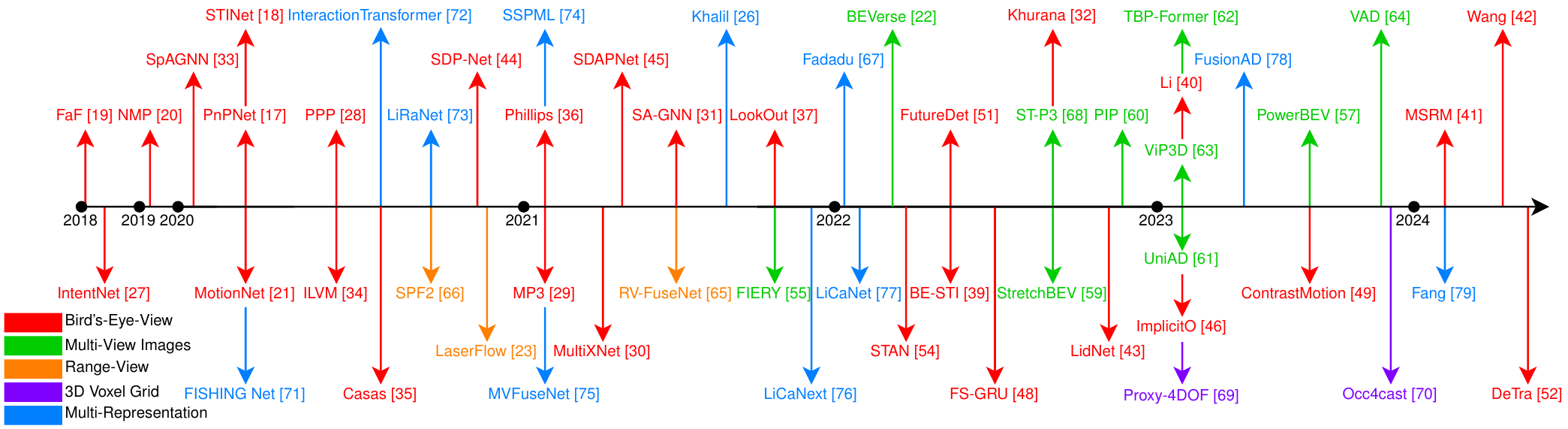}
    %the \hypersetup{pdfborder={0 0 0}} modifier included to avoid showing locally the hyperlink boxes. %%vsantos
% \includesvg[width=1\textwidth,pretex=\sffamily\tiny\hypersetup{pdfborder={0 0 0}}]{figures/temporal_summary_inputrepresentation_refs.svg}   
    
% \input{figures/gentimelineB} 
\caption{Chronological overview of the joint perception and prediction approaches according to the \textbf{input representation} level of the taxonomy.}
% Below an attempt to create automatically this timeline after a data file... it's a sample only (wrong categories), but it's highly configurable (just edit the data file): check if it is useful and viable.}
\label{fig:temporal_summary_input_representation}
\end{figure*}

\autoref{fig:temporal_summary_input_representation} provides a chronological overview of these approaches based on their input representations. Notably, BEV representation was the pioneering choice for joint perception and prediction and continues to be widely adopted. However, multi-view images have gained significant attention in recent years, highlighting the growing importance of dense and semantic representations. Multi-representation approaches saw a surge in interest during 2020 and 2021, and they remain a viable option due to the benefits of leveraging complementary sensors and representations. Finally, 3D voxel grid approaches are still emerging, driven by the appearance of self-supervised point cloud forecasting methods for joint perception and prediction.

\subsection{Scene Context} \label{subsec:scene_context_modeling}
% Try to talk about implicitly and explicitly scene context modeling. 

\begin{figure*}[t]
    \centering
   \includegraphics[width=0.95\linewidth]{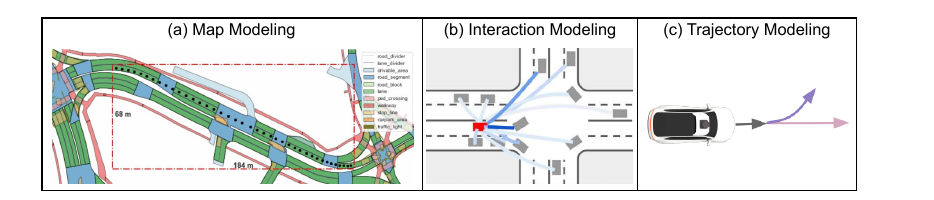}
    % \includesvg[width=0.95\linewidth]{figures/scene_context.svg}
    
    \caption{Illustration of the types of \textbf{scene context} modeling: (a) \textbf{map modeling}, (b) \textbf{interaction modeling}, and (c) \textbf{trajectory modeling}. Figure created based on \cite{Caesar2020, Li2020, Cui2021}.}
    \label{fig:scene_context_illustration}
\end{figure*}

Scene context modeling enables autonomous driving systems to understand both the static and dynamic elements of the environment, facilitating accurate detection of agents and prediction of their future movements. By modeling scene context, these systems can better interpret constraints and uncertainties inherent to perception and prediction tasks, such as road topology, traffic rules, the stochastic nature of agent motion, and the interactions between agents. This section is divided into three key areas: \textbf{map modeling}, \textbf{interaction modeling}, and \textbf{trajectory modeling}, each addressing different facets of the scene context. \autoref{fig:scene_context_illustration} illustrates these types of scene context modeling.

\subsubsection{Map Modeling} \label{subsubsec:map_modeling}
% Talk about the map modeling in general way, whether it is implicitly or explicitly. Talk initially about HD maps, and how they model and rely on HD maps for understand the context of the scene, and also talk about the approaches that output a semantic map from sensory inputs.
% Try to search for some other dimension about map modeling.

Map modeling provides autonomous driving systems with the context for understanding traffic constraints, including drivable surfaces, lane boundaries and directions, intersections, and pedestrian crossings. These elements deliver contextual insight into the possible actions and constraints for each agent within the scene.

Approaches to map modeling can be categorized based on whether they implicitly or explicitly incorporate the static map context. Most approaches \cite{Luo2018, Wu2020, Zhang2020, Hendy2020, Weng2020, Zhang2021, Meyer2021, Luo2021Pillar, Ye2021, Laddha2021, Khalil2021, Hu2021, Khalil2021LiCaNext, Khalil2022LiCaNet, Wei2022, Wang2022, Peri2022, Akan2022, Khurana2022, Khalil2022, Li2023, Khurana2023, Li2023PowerBEV, Jia2023, Liu2023, Wang2024semi, Fang2024, Wang2024self} rely on implicit modeling, where the spatial structure, dynamics, and environmental constraints are indirectly learned through the sensor inputs and the training process of the model, rather than being explicitly represented within the network. On the other hand, other approaches \cite{Zeng2019, Liang2020, Li2020, Casas2018, Casas2020ILVM, Sadat2020, Casas2020SpAGNN, Casas2020, Shah2020, Phillips2021, Cui2021, Djuric2021, Luo2021, Fadadu2022, Chen2022, Agro2023, Laddha2021MVFuseNet, Gu2023, Casas2024, Casas2021, Zhang2022, Hu2022, Fang2023, Jiang2022, Jiang2023, Hu2023, Ye2023} explicitly represent map information, either by using High-Definition Maps (HD Maps) as inputs or by constructing semantic maps from sensor data in real-time. 

HD maps are leveraged as an input source of detailed prior knowledge, allowing models to integrate static information on traffic rules and constraints directly. HD maps provide detailed and structured a priori information about the driving environment, including road geometry, road lanes, drivable surfaces, traffic signs, intersections, pedestrian crossings, and other key static elements. However, they cannot be used as the sole input for autonomous driving tasks due to their lack of real-time information, being combined with sensor data. Approaches using HD maps can be subdivided based on the map encoding strategies and the stage at which the fusion takes place. Rasterized BEV representation is the most common method, where HD maps are converted into multi-channel BEV binary masks, with each channel containing different semantic features like lanes, road surfaces, intersections, among others. Some methods (\cite{Zeng2019,Liang2020,Li2020}) perform fusion at the raw input level, merging rasterized BEV HD maps with BEV LiDAR data through straightforward concatenation along the channel dimension, since both share a top-down view of the environment. These inputs are then processed together by convolutional networks to extract joint features. However, the majority of approaches \cite{Casas2018, Casas2020ILVM, Sadat2020, Casas2020SpAGNN, Casas2020, Shah2020, Phillips2021, Cui2021, Djuric2021, Luo2021, Fadadu2022, Chen2022, Agro2023, Laddha2021MVFuseNet} perform fusion at the feature level, where HD maps and sensor data are processed separately through backbone networks, typically CNNs, before the extracted feature maps are concatenated and passed through another network to obtain the fused features. Processing HD maps and sensor data separately allows backbone networks to independently extract meaningful features from each source, which are then fused to ensure balanced integration between their unique features without one overshadowing the other. Some approaches have moved beyond rasterized representations; for instance, ViP3D \cite{Gu2023} employs a vectorized representation of HD maps, using Graph Neural Networks (GNNs) to extract features from the vectorized map elements. Similarly, DeTra \cite{Casas2024} uses a graph-based representation, where lane centerlines are divided into segments and each center of the segment forms a node in the lane graph. Graph Convolution Networks (GCN) are then used to extract map embeddings for each lane node.

Other approaches opt to construct their semantic maps dynamically, directly from sensor data. These online semantic maps are built in real-time, predicting updated and adaptive features such as drivable areas, intersections, construction zones, and reachable lanes, making them an alternative to static HD maps. For example, MP3 \cite{Casas2021} constructs an online map from BEV point cloud feature maps using a CNN decoder, outputting probabilistic estimates organized in channels to handle uncertainties. The drivable area and intersections channels are modeled as Bernoulli random variables, and reachable lanes channel as a Laplacian distribution. BEVerse \cite{Zhang2022} and ST-P3 \cite{Hu2022} similarly use CNN decoders to create online semantic maps, operating on multi-view images encoded by ConvGRUs. TBP-Former \cite{Fang2023} adopts a transformer-based framework to encode multi-view images, using a decoder to predict essential traffic elements like drivable areas and lanes. Some methods, such as PIP \cite{Jiang2022} and VAD \cite{Jiang2023}, use map queries to extract BEV map features, producing map vectors that represent road topology instead of multi-channel map layers, while UniAD \cite{Hu2023} and FusionAD \cite{Ye2023} employ map queries to perform panoptic segmentation of the map.

In joint perception and prediction, map modeling ranges from implicit modeling to full HD map dependency or on-the-fly semantic map construction. Systems that generate online semantic maps can operate with less dependency on precise localization than those relying on HD maps, which require centimeter-level accuracy \cite{Casas2021}. These online maps update dynamically based on sensor inputs, making them highly responsive in situations like construction zones or temporary road alterations. Although HD maps offer superior detail, they are resource-intensive, costly, and difficult to maintain, particularly over large geographic areas, often leading to outdated information. Online semantic maps, despite potential uncertainties, address these challenges by providing adaptive and up-to-date environmental context.

\subsubsection{Interaction Modeling} \label{subsubsec:interaction_modeling}
% Talk about 2 main concepts, agent-agent interaction and scene-agent interaction. Within agent-agent interaction, there are two more concepts: intra-class agent-agent interactions and inter-class agent-agent interactions. Try to structure the sections using these concepts. Also, put the remaining approaches as having implicit interactions, which can be learned by the neural network.
% Interaction modeling is crucial to achieve accurate and reliable motion prediction of dynamic agents in complex scenarios. 
Vehicles on the road are constantly interacting with other road users, including other vehicles, pedestrians, and bicycles, as well as with the spatial layout and traffic rules of the environment. These interactions influence the intentions and future movements of each agent in the scene. For example, as a vehicle approaches a roundabout, it yields to cars already circulating within, adjusting its entry speed based on the flow of traffic while adhering to rules that give priority to those within the roundabout. Modeling these interactions is critical for capturing the nuances of regulatory and social behaviors on the road, enabling the prediction of accurate and collision-free future trajectories in complex scenarios.

Interaction modeling can be categorized based on whether these social behaviors are incorporated implicitly or explicitly into the system. Implicit interaction modeling approaches \cite{Luo2018, Casas2018, Zeng2019, Liang2020, Wu2020, Hendy2020, Sadat2020, Weng2020, Shah2020, Zhang2021, Meyer2021, Casas2021, Luo2021Pillar, Laddha2021MVFuseNet, Djuric2021, Ye2021, Laddha2021, Khalil2021, Hu2021, Khalil2021LiCaNext, Fadadu2022, Khalil2022LiCaNet, Zhang2022, Wei2022, Wang2022, Peri2022, Chen2022, Hu2022ST-P3, Akan2022, Khurana2022, Khalil2022, Fang2023, Gu2023, Li2023, Agro2023, Khurana2023, Li2023PowerBEV, Jia2023, Liu2023, Wang2024semi, Fang2024, Wang2024self} do not employ a dedicated model to capture interactions between agents or between agents and the scene, whereas explicit interaction modeling \cite{Casas2020SpAGNN, Casas2020ILVM, Casas2020, Phillips2021, Zhang2020, Luo2021, Cui2021, Li2020, Jiang2022, Jiang2023, Hu2023, Ye2023, Casas2024} applies a specific model designed for this purpose. This omission increases the possibility of predicting overlapping or unrealistic future trajectories. Explicit interaction modeling can be further divided into agent-agent and agent-scene interactions. Agent-agent interactions capture the mutual influence of agents on the future trajectories of each other, as seen in behaviors such as overtaking or yielding to other vehicles. Agent-scene interactions, on the other hand, refer to the influence of static elements in the scene, such as lane boundaries, traffic signals, and road geometry, on the predicted movements of agents. It is important to note that map modeling is often confused with agent-scene interaction modeling. Map modeling provides detailed, static information about the road layout, such as lane boundaries and traffic signs, whereas agent-scene interaction modeling focuses on how these static elements dynamically influence the movement and behavior of agents in the environment. Furthermore, agent-agent interactions can be distinguished between intra-class and inter-class interactions. Intra-class interactions occur among agents of the same class, such as vehicle-vehicle interactions where one car yields to another. In contrast, inter-class interactions occur between agents of different classes; for example, the interaction between a vehicle and a pedestrian crossing the street.

SpAGNN \cite{Casas2020SpAGNN} was the first joint perception and prediction approach to incorporate interaction modeling. It encodes agent-agent interactions between vehicles through a fully connected directed graph, where each agent is represented as a node, with bidirectional connections indicating interactions between them. Bidirectionality is important because the relationships between actors can be asymmetric; for example, a vehicle following another vehicle in the front. The Spatially-Aware Graph Neural Network (SpAGNN), inspired by the Gaussian Markov Random Field (Gaussian MRF), performs message passing between nodes to update its state based on neighboring nodes, capturing interaction dynamics. For each agent in the scene, the positions of all other agents are transformed to the local coordinate system of that agent, making the graph aware of the spatial relationships between actors. Two other approaches \cite{Casas2020ILVM, Casas2020} built on SpAGNN to model interactions between vehicles but improved upon a few aspects. ILVM \cite{Casas2020ILVM} represented the scene through a distributed latent space shared among actors, efficiently sampling multiple scene-consistent trajectories for all agents. Meanwhile, the authors of \cite{Casas2020} additionally introduced agent-scene interactions by incorporating prior knowledge of road topology, traffic rules, and the structure of the driving scene as a loss function to enforce rule-following behaviors among traffic participants. The authors of \cite{Phillips2021} used the same interaction modeling as in ILVM. STINet \cite{Zhang2020} and SA-GNN \cite{Luo2021} also used GNNs, similar to previous approaches, for modeling agent-agent interactions, but specifically focusing on pedestrians as nodes within the interaction graph. Pedestrians are important agents in the scene but exhibit unique behaviors, which pose additional challenges. For example, a pedestrian walking within a group may tend to follow the group closely. LookOut \cite{Cui2021} goes further by using a GNN to perform interaction modeling between multi-class agents, e.g., vehicles, pedestrians and bicycles, to achieve both inter-class and intra-class interactions. The authors of \cite{Li2020} proposed an interaction transformer where the queries are the set of detected vehicles along with their feature vectors and spatial information, the keys and values represent neighboring contextual information from other agents, and the positional embeddings encode the relative positions and orientations between agents. PIP \cite{Jiang2022} and VAD \cite{Jiang2023} advanced this further by modeling agent-agent interactions with a self-attention mechanism in the motion queries and agent-scene interactions with cross-attention modules between motion queries and map queries. UniAD \cite{Hu2023} and FusionAD \cite{Ye2023} perform agent-agent and agent-scene interactions similarly to PIP but expand to detect both vehicles and pedestrians. They also introduced an agent-goal interaction using deformable attention based on the endpoint of the predicted trajectory from the previous layer. DeTra \cite{Casas2024} performs iterative updates on object poses to encode agent-agent and agent-scene interactions, where an object pose comprises the multiple future predicted trajectories for each agent. The agent-agent interactions are computed using time self-attention, mode self-attention, and object self-attention. In time self-attention, queries attend only to other queries from the same agent and mode (single trajectory). In mode self-attention, the queries attend only to other queries from the same agent and time, i.e., waypoints between modes (multiple trajectories) at the same time step of a single agent. Finally, in object self-attention, the queries attend only to other queries from other agents at the same time step and mode. Regarding agent-scene interaction, object queries attend to neighboring map tokens using cross-attention.

Interaction modeling approaches in joint perception and prediction range from implicit to explicit methods. Explicit methods are essential to capture the nuanced interactions between agents and between agents and the scene, enabling socially-aware and rule-following future trajectory estimates. Recent developments have introduced innovative ways to integrate these interactions, such as through GNNs and transformers, refining the predictive capabilities of autonomous systems. However, inter-class and intra-class interaction modeling are not yet fully explored. These interactions potentially offer even greater understanding, as different classes of objects exhibit unique movement behaviors and follow varied interaction rules.

% Please add the following required packages to your document preamble:
% \usepackage{multirow}
% \usepackage{graphicx}
%% multirow does not work very well when there are cells in the rows that span more than one line of text! You either use {NiceTabular} of nicematrix, or force a manual fixup :-( % vsantos
\begin{table*}[!htb]
\renewcommand{\arraystretch}{1.4}
\centering
\caption{Summary of the joint perception and prediction approaches according to the \textbf{scene context} level of the taxonomy. Works are sorted in ascending chronological order.}
\label{tab:scene_context_summary}
\resizebox{\textwidth}{!}{%
\begin{tabular}{m{1.75cm} m{1.1cm} m{1.25cm} m{10.15cm}} % Adjust the widths as necessary
% \toprule
\textbf{Scene Context} & \multicolumn{2}{l}{\textbf{Subclasses}}                                                        & \textbf{Works}                                                                                                                                                                                                                                                                                                                                                                                                                                                                                                                                                                                                                                                                                                                                                                                                                                                                                                                                                                                                                                                                                                                                                                      \\ \toprule
\multirow{3}{*}[-2.7em]{\parbox{3cm}{\raggedright Map\\Modeling}}         & \multicolumn{2}{l}{Implicit}                      & FaF \cite{Luo2018}, MotionNet \cite{Wu2020}, STINet \cite{Zhang2020}, FISHING Net \cite{Hendy2020}, SPF2 \cite{Weng2020}, SDP-Net \cite{Zhang2021}, LaserFlow \cite{Meyer2021}, SSPML \cite{Luo2021Pillar}, SDAPNet \cite{Ye2021}, RV-FuseNet \cite{Laddha2021}, Khalil et al. \cite{Khalil2021}, FIERY \cite{Hu2021}, LiCaNext \cite{Khalil2021LiCaNext}, LiCaNet \cite{Khalil2022LiCaNet}, STAN \cite{Wei2022}, BE-STI \cite{Wang2022}, FutureDet \cite{Peri2022}, StretchBEV \cite{Akan2022}, Khurana et al. \cite{Khurana2022}, LidNet \cite{Khalil2022}, WeakMotionNet \cite{Li2023}, Proxy-4DOF \cite{Khurana2023}, PowerBEV \cite{Li2023PowerBEV}, ContrastMotion \cite{Jia2023}, Occ4cast \cite{Liu2023}, MSRM \cite{Wang2024semi}, Fang et al. \cite{Fang2024}, and Wang et al. \cite{Wang2024self}                                                                                                                                                                                                                                                                                                                                                      \\ \cline{2-4} 
                                      & \multirow{2}{*}[-1em]{Explicit}  & HD Maps              & IntentNet \cite{Casas2018}, NMP \cite{Zeng2019}, SpAGNN \cite{Casas2020SpAGNN}, PnPNet \cite{Liang2020}, ILVM \cite{Casas2020ILVM}, PPP \cite{Sadat2020}, Casas et al. \cite{Casas2020}, InteractionTransformer \cite{Li2020}, LiRaNet \cite{Shah2020}, Phillips et al. \cite{Phillips2021}, MVFuseNet \cite{Laddha2021MVFuseNet}, MultiXNet \cite{Djuric2021}, SA-GNN \cite{Luo2021}, LookOut \cite{Cui2021}, Fadadu et al. \cite{Fadadu2022}, FS-GRU \cite{Chen2022}, ViP3D \cite{Gu2023}, ImplicitO \cite{Agro2023}, and DeTra \cite{Casas2024}                                                                                                                                                                                                                                                                                                                                                                                                                                                                                                                                                                                                                               \\ \cline{3-4} 
                                      &                            & Online Semantic Maps & MP3 \cite{Casas2021}, BEVerse \cite{Zhang2022}, ST-P3 \cite{Hu2022}, PIP \cite{Jiang2022}, TBP-Former \cite{Fang2023}, UniAD \cite{Hu2023}, FusionAD \cite{Ye2023}, and VAD \cite{Jiang2023}                                                                                                                                                                                                                                                                                                                                                                                                                                                                                                                                                                                                                                                                                                                                                                                                                                                                                                                                                                                                                                         \\ \hline
\multirow{3}{*}[-1.5em]{\parbox{3cm}{\raggedright Interaction\\Modeling}} & \multicolumn{2}{l}{Implicit}                      & FaF \cite{Luo2018}, IntentNet \cite{Casas2018}, NMP \cite{Zeng2019}, PnPNet \cite{Liang2020}, MotionNet \cite{Wu2020}, FISHING Net \cite{Hendy2020}, PPP \cite{Sadat2020}, SPF2 \cite{Weng2020}, LiRaNet \cite{Shah2020}, SDP-Net \cite{Zhang2021}, LaserFlow \cite{Meyer2021}, MP3 \cite{Casas2021}, SSPML \cite{Luo2021Pillar}, MVFuseNet \cite{Laddha2021MVFuseNet}, MultiXNet \cite{Djuric2021}, SDAPNet \cite{Ye2021}, RV-FuseNet \cite{Laddha2021}, Khalil et al. \cite{Khalil2021}, FIERY \cite{Hu2021}, LiCaNext \cite{Khalil2021LiCaNext}, Fadadu et al. \cite{Fadadu2022}, LiCaNet \cite{Khalil2022LiCaNet}, BEVerse \cite{Zhang2022}, STAN \cite{Wei2022}, BE-STI \cite{Wang2022}, FutureDet \cite{Peri2022}, FS-GRU \cite{Chen2022}, ST-P3 \cite{Hu2022ST-P3}, StretchBEV \cite{Akan2022}, Khurana et al. \cite{Khurana2022}, LidNet \cite{Khalil2022}, TBP-Former \cite{Fang2023}, ViP3D \cite{Gu2023}, WeakMotionNet \cite{Li2023}, ImplicitO \cite{Agro2023}, Proxy-4DOF \cite{Khurana2023}, PowerBEV \cite{Li2023PowerBEV}, ContrastMotion \cite{Jia2023}, Occ4cast \cite{Liu2023}, MSRM \cite{Wang2024semi}, Fang et al. \cite{Fang2024}, and Wang et al. \cite{Wang2024self} \\ \cline{2-4} 
                                      & \multirow[b]{2}{*}{Explicit}  & Agent-Agent          & SpAGNN \cite{Casas2020SpAGNN}, STINet \cite{Zhang2020}, ILVM \cite{Casas2020ILVM}, Casas et al. \cite{Casas2020}, Interaction Transformer \cite{Li2020}, Phillips et al. \cite{Phillips2021}, SA-GNN \cite{Luo2021}, LookOut \cite{Cui2021}, PIP \cite{Jiang2022}, UniAD \cite{Hu2023}, FusionAD \cite{Ye2023}, VAD \cite{Jiang2023}, and DeTra \cite{Casas2024}                                                                                                                                                                                                                                                                                                                                                                                                                                                                                                                                                                                                                                                                                                                                                                                                                      \\ \cline{3-4} 
                                      &                            & Agent-Scene          & Casas et al. \cite{Casas2020}, SA-GNN \cite{Luo2021}, PIP \cite{Jiang2022}, UniAD \cite{Hu2023}, FusionAD \cite{Ye2023}, VAD \cite{Jiang2023}, and DeTra \cite{Casas2024}                                                                                                                                                                                                                                                                                                                                                                                                                                                                                                                                                                                                                                                                                                                                                                                                                                                                                                                                                                                                              \\ \hline
\multirow{6}{*}[-5.0em]{\parbox{3cm}{\raggedright Trajectory\\Modeling}}  & \multirow{4}{*}[-2.8em]{Waypoints} & Unimodal             & FaF \cite{Luo2018}, IntentNet \cite{Casas2018}, NMP \cite{Zeng2019}, PnPNet \cite{Liang2020}, SpAGNN \cite{Casas2020SpAGNN}, STINet \cite{Zhang2020}, InteractionTransformer \cite{Li2020}, SPF2 \cite{Weng2020}, SDP-Net \cite{Zhang2021}, LaserFlow \cite{Meyer2021}, MVFuseNet \cite{Laddha2021MVFuseNet}, SDAPNet \cite{Ye2021}, RV-FuseNet \cite{Laddha2021}, and FS-GRU \cite{Chen2022}                                                                                                                                                                                                                                                                                                                                                                                                                                                                                                                                                                                                                                                                                                                                                                  \\ \cline{3-4} 
                                      &                            & Multimodal           & ILVM \cite{Casas2020ILVM}, Casas et al. \cite{Casas2020}, LiRaNet \cite{Shah2020}, Phillips et al. \cite{Phillips2021}, MultiXNet \cite{Djuric2021}, LookOut \cite{Cui2021}, Fadadu et al. \cite{Fadadu2022}, FutureDet \cite{Peri2022}, PIP \cite{Jiang2022}, UniAD \cite{Hu2023}, ViP3D \cite{Gu2023}, FusionAD \cite{Ye2023}, VAD \cite{Jiang2023}, and DeTra \cite{Casas2024}                                                                                                                                                                                                                                                                                                                                                                                                                                                                                                                                                                                                                                                                                                                                                                                                       \\ \cline{3-4} 
                                      &                            & Deterministic        & FaF \cite{Luo2018}, IntentNet \cite{Casas2018}, NMP \cite{Zeng2019}, PnPNet \cite{Liang2020}, STINet \cite{Zhang2020}, InteractionTransformer \cite{Li2020}, SPF2 \cite{Weng2020}, SDP-Net \cite{Zhang2021}, SDAPNet \cite{Ye2021}, FS-GRU \cite{Chen2022}, PIP \cite{Jiang2022}, and VAD \cite{Jiang2023}                                                                                                                                                                                                                                                                                                                                                                                                                                                                                                                                                                                                                                                                                                                                                                                                                                                         \\ \cline{3-4} 
                                      &                            & Probabilistic        & SpAGNN \cite{Casas2020SpAGNN}, ILVM \cite{Casas2020ILVM}, Casas et al. \cite{Casas2020}, LiRaNet \cite{Shah2020}, LaserFlow \cite{Meyer2021}, Phillips et al. \cite{Phillips2021}, MVFuseNet \cite{Laddha2021MVFuseNet}, MultiXNet \cite{Djuric2021}, RV-FuseNet \cite{Laddha2021}, LookOut \cite{Cui2021}, Fadadu et al. \cite{Fadadu2022}, FutureDet \cite{Peri2022}, UniAD \cite{Hu2023}, ViP3D \cite{Gu2023}, FusionAD \cite{Ye2023}, and DeTra \cite{Casas2024}                                                                                                                                                                                                                                                                                                                                                                                                                                                                                                                                                                                                                                                                                           \\ \cline{2-4} 
                                      & \multicolumn{2}{l}{Displacement Vectors}          & MotionNet \cite{Wu2020}, SSPML \cite{Luo2021Pillar}, Khalil et al. \cite{Khalil2021}, LiCaNext \cite{Khalil2021LiCaNext}, LiCaNet \cite{Khalil2022LiCaNet}, STAN \cite{Wei2022}, BE-STI \cite{Wang2022}, LidNet \cite{Khalil2022}, WeakMotionNet \cite{Li2023}, ContrastMotion \cite{Jia2023}, MSRM \cite{Wang2024semi}, Fang et al. \cite{Fang2024}, and Wang et al. \cite{Wang2024self}                                                                                                                                                                                                                                                                                                                                                                                                                                                                                                                                                                                                                                                                                                                                                                                               \\ \cline{2-4} 
                                      & \multicolumn{2}{l}{Occupancy Flow Maps}           & FISHING Net \cite{Hendy2020}, PPP \cite{Sadat2020}, MP3 \cite{Casas2021}, SA-GNN \cite{Luo2021}, FIERY \cite{Hu2021}, BEVerse \cite{Zhang2022}, ST-P3 \cite{Hu2022}, StretchBEV \cite{Akan2022}, Khurana et al. \cite{Khurana2022}, TBP-Former \cite{Fang2023}, UniAD \cite{Hu2023}, ImplicitO \cite{Agro2023}, Proxy-4DOF \cite{Khurana2023}, FusionAD \cite{Ye2023}, PowerBEV \cite{Li2023PowerBEV}, and Occ4cast \cite{Liu2023}                                                                                                                                                                                                                                                                                                                                                                                                                                                                                                                                                                                                                                                                                                                                                     \\ \bottomrule                                      
\end{tabular}%
}
\end{table*}

\begin{figure*}[!ht]
    \centering
    \begin{subfigure}[b]{\textwidth}
        \centering
        % \includesvg[width=\textwidth,pretex=\sffamily\tiny\hypersetup{pdfborder={0 0 0}}]{figures/temporal_summary_mapmodeling_refs.svg}
        \includegraphics[width=1\textwidth]{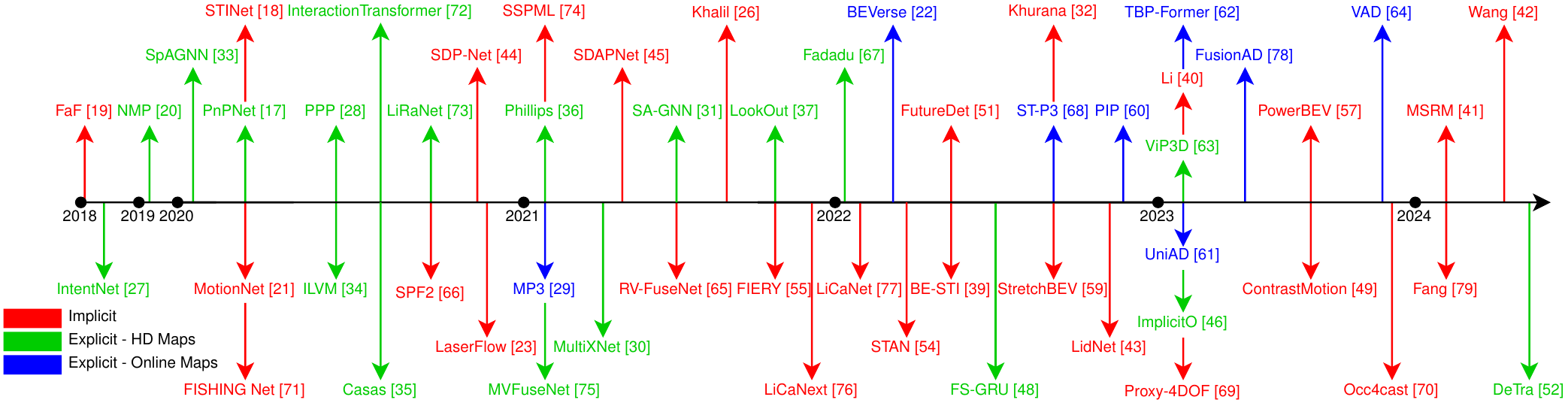}
        \caption{\textbf{Map modeling}.}
        \label{fig:subfig_map_modeling}
    \end{subfigure}
    \par\vspace{0.8cm} % Add consistent vertical spacing
    \begin{subfigure}[b]{\textwidth}
        \centering
        % \includesvg[width=\textwidth,pretex=\sffamily\tiny\hypersetup{pdfborder={0 0 0}}]{figures/temporal_summary_interactionmodeling_refs.svg}
        \includegraphics[width=1\textwidth]{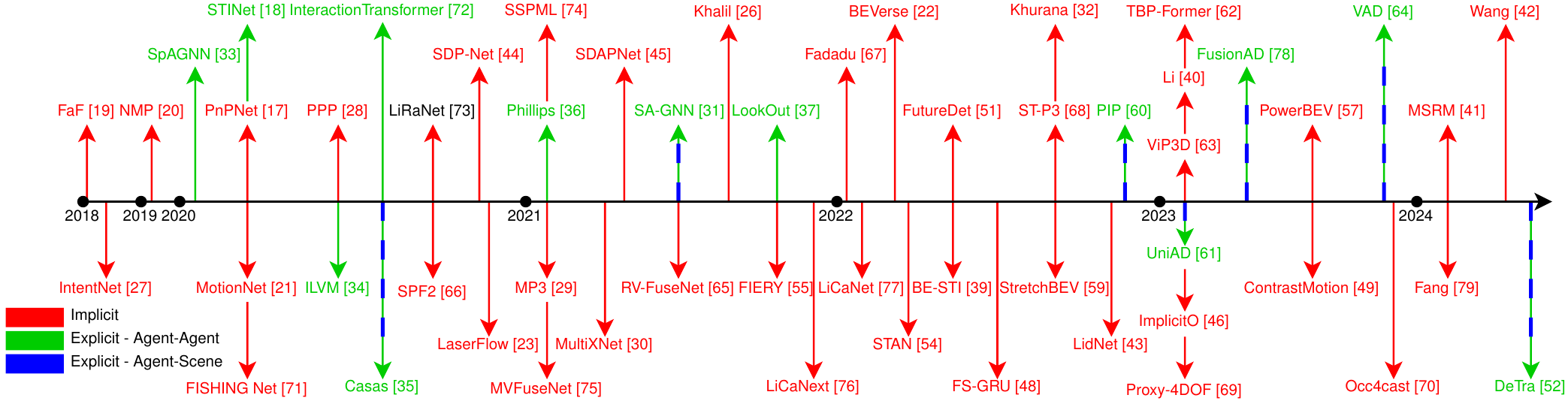}
        \caption{\textbf{Interaction modeling}.}
        \label{fig:subfig_interaction_modeling}
    \end{subfigure}
    \par\vspace{0.8cm} % Add consistent vertical spacing
    \begin{subfigure}[b]{\textwidth}
        \centering
        % \includesvg[width=\textwidth,pretex=\sffamily\tiny\hypersetup{pdfborder={0 0 0}}]{figures/temporal_summary_trajectorymodeling_refs.svg}
        \includegraphics[width=1\textwidth]{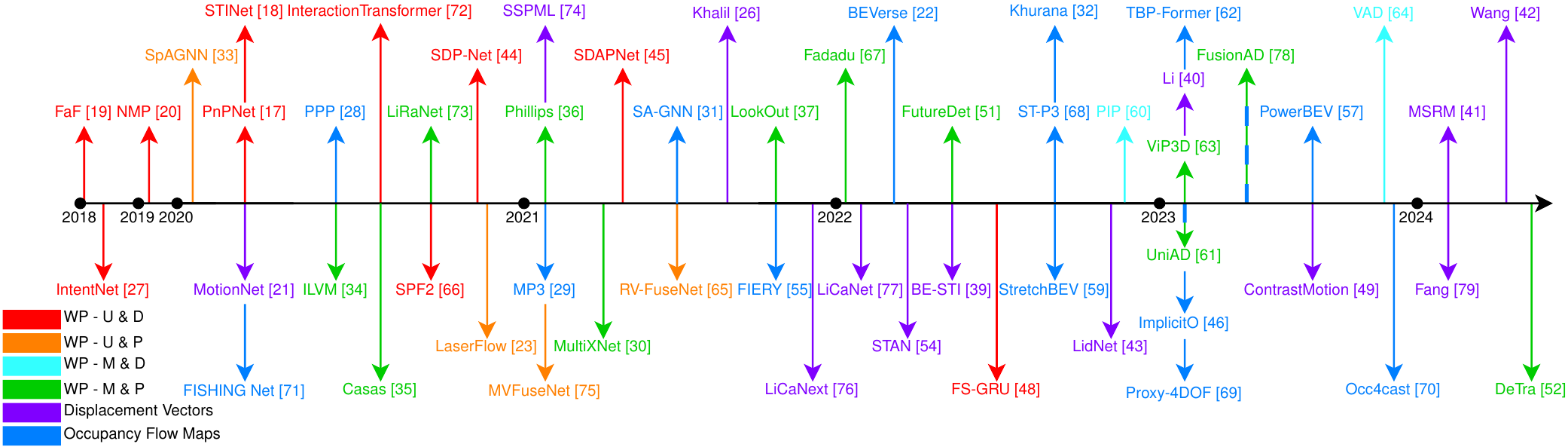}
        \caption{\textbf{Trajectory modeling}. Acronyms: WP - Waypoints, U - Unimodal, M - Multimodal, D - Deterministic, and P - Probabilistic.}
        \label{fig:subfig_trajectory_modeling}
    \end{subfigure}
    \par\vspace{0.4cm} % Add consistent vertical spacing
    \caption{Chronological overview of the joint perception and prediction approaches according to the \textbf{scene context} level of the taxonomy: (a) \textbf{map modeling}, (b) \textbf{interaction modeling}, and (c) \textbf{trajectory modeling}.}
    \label{fig:temporal_summary_scene_context}
\end{figure*}

\subsubsection{Trajectory Modeling} \label{subsubsec:trajectory_modeling}
% Talk about the trajectory modeling such as unimodal, multimodal, waypoints, intentions, gaussians waypoints, etc.
Trajectory modeling is central to predicting the future paths of dynamic agents in autonomous driving scenarios, such as vehicles, pedestrians, and cyclists. Different trajectory modeling methods handle the complexity and uncertainty of these predictions in different ways, with representations typically taking the form of waypoints, displacement vectors, or semantic occupancy flow maps.

% Waypoints: could be bounding boxes, waypoints directly, motion offsets, ,
% Then, the waypoints model could be described by a probability distribution to count with uncertainty. It also can be unimodal or multimodal (multiple hypothetesis). ILVM predicts joint distribution over all actors.
The most commonly used approach for trajectory modeling is based on waypoints, an explicit trajectory prediction output that map the position and orientation of agents within a BEV grid at each future time step. Numerous approaches \cite{Luo2018, Casas2018, Zeng2019, Liang2020, Casas2020SpAGNN, Zhang2020, Casas2020ILVM, Casas2020, Li2020, Weng2020, Shah2020, Zhang2021, Meyer2021, Phillips2021, Laddha2021MVFuseNet, Djuric2021, Ye2021, Laddha2021, Cui2021, Fadadu2022, Peri2022, Chen2022, Jiang2022, Gu2023, Jiang2023, Casas2024, Hu2023, Luo2021, Ye2023} have adopted this format, using waypoints derived directly from model decoders or inferred from predicted bounding boxes and motion offsets. Waypoint-based methods can provide either unimodal predictions or multimodal predictions. Unimodal trajectories \cite{Luo2018, Casas2018, Zeng2019, Casas2020SpAGNN, Liang2020, Zhang2020, Li2020, Weng2020, Zhang2021, Meyer2021, Laddha2021MVFuseNet, Ye2021, Laddha2021, Chen2022} predict a single outcome, emphasizing accuracy for the most likely scenario. In contrast, multimodal trajectory prediction \cite{Casas2020ILVM, Shah2020, Casas2020, Phillips2021, Djuric2021, Cui2021, Fadadu2022, Peri2022, Jiang2022, Gu2023, Jiang2023, Casas2024, Hu2023, Ye2023} generates multiple possible future outcomes, capturing the uncertainty in dynamic environments. These multiple possible trajectories capture important diversity in agent behavior, such as turning right versus continuing straight at an intersection, or deciding to proceed or stop at a yield sign. Additionally, these models can be categorized based on whether they produce deterministic or probabilistic waypoint trajectories. Deterministic methods \cite{Luo2018, Casas2018, Zeng2019, Liang2020, Zhang2020, Li2020, Weng2020, Zhang2021, Ye2021, Chen2022, Jiang2022, Jiang2023} predict future positions or motion offsets, providing a single, specific outcome without accounting for uncertainty. On the other hand, probabilistic models \cite{Casas2020SpAGNN, Casas2020ILVM, Casas2020, Shah2020, Meyer2021, Phillips2021, Laddha2021MVFuseNet, Djuric2021, Laddha2021, Cui2021, Fadadu2022, Peri2022, Gu2023, Casas2024, Hu2023, Ye2023} leverage distributions to account for uncertainty, generating multiple possible future trajectories by representing waypoints as probability distributions. Common choices include Gaussian, Von Mises, Laplace, and various mixture models, enabling a richer representation of potential future paths. The prediction horizon for approaches using this type of trajectory modeling typically spans 3 to 6 seconds, allowing for long-term forecasting.

% One paragraph about displacement vectors approaches.
Some approaches \cite{Wu2020, Khalil2021, Wei2022, Wang2022, Khalil2022, Li2023, Jia2023, Wang2024semi, Wang2024self, Khalil2021LiCaNext, Luo2021Pillar, Khalil2022LiCaNet, Fang2024} modeled the future position of agents as cell-based displacement vectors within a BEV map. Each cell in the BEV map represents motion as a relative displacement vector between time stamps. Cells with similar motion vectors across neighboring positions and consecutive frames are grouped to represent a single agent, thereby maintaining spatial and temporal consistency. However, displacement vector approaches typically do not generate multiple motion hypotheses or incorporate probabilistic modeling, making them less suited for capturing uncertainty compared to waypoint-based methods. Another drawback of these approaches is their limited prediction horizon, which currently extends only up to 1 second.

% One paragraph about occupancy motion segmentations approaches.
Another method of modeling the future states of agents is through semantic occupancy flow maps. The occupancy maps predict the probability of each cell in a BEV grid being occupied or free at a particular time step, and when combined with temporal flow, this forecasting becomes an occupancy flow map. Several studies \cite{Sadat2020, Hendy2020, Casas2021, Luo2021, Hu2021, Zhang2022, Hu2022, Akan2022, Khurana2022, Fang2023, Hu2023, Agro2023, Khurana2023, Ye2023, Li2023PowerBEV, Liu2023} have adopted semantic occupancy flow maps, which offer a flexible representation that captures multiple potential future states in a scene. This representation is achieved by modeling probabilistic distributions for each cell, allowing the map to incorporate uncertainty and reflect variability in possible future occupancy. Approaches using this type of trajectory modeling currently offer prediction horizons ranging from 2 to 7 seconds, enabling long-term predictions.

Modeling the inherent uncertainty in motion prediction using multiple hypotheses and probabilistic distributions is crucial for downstream tasks like motion planning. Autonomous systems rely on these uncertainties to navigate safely through dynamic and dense environments. Given the stochastic nature of future motion, various approaches have been developed to model the future movements of agents in diverse ways. The diversity in trajectory modeling methods, whether using explicit waypoints, cell-based displacement vectors, or semantic occupancy flow maps, provides essential insights for predicting motion in complex scenarios.

\subsubsection*{Summary} \label{subsubsec:summary_scene_context_modeling}
To summarize the scene context modeling section of the taxonomy, \autoref{tab:scene_context_summary} categorizes the approaches based on their map, interaction, and trajectory modeling.
Implicit map and interaction modeling remain the most widely adopted strategies, as they avoid increasing the complexity and parameter count of neural network architectures.
However, in terms of map modeling, the use of HD maps is gaining popularity due to their detailed semantic features, while online semantic maps are also becoming important for their ability to update in real time. Notably, methods exploring the construction of online semantic maps predominantly rely on multi-view images as their primary input, with MP3 \cite{Casas2021} being the exception, using BEV instead.
In explicit interaction modeling, agent-agent interactions are the most commonly explored, with a smaller number of approaches also addressing agent-scene interactions. It is worth noting that all the approaches that model agent-scene interactions also account for agent-agent interactions, but the reverse is not always true.
For trajectory modeling, waypoints are the most prevalent method for predicting future agent movements. This output representation is more mature in researches in the field of perception for autonomous driving, with clear distinctions between unimodal and multimodal methods, as well as deterministic and probabilistic representations. Due to this well-established foundation, it is unsurprising that numerous contributions continue to enhance waypoint-based modeling. Meanwhile, occupancy flow maps are emerging as a significant alternative, offering the ability to capture stochastic motion uncertainty through multimodal and probabilistic modeling, alongside class-agnostic capabilities.

\autoref{fig:temporal_summary_scene_context} presents a chronological overview of the joint perception and prediction approaches based on their map modeling (see \autoref{fig:subfig_map_modeling}), interaction modeling (see \autoref{fig:subfig_interaction_modeling}), and trajectory modeling (see \autoref{fig:subfig_trajectory_modeling}). 
For map modeling, HD maps have been widely used historically and continue to serve as a reliable option today. However, in recent years, online semantic maps have emerged as a viable alternative due to their adaptability and real-time contextual updates, even though they are less detailed than HD maps.
Regarding interaction modeling, despite the critical role of capturing interactions between agents and between agents and the scene, most approaches still avoid explicitly modeling these interactions to reduce model complexity, relying instead on the model to capture them implicitly. Nevertheless, some recent methods have emphasized explicit modeling of both agent-agent and agent-scene interactions, recognizing the importance of modeling these interactions.
In trajectory modeling, waypoint-based approaches dominate the field, but there is a noticeable shift toward displacement vectors and occupancy flow maps. This shift can be attributed to their ability to provide more detailed motion dynamics and spatial occupancy changes over time. Furthermore, waypoint-based approaches that output unimodal trajectories, are often modeled deterministically. In contrast, waypoint-based approaches that output multimodal trajectories are generally modeled probabilistically, as this approach effectively captures uncertainty in future motion and accounts for multiple potential trajectories. Consequently, deterministic modeling for unimodal outputs and probabilistic modeling for multimodal outputs frequently appear together in the literature. Notably, UniAD \cite{Hu2023} and FusionAD \cite{Ye2023} offer hybrid trajectory modeling methods that include both waypoints and occupancy flow maps.

\subsection{Output Representation} \label{subsec:output_representation}
\begin{figure*}[t]
    \centering
   \includegraphics[width=0.75\linewidth]{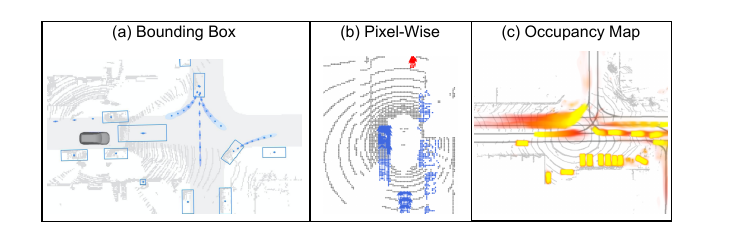}
    % \includesvg[width=0.75\linewidth]{figures/output_representation.svg}
    \caption{Illustration of the \textbf{output representations} used in joint perception and prediction for autonomous driving: (a) \textbf{bounding box}, (b) \textbf{pixel-wise}, and (c) \textbf{occupancy map}. Figure created based on \cite{Djuric2021, Wu2020, Agro2023}.}
    \label{fig:output_representation_illustration}
\end{figure*}

% Try to talk about the output representation in order to talk about the decoding process to achieve these output representations.
The choice of output representation is essential for joint perception and prediction approaches, as it defines how the model decodes and structures its understanding of the environment for downstream tasks like motion planning and risk assessment. The output representations can be categorized into \textbf{bounding box}, \textbf{pixel-wise}, and \textbf{occupancy map} formats, each offering distinct ways to interpret and convey spatial, dynamic, and contextual information about surrounding agents in the scene. \autoref{fig:output_representation_illustration} illustrates these representations. In this section, we examine the unique characteristics and contributions of each output type to accurate and efficient perception and prediction. In addition, we explore the specific decoding strategies employed by each approach within each representation type, which aim to enhance model performance.

\subsubsection{Bounding Box} \label{subsubsec:bbox}
% Talk about the evolution of these approaches. 
The bounding box output representation is a widely adopted approach for predicting the positions and trajectories of dynamic agents in autonomous driving. Models that perform joint perception and prediction begin by detecting objects in the current frame, outputting bounding boxes that capture the spatial extent of each agent, including center coordinates, dimensions, and orientation. To predict future movements, these models extend this representation to output waypoints in the form of motion offsets or as centers of future bounding boxes, effectively constructing a sequence of predicted positions over a preset time horizon. This provides a structured and interpretable tool for visualizing the predicted paths of each agent in the scene.

FaF \cite{Luo2018} was the pioneering approach to use bounding box output representation for joint perception and prediction. They used two branches of CNN as decoder heads: one for binary classification to determine if an object was a vehicle and another to predict bounding boxes in both the current and future frames, resulting in unimodal future trajectories. Inspired by SSD \cite{Liu2016}, they used anchor boxes for each location on the feature map, predicting for each anchor box the corresponding normalized offsets, sizes, and heading parameters. Predefined anchor boxes reduce the variance of the bounding box prediction targets, making the training process more stable.
% As a post-processing step, the model also decodes object tracklets by pooling past and present detections, smoothing them with an average operation when there is overlap.
NMP \cite{Zeng2019}, SDAPNet \cite{Ye2021} and FS-GRU \cite{Chen2022} decoded future trajectories using CNN heads and anchor boxes, similar to FaF, while SDP-Net \cite{Zhang2021} predicted without relying on anchor boxes. IntentNet \cite{Casas2018} decoded detected agents and their future trajectories similarly to FaF, but introduced an intention network that performed multi-class classification over a set of high-level actions, such as keep lane, turn left, turn right, left change lane, right change lane, among others. The intention scores were then fed into a convolutional layer to provide additional features for conditioning future trajectories. SpAGNN \cite{Casas2020SpAGNN} outputs 2D waypoints with positions and orientations, using a CNN decoder that takes RoI-aligned features as input, refined further by a GNN interaction model. The model assumes the marginal distribution of the future waypoints of each actor follows a Gaussian distribution for location and a Von Mises distribution for orientation. LaserFlow \cite{Meyer2021} and RV-FuseNet \cite{Laddha2021} also employ CNN decoders, modeling waypoints with Laplacian distributions.
PnPNet \cite{Liang2020} introduces tracking in the loop by associating current detections with previous tracks, estimating the trajectory to the next frame to reduce false positives and localization error. An LSTM network extracts temporal information from object tracking and feeds it into an MLP to predict future unimodal waypoints. STINet \cite{Zhang2020} combines local geometry, dynamic features, and history path features, feeding them to two MLP heads for current frame bounding boxes (classification and regression) and to another MLP head for unimodal waypoints in future frames. InteractionTransformer \cite{Li2020} outputs one prediction waypoint per step, dependent on the previous waypoint, using a recurrent model with an interaction transformer module and MLPs for waypoint refinement and next-step prediction. SPF2 \cite{Weng2020} inverts the detect-then-forecast pipeline by forecasting future point clouds, followed by detection and tracking on these predicted point clouds to extract object trajectories, decoded through a 2D CNN before applying an off-the-shelf detector and tracker. 

The authors of \cite{Casas2020} introduced multimodal trajectories, by using a mixture of Gaussian distributions, inspired by MTP \cite{Cui2019}, while retaining a decoding process similar to SpAGNN. MultiXNet \cite{Djuric2021} extends IntentNet to predict multimodal trajectories, by adding a refinement stage that crops an RoI around each detected agent to focus on surrounding context. This refined feature is passed through a CNN to produce new waypoints parameterized by Laplace distributions. The MultiXNet decoding process became a foundation for approaches like \cite{Shah2020, Laddha2021MVFuseNet, Fadadu2022}, which focused on encoding improvements. ILVM \cite{Casas2020ILVM} learns a scene latent variable using a GNN, which captures all stochastic information about the scene, including actor interactions, possible behaviors, and environmental factors (e.g., traffic dynamics). Given the input features and the scene-consistent latent variable, the decoder, using an MLP, generates waypoints by considering the underlying dynamics of the entire scene, allowing multiple possible future trajectories for all actors to be decoded simultaneously. Philips et al. \cite{Phillips2021} and LookOut \cite{Cui2021} rely on ILVM for decoding, with LookOut improving this approach by adding a specialized sampling network that enhances the diversity of generated futures, creating a range of most likely scenarios. UniAD \cite{Ye2023} and FusionAD \cite{Ye2023} leverage transformer-based attention mechanisms through their MotionFormer architecture to predict multimodal future trajectories for all agents within a scene, refining each trajectory through successive layers of self, cross and deformable attention. During training, a non-linear smoother applies kinematic constraints to enhance the realism of trajectories, guiding them with both scene and agent-level positional anchors. DeTra \cite{Casas2024} uses a single-layer bi-directional GRU to process temporal features for each mode and agent in parallel, with an MLP predicting location and scale of an isotropic Laplacian distributions from the GRU hidden state at each time step. FutureDet \cite{Peri2022} extends a robust object detector \cite{Yin2021} to detect objects in future, unobserved LiDAR sweeps with ground truth supervision, followed by backcasting each future detection back to the current timestep detection. This process is the reverse of forecasting position offsets from current-frame detections and associates multiple possible future positions with a single current detection, naturally producing multi-future predictions.

All the previous approaches use regression-based trajectory prediction, generating continuous paths by directly outputting future waypoints of an agent and minimizing the error between predicted and ground truth trajectory points. ViP3D \cite{Gu2023} and PIP \cite{Jiang2022} support not only regression-based approaches but also goal-based and heatmap-based methods. Goal-based decoding predicts possible trajectory endpoints first, refining paths to align with these goals, while heatmap-based decoding generates spatial heatmaps that represent end position probabilities, completing trajectories based on these points. VAD \cite{Jiang2023} uses a vectorized scene representation, avoiding explicit waypoints by outputting multimodal motion vectors to represent future agent trajectories. 

In summary, bounding box-based output representation is the most widely used method for joint perception and prediction approaches. These range from unimodal to multimodal trajectories, with a variety of decoding architectures and probabilistic models for waypoints. However, this representation depends on predefined object classes, limiting its ability to predict dynamic objects not present in the training set. Additionally, bounding boxes often rely on predefined anchor boxes, which can hinder detection of small and distant objects. Nevertheless, relying on object classes simplifies modeling intra-class and inter-class interactions, which is important since different dynamic object classes exhibit distinct motion behaviors.

\subsubsection{Pixel-Wise} \label{subsubsec:pixel-wise}

The pixel-wise output representation in joint perception and prediction involves classifying each cell in a BEV map and predicting displacement vectors that describe the motion of each individual cell. Each cell may be classified as static or dynamic, or assigned a specific object class. Cells with similar motion vectors across neighboring positions and consecutive frames are grouped to represent single agents, maintaining both spatial and temporal consistency.

MotionNet \cite{Wu2020} pioneered the use of this output representation for joint perception and prediction. After feature extraction through a Spatio-Temporal Pyramid Network (STPN), it applies three output heads using two-layer 2D convolutions: cell classification, motion prediction, and state estimation. The cell classification head categorizes each cell (e.g., vehicle, pedestrian, or background), the motion prediction head forecasts future cell positions as displacement vectors over time, and the state estimation head determines whether a cell is static or moving. The final output is a BEV map matching the resolution of the input. To refine predicted motion and prevent minor disturbances in static cells, such as background areas or stationary vehicles, the outputs of the state estimation and cell classification heads are incorporated into the motion prediction head. Specifically, if a cell is classified as background by the cell classification head or as static by the state estimation head, the motion prediction for that cell is set to zero. MotionNet established a critical foundation in this field, and subsequent approaches \cite{Khalil2021, Khalil2022LiCaNet, Khalil2021LiCaNext} used the STPN and output heads from MotionNet as backbone networks, instead focusing on fusion processes for different input representations. LidNet \cite{Khalil2022} also built on MotionNet’s architecture but enhanced the STPN feature extractor, while the authors of \cite{Wei2022} proposed a transformer-based backbone to replace STPN, feeding it to the same three output heads as MotionNet. BE-STI \cite{Wang2022} introduced additional stages: a spatial semantic decoder and a temporal motion decoder. The spatial semantic decoder takes the multi-scale spatial features extracted by a backbone network and merges them through a bottom-up semantic decoder, supervised by a semantic segmentation task to strengthen spatial feature learning. The temporal motion decoder stage processes the upsampled multi-scale features from the semantic decoder, applying spatiotemporal enhancement to capture discriminative spatial features, which are upsampled in the same manner as in the semantic decoder. The output from the temporal motion decoder is then fed into the same output heads as in MotionNet.

While previous approaches rely on supervised learning, more recent work has addressed the challenges of requiring large amounts of labeled data by introducing weakly supervised, semi-supervised and self-supervised learning methods. Luo et al. \cite{Luo2021Pillar} applied self-supervised learning, combining structural consistency and cross-sensor regularization (using camera images) to predict a dense, class-agnostic motion for each BEV cell over time. This structural consistency loss aligns spatial positions of pillars across frames, while cross-sensor regularization uses optical flow from images to refine alignment by compensating for ego-vehicle movement. ContrastMotion \cite{Jia2023} introduced a pillar association technique that predicts pillar correspondence probabilities by analyzing feature distances between frames, which are then used to estimate motion. This linear extrapolation assumes relatively constant velocities over short intervals, a reasonable assumption in autonomous driving scenarios where objects typically exhibit steady movement over brief periods. The authors of \cite{Fang2024} focused on novel self-supervised losses for dynamic object motion prediction, employing Masked Chamfer Distance Loss to align dynamic points, Piecewise Rigidity Loss to ensure consistent motion within rigid objects, and Temporal Consistency Loss to smooth trajectories across frames. WeakMotionNet \cite{Li2023} uses MotionNet as a backbone but adds a two-stage weakly supervised framework. Stage 1, called PreSegNet, trains a foreground/background (FG/BG) segmentation network on partially annotated FG/BG masks to classify each BEV cell. In Stage 2, WeakMotionNet performs motion prediction using only moving foreground cells. Expanding this idea, the authors of \cite{Wang2024semi} implemented a semi-supervised framework with a Mean-Teacher model for pseudo-labeling, where the teacher generates pseudo-labels for motion vectors that the student then learns to predict. Finally, the authors of \cite{Wang2024self} introduced a pseudo motion labeling approach using an optimal transport solver to find correspondences between BEV cells across frames, with each BEV cell’s movement defining its displacement vector.

The pixel-wise output representation enables fine-grained, dense motion prediction across BEV maps, with approaches ranging from fully supervised to self-supervised methods. Compared to bounding-box-based methods, this design improves perception of unseen objects by decomposing regions into grid cells that capture shared local features, and avoids object proposals and Non-Maximum Suppression (NMS) that can miss uncertain detections \cite{Wu2020}. However, while pixel-wise approaches capture detailed motion, they may struggle to maintain coherence between cells for larger objects, potentially leading to fragmented predictions.

\begin{table*}[t]  % Use table* to span both columns
\renewcommand{\arraystretch}{1.4}
\centering
\caption{Summary of the joint perception and prediction approaches according to the \textbf{output representation} level of the taxonomy. Works are sorted in ascending chronological order.}
\label{tab:summary_output_representation}
\resizebox{\textwidth}{!}{%
\begin{tabular}{m{1.9cm}m{4.3cm}m{9.6cm}}  % Adjust widths to fit content
% \toprule
\textbf{Output\newline Representation} & \textbf{Characteristics}                                                                                                     & \textbf{Works}                                                                                                                                                                                                                                                                                                                                                                                                                                                                                                                                                                                                                                                                                                                                                           \\ \toprule
Bounding Box                   & Object’s spatial extent (position, dimensions, and orientation), with sequential waypoints for predicting future trajectory. & FaF \cite{Luo2018}, IntentNet \cite{Casas2018}, NMP \cite{Zeng2019}, SpAGNN \cite{Casas2020SpAGNN}, PnPNet \cite{Liang2020}, STINet \cite{Zhang2020}, ILVM \cite{Casas2020ILVM}, Casas et al. \cite{Casas2020}, InteractionTransformer \cite{Li2020}, SPF2 \cite{Weng2020}, LiRaNet \cite{Shah2020}, SDP-Net \cite{Zhang2021}, LaserFlow \cite{Meyer2021}, Phillips et al. \cite{Phillips2021}, MVFuseNet \cite{Laddha2021MVFuseNet}, MultiXNet \cite{Djuric2021}, SDAPNet \cite{Ye2021}, RV-FuseNet \cite{Laddha2021}, LookOut \cite{Cui2021}, Fadadu et al. \cite{Fadadu2022}, FutureDet \cite{Peri2022}, FS-GRU \cite{Chen2022}, PIP \cite{Jiang2022}, UniAD \cite{Hu2023}, ViP3D \cite{Gu2023}, FusionAD \cite{Ye2023}, VAD \cite{Jiang2023}, and DeTra \cite{Casas2024} \\ \hline
Pixel-Wise                     & Classifies each cell in a grid and assigns displacement vectors to describe the motion of each cell individually.            & MotionNet \cite{Wu2020}, SSPML \cite{Luo2021Pillar}, Khalil et al. \cite{Khalil2021}, LiCaNext \cite{Khalil2021LiCaNext}, LiCaNet \cite{Khalil2022LiCaNet}, STAN \cite{Wei2022}, BE-STI \cite{Wang2022}, LidNet \cite{Khalil2022}, WeakMotionNet \cite{Li2023}, ContrastMotion \cite{Jia2023}, MSRM \cite{Wang2024semi}, Fang. et al. \cite{Fang2024}, and Wang et al. \cite{Wang2024self}                                                                                                                                                                                                                                                                                                                                                                                       \\ \hline
Occupancy Map                     & Occupancy state of each grid cell and predicts cell movements over time.                                                     & FISHING Net \cite{Hendy2020}, PPP \cite{Sadat2020}, MP3 \cite{Casas2021}, SA-GNN \cite{Luo2021}, FIERY \cite{Hu2021}, BEVerse \cite{Zhang2022}, ST-P3 \cite{Hu2022}, StretchBEV \cite{Akan2022}, Khurana et al. \cite{Khurana2022}, TBP-Former \cite{Fang2023}, UniAD \cite{Hu2023}, ImplicitO \cite{Agro2023}, Proxy-4DOF \cite{Khurana2023}, FusionAD \cite{Ye2023}, PowerBEV \cite{Li2023PowerBEV}, and Occ4cast \cite{Liu2023}                                                                                                                                                                                                                                                                                                                                \\ \bottomrule
\end{tabular}%
}
\end{table*}

\subsubsection{Occupancy Map} \label{subsubsec:occupancy_map}

Occupancy maps and occupancy flow maps are powerful representations for detecting and predicting dynamic objects in a scene. An occupancy map captures the occupancy state of space within a grid, indicating whether each cell is occupied or free at a given time. Occupancy flow maps extend this concept by predicting the movement of occupied cells over time, providing motion information for dynamic objects across multiple future frames. These representations are advantageous as they encompass the entire scene as a whole by distinguishing between occupied and unoccupied spaces, including objects that may not be entirely visible or labeled. This leads to a more complete understanding of the environment and the movement of agents within it.

The first approach to explore occupancy flow maps was FISHING Net \cite{Hendy2020}, which used multiple sensor modalities to independently predict semantic grids for current and future frames. An aggregation function (average or priority pooling) then fused these grids into a unified top-down occupancy map across multiple time frames, outputting probabilities for each cell. PPP \cite{Sadat2020} introduced a recurrent mechanism that updates occupancy predictions by incorporating previous predictions with new features from the encoding network, which are then processed by a softmax layer to generate a categorical distribution across grid cells, indicating class-specific occupancy at each future timestep. MP3 \cite{Casas2021} extends PPP by predicting 2D velocity vectors for each occupied cell, estimating future occupancy by iteratively updating each cell based on the likelihood of movement from neighboring cells. This approach models occupancy using probabilistic distributions for each object class, accounting for multiple potential behaviors. SA-GNN \cite{Luo2021} was the first to model interactions explicitly, using a CNN to independently predict the probability of each actor’s location at future timesteps, followed by an integration step that combines these predictions with scene-level features to produce a unified occupancy map for each future frame.

FIERY \cite{Hu2021} introduced a decoder network that served as a foundation for several subsequent approaches \cite{Zhang2022, Hu2022, Akan2022, Fang2023, Li2023PowerBEV}. FIERY employed a ConvGRU followed by residual blocks to recursively predict future states from the current state and a latent code sampled from diagonal Gaussian distributions. A decoder network with multiple CNN heads then predicted semantic segmentation, instance centerness, instance offset, and future instance flow. For each future timestep, instance centerness locates instance centers, offsets and the segmentation map assign neighboring pixels to these centers for instance segmentation, and future flow is used to track instances over time, ensuring consistency through Hungarian matching \cite{Kuhn1955} across frames. BEVerse \cite{Zhang2022} builds on FIERY, predicting a unique latent map for each BEV cell rather than a single global latent vector, enabling the capture of uncertainties across different objects. ST-P3 \cite{Hu2022} uses the same output heads as FIERY but introduces a dual pathway prediction strategy, with one pathway processing historical BEV features through a GRU and the other using Gaussian samples for GRU inputs. StretchBEV \cite{Akan2022} improves long-sequence prediction diversity by sampling at each timestep rather than once for the entire sequence, as FIERY does, allowing it to capture complex future interactions. TBP-Former \cite{Fang2023} introduces a Spatial-Temporal Pyramid Transformer (STPT) encoder to feed FIERY’s decoder heads. PowerBEV \cite{Li2023PowerBEV} simplifies FIERY to use only two output heads: a semantic segmentation map and a backward centripetal flow. The semantic segmentation map directly identifies object centers without needing a separate centerness head, while the backward centripetal flow links each pixel to its prior position, allowing efficient pixel-level association across frames. This approach simplifies instance tracking by reducing dependency on forward flow and Hungarian matching, enhancing stability, accuracy, and robustness against tracking errors.

UniAD \cite{Hu2023} diverges from traditional RNN-based methods, which often compress features and require complex post-processing. Its OccFormer architecture uses dense scene features with agent-specific information, attention-based pixel-agent interactions, and sequential blocks to directly predict instance-wise occupancy. This method employs self- and cross-attention with a constrained mask to align pixels with agents, producing detailed agent-specific occupancy maps without clustering. FusionAD \cite{Ye2023} builds on the UniAD model, with its primary contribution focused on integrating multiple input modalities. ImplicitO \cite{Agro2023} predicts occupancy and flow at spatio-temporal query points by gathering local features around each query, employing deformable sampling to capture relevant context, and using cross-attention to focus on the most important information for each prediction. These aggregated features are then processed by a ResNet-based network with separate heads for occupancy and flow predictions.

Occupancy maps separate environmental motion from ego-vehicle motion, but full 3D occupancy is challenging to observe directly due to occlusions. To address this, Khurana et al. \cite{Khurana2022} apply differentiable raycasting, incorporated into the network architecture proposed by NMP \cite{Zeng2019}, to convert future occupancy predictions into LiDAR sweep predictions, enabling self-supervised learning by comparing these with ground-truth sweeps and allowing occupancy to emerge as an internal representation within the network. By shifting the goal from semantic occupancy to geometric occupancy, identifying whether a location is occupied without specifying the type of object occupying it, BEV occupancy can be learned effectively from unannotated LiDAR sequences. The authors of \cite{Khurana2023} further simplified this raycasting approach by reshaping the 4D voxel grid (X$\times$Y$\times$Z$\times$T) into a 3D format (X$\times$Y$\times$ZT), treating the height and time dimensions as channels to enable efficient 2D convolutions. PCF \cite{Liu2023} advances these previous works by using ConvLSTMs, replacing the concatenation step with recurrent temporal processing and a shared 2D convolutional encoder across frames.

Occupancy flow maps model the space of all objects occupy and move through, offering a more complete representation of both static and dynamic elements in the scene. Occupancy flow maps offer advantages over bounding box-based approaches by capturing diverse object shapes and effectively handling low-confidence detections \cite{Sadat2020}. Although pixel-wise approaches offer a dense motion field, their parameterization does not capture multimodal behaviors, a limitation mitigated by occupancy flow maps, which can use probabilistic distributions to represent the uncertainty of multimodal interactions \cite{Casas2021}. However, occupancy flow maps also present challenges. Predicting motion on a grid-based map can be computationally intensive, particularly for high-resolution grids or extended temporal horizons. Furthermore, most occupancy-based approaches lack object-specific information, making it difficult to distinguish between different dynamic agents based solely on occupancy.

\begin{figure*}
    \centering
    \includegraphics[width=1\linewidth]{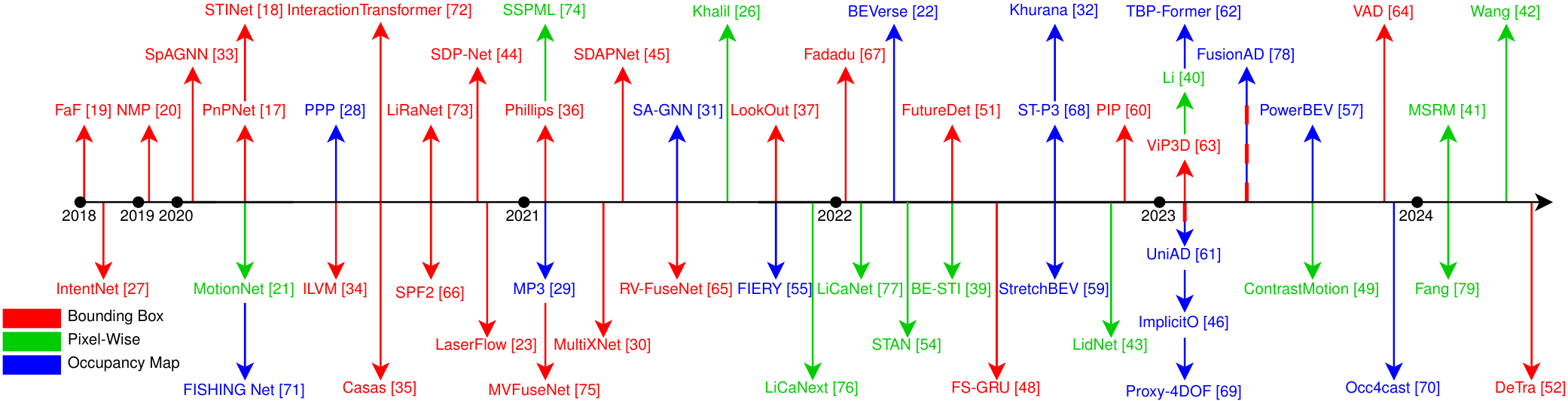}
     %the \hypersetup{pdfborder={0 0 0}} modifier included to avoid showing locally the hyperlink boxes. %%vsantos
% \includesvg[width=1\textwidth,pretex=\sffamily\tiny\hypersetup{pdfborder={0 0 0}}]{figures/temporal_summary_outputrepresentation_refs.svg}   

% \input{figures/gentimelineB} 
% \todo[inline]{The UniAD is hard to spot as a dual approach technique (arrow too short). Also you could perhaps thicken the line for FusionAD so the dual approach is easier to catch... just a suggestion!}
\caption{Chronological overview of the joint perception and prediction approaches according to the \textbf{output representation} level of the taxonomy.}
% Below an attempt to create automatically this timeline after a data file... it's a sample only (wrong categories), but it's highly configurable (just edit the data file): check if it is useful and viable.}
\label{fig:temporal_summary_output_representation}
\end{figure*}

\subsubsection*{Summary} \label{subsubsec:summary_output_representation}

\autoref{tab:summary_output_representation} summarizes the joint perception and prediction approaches according to the output representation level of the taxonomy. This table reveals that bounding boxes are significantly more common than pixel-wise and occupancy map representations. This is unsurprising, as the bounding box output representation is a more established and mature approach compared to the other two. Notably, UniAD \cite{Hu2023} and FusionAD \cite{Ye2023} appear under both bounding box and occupancy map categories, as their networks can predict both outputs within a single model. 

\autoref{fig:temporal_summary_output_representation} provides a chronological overview of these approaches based on output representation. While bounding box (red arrows) representations were predominant in the early stages of joint perception and prediction research, recent trends show a clear shift toward pixel-wise (green arrows) and occupancy map (blue arrows) representations. This transition can be attributed to the class-agnostic capability of pixel-wise and occupancy map representations, which offer greater flexibility compared to the bounding box representation. The adoption of self-supervised learning paradigms in joint perception and prediction has further emphasized the advantages of class-agnostic capabilities, accelerating this shift and reducing the dependence on large-scale, human-annotated datasets, which are both resource-intensive and costly to produce. It is worth noting that UniAD and FusionAD are represented with dual-colored arrows (red and blue) to indicate their dual output capabilities.

\section{Evaluation} \label{sec:evaluation}

This section evaluates the approaches for joint perception and prediction in autonomous driving through both qualitative analysis and quantitative comparison. The \textbf{qualitative analysis} provides a detailed one-by-one overview of all works discussed in this survey. The \textbf{quantitative comparison} evaluates the approaches based on their output representations, as the evaluation metrics are inherently tied to the nature of the outputs, whether they are bounding boxes, pixel-wise predictions, or occupancy maps.

% Please add the following required packages to your document preamble:
% \usepackage{graphicx}
\makesavenoteenv{table*} % Allow footnotes in table environment
% Suggestion with colors in the xmark and cmark :-). %vsantos
\newcommand{\xxmark}{\textcolor{red}{\xmark}}
\newcommand{\ccmark}{\textcolor{green}{\cmark}}
\begin{table*}[t]
\renewcommand{\arraystretch}{1.15}
\centering
\caption{Qualitative analysis between all 55 approaches of joint perception and prediction. Works are sorted in ascending chronological order. Columns abbreviations are in the footnote of the table.}
\label{tab:qualitative_analysis}
\begin{threeparttable}
\resizebox{\textwidth}{!}{%
\begin{tabular}{lcllllcclllc}
\textbf{Work}                        & \textbf{Year} & \textbf{Input\tnote{1}}   & \textbf{Map\tnote{2}} & \textbf{Interaction\tnote{3}} & \textbf{Output\tnote{4}} & \textbf{Agents\tnote{5}} & \textbf{PH (s)\tnote{6}} & \textbf{Base Architectures\tnote{7}} & \textbf{Supervision} & \textbf{Dataset/Simulator\tnote{8}}   & \textbf{Code\tnote{9}}                                                             \\ \toprule
FaF \cite{Luo2018}                   & 2018          & BEV              & Implicit     & Implicit             & BBox            & V               & 1               & CNN                         & Full                & ATG4D                        & \xxmark                                                                    \\
IntentNet \cite{Casas2018}           & 2018          & BEV              & HD Map       & Implicit             & BBox            & V               & 3               & CNN                         & Full                & ATG4D                        & \xxmark                                                                    \\ \hline
NMP \cite{Zeng2019}                  & 2019          & BEV              & HD Map       & Implicit             & BBox            & V               & 3               & CNN                         & Full                & ATG4D                        & \xxmark                                                                    \\ \hline
SpAGNN \cite{Casas2020SpAGNN}        & 2020          & BEV              & HD Map       & A-A                  & BBox            & V               & 3               & CNN, GNN, GRU               & Full                & nuScenes, ATG4D              & \xxmark                                                                    \\
PnPNet \cite{Liang2020}              & 2020          & BEV              & HD Map       & Implicit             & BBox            & V, P            & 3               & CNN, LSTM                   & Full                & nuScenes, ATG4D              & \xxmark                                                                    \\
MotionNet \cite{Wu2020}              & 2020          & BEV              & Implicit     & Implicit             & P-W             & V, P, B         & 1               & CNN                         & Full                & nuScenes                     & \href{https://github.com/pxiangwu/MotionNet}{\ccmark}                      \\
STINet \cite{Zhang2020}              & 2020          & BEV              & Implicit     & A-A                  & BBox            & P               & 3               & CNN, GNN                    & Full                & Lyft, WOD                    & \xxmark                                                                    \\
FISHING Net \cite{Hendy2020}         & 2020          & BEV, MVI, RADAR  & Implicit     & Implicit             & Occ             & V, P, B         & 2               & CNN                         & Full                & nuScenes, Lyft               & \xxmark                                                                    \\
ILVM \cite{Casas2020ILVM}            & 2020          & BEV              & HD Map       & A-A                  & BBox            & V               & 5               & CNN, GNN, GRU               & Full                & nuScenes, ATG4D              & \xxmark                                                                    \\
PPP \cite{Sadat2020}                 & 2020          & BEV              & HD Map       & Implicit             & Occ             & V, P, B         & 5               & CNN                         & Full                & ATG4D                        & \xxmark                                                                    \\
Casas et al. \cite{Casas2020}        & 2020          & BEV              & HD Map       & A-A/A-S              & BBox            & V               & 5               & CNN, GNN, GRU               & Full                & nuScenes, ATG4D              & \xxmark                                                                    \\
Inter.Trans. \cite{Li2020}           & 2020          & BEV, FVI         & HD Map       & A-A                  & BBox            & V               & 3               & CNN, T                 & Full                & nuScenes, ATG4D              & \xxmark                                                                    \\
SPF2 \cite{Weng2020}                 & 2020          & RV               & Implicit     & Implicit             & BBox            & C-A             & 3               & CNN, LSTM                   & Self                 & nuScenes, KITTI              & \xxmark                                                                    \\
LiRaNet \cite{Shah2020}              & 2020          & BEV, RADAR       & HD Map       & Implicit             & BBox            & V               & 3               & CNN                         & Full                & nuScenes, X17k               & \xxmark                                                                    \\
SDP-Net \cite{Zhang2021}             & 2020          & BEV              & Implicit     & Implicit             & BBox            & V               & 0.5             & CNN                         & Full                & KITTI                        & \xxmark                                                                    \\
LaserFlow \cite{Meyer2021}           & 2020          & RV               & Implicit     & Implicit             & BBox            & V, P, B         & 3               & CNN                         & Full                & nuScenes, ATG4D              & \xxmark                                                                    \\ \hline
Phillips et al. \cite{Phillips2021}  & 2021          & BEV              & HD Map       & A-A                  & BBox            & V, P            & 5               & CNN, GNN, GRU               & Full                & ATG4D, LP3                   & \xxmark                                                                    \\
MP3 \cite{Casas2021}                 & 2021          & BEV              & OSM          & Implicit             & Occ             & V, P, B         & 5               & CNN                         & Full                & URBANEXPERT                  & \xxmark                                                                    \\
SSPML \cite{Luo2021Pillar}           & 2021          & BEV, FVI         & Implicit     & Implicit             & P-W             & C-A             & 0.5             & CNN                         & Self                 & nuScenes                     & \href{https://github.com/qcraftai/pillar-motion}{\ccmark}                  \\
MVFuseNet \cite{Laddha2021MVFuseNet} & 2021          & BEV, RV          & HD Map       & Implicit             & BBox            & V, P, B         & 3               & CNN                         & Full                & nuScenes                     & \xxmark                                                                    \\
MultiXNet \cite{Djuric2021}          & 2021          & BEV              & HD Map       & Implicit             & BBox            & V, P, B         & 3               & CNN                         & Full                & nuScenes, ATG4D              & \xxmark                                                                    \\
SDAPNet \cite{Ye2021}                & 2021          & BEV              & Implicit     & Implicit             & BBox            & V               & 2.5             & CNN                         & Full                & nuScenes                     & \xxmark                                                                    \\
SA-GNN \cite{Luo2021}                & 2021          & BEV              & HD Map       & A-A/A-S              & Occ             & P               & 7               & CNN, GNN                    & Full                & nuScenes, ATG4D              & \xxmark                                                                    \\
RV-FuseNet \cite{Laddha2021}         & 2021          & RV               & Implicit     & Implicit             & BBox            & V               & 3               & CNN                         & Full                & nuScenes                     & \xxmark                                                                    \\
Khalil et al. \cite{Khalil2021}      & 2021          & BEV, RV          & Implicit     & Implicit             & P-W             & V, P, B         & 1               & CNN                         & Full                & nuScenes                     & \xxmark                                                                    \\
FIERY \cite{Hu2021}                  & 2021          & MVI              & Implicit     & Implicit             & Occ             & V               & 2               & CNN, GRU                    & Full                & nuScenes, Lyft               & \href{https://github.com/wayveai/fiery}{\ccmark}                           \\
LookOut \cite{Cui2021}               & 2021          & BEV              & HD Map       & A-A                  & BBox            & V, P, B         & 5               & CNN, GNN, VAE               & Full                & ATG4D                        & \xxmark                                                                    \\
LiCaNext \cite{Khalil2021LiCaNext}   & 2021          & BEV, FVI, RV, RI & Implicit     & Implicit             & P-W             & V, P, B         & 1               & CNN                         & Full                & nuScenes                     & \xxmark                                                                    \\ \hline
Fadadu et al. \cite{Fadadu2022}      & 2022          & BEV, FVI, RV     & HD Map       & Implicit             & BBox            & V, P, B         & 3               & CNN                         & Full                & nuScenes, ATG4D              & \xxmark                                                                    \\
LiCaNet \cite{Khalil2022LiCaNet}     & 2022          & BEV, FVI, RV     & Implicit     & Implicit             & P-W             & V, P, B         & 1               & CNN                         & Full                & nuScenes                     & \xxmark                                                                    \\
BEVerse \cite{Zhang2022}             & 2022          & MVI              & OSM          & Implicit             & Occ             & V, P, B         & 2               & T, CNN, GRU            & Full                & nuScenes                     & \href{https://github.com/zhangyp15/BEVerse}{\ccmark}                       \\
STAN \cite{Wei2022}                  & 2022          & BEV              & Implicit     & Implicit             & P-W             & V, P, B         & 1               & CNN, T                 & Full                & nuScenes                     & \xxmark                                                                    \\
BE-STI \cite{Wang2022}               & 2022          & BEV              & Implicit     & Implicit             & P-W             & V, P, B         & 1               & CNN                         & Full                & nuScenes, WOD                & \xxmark                                                                    \\
FutureDet \cite{Peri2022}            & 2022          & BEV              & Implicit     & Implicit             & BBox            & V, P            & 3               & CNN                         & Full                & nuScenes                     & \href{https://github.com/neeharperi/FutureDet}{\ccmark}                    \\
FS-GRU \cite{Chen2022}               & 2022          & BEV              & HD Map       & Implicit             & BBox            & V               & 3               & CNN, GRU                    & Full                & Argoverse                    & \xxmark                                                                    \\
ST-P3 \cite{Hu2022}                  & 2022          & MVI              & OSM          & Implicit             & Occ             & V, P            & 2               & CNN, GRU                    & Full                & nuScenes, CARLA              & \href{https://github.com/OpenPerceptionX/ST-P3}{\ccmark}                   \\
StretchBEV \cite{Akan2022}           & 2022          & MVI              & Implicit     & Implicit             & Occ             & V               & 6               & CNN, GRU                    & Full                & nuScenes                     & \href{https://github.com/kaanakan/stretchbev}{\ccmark}                     \\
Khurana et al. \cite{Khurana2022}    & 2022          & BEV              & Implicit     & Implicit             & Occ             & C-A             & 3               & CNN                         & Self                & nuScenes                     & \href{https://github.com/tarashakhurana/emergent-occ-forecasting}{\ccmark} \\
LidNet \cite{Khalil2022}             & 2022          & BEV              & Implicit     & Implicit             & P-W             & V, P, B         & 1               & CNN                         & Full                & nuScenes                     & \xxmark                                                                    \\
PIP \cite{Jiang2022}                 & 2022          & MVI              & OSM          & A-A/A-S              & BBox            & V               & 6               & CNN, T                 & Full                & nuScenes                     & \xxmark                                                                    \\ \hline
TBP-Former \cite{Fang2023}           & 2023          & MVI              & OSM          & Implicit             & Occ             & V, P            & 2               & CNN, T                 & Full                & nuScenes                     & \href{https://github.com/MediaBrain-SJTU/TBP-Former}{\ccmark}              \\
UniAD \cite{Hu2023}                  & 2023          & MVI              & OSM          & A-A/A-S              & BBox/Occ        & V, P, B         & 6               & CNN, T                 & Full                & nuScenes                     & \href{https://github.com/OpenDriveLab/UniAD}{\ccmark}                      \\
ViP3D \cite{Gu2023}                  & 2023          & MVI              & HD Map       & Implicit             & BBox            & V, P            & 6               & T                      & Full                & nuScenes                     & \href{https://github.com/Tsinghua-MARS-Lab/ViP3D}{\ccmark}                 \\
WeakMotionNet \cite{Li2023}          & 2023          & BEV              & Implicit     & Implicit             & P-W             & C-A             & 1               & CNN                         & Weak               & nuScenes, WOD                & \href{https://github.com/L1bra1/WeakMotion}{\ccmark}                       \\
ImplicitO \cite{Agro2023}            & 2023          & BEV              & HD Map       & Implicit             & Occ             & V               & 5               & CNN, T                 & Full                & Argoverse 2, High-waysim     & \xxmark                                                                    \\
Proxy-4DOF \cite{Khurana2023}        & 2023          & 3D Voxel Grid    & Implicit     & Implicit             & Occ             & C-A             & 3               & CNN                         & Self                 & nuScenes, KITTI, Argoverse 2 & \href{https://github.com/tarashakhurana/4d-occ-forecasting}{\ccmark}       \\
FusionAD \cite{Ye2023}               & 2023          & BEV, MVI         & OSM          & A-A/A-S              & BBox/Occ        & V, P, B         & 6               & CNN, T                 & Full                & nuScenes                     & \href{https://github.com/westlake-autolab/FusionAD}{\ccmark}               \\
PowerBEV \cite{Li2023PowerBEV}       & 2023          & MVI              & Implicit     & Implicit             & Occ             & V               & 2               & CNN                         & Full                & nuScenes                     & \href{https://github.com/edwardleelpz/powerbev}{\ccmark}                   \\
ContrastMotion \cite{Jia2023}        & 2023          & BEV              & Implicit     & Implicit             & P-W             & C-A             & 1               & CNN                         & Self                 & nuScenes, KITTI              & \xxmark                                                                    \\
VAD \cite{Jiang2023}                 & 2023          & MVI              & OSM          & A-A/A-S              & BBox            & V, P            & 3               & CNN, T                 & Full                & nuScenes, CARLA, Town05      & \href{https://github.com/hustvl/VAD}{\ccmark}                              \\
Occ4cast \cite{Liu2023}              & 2023          & 3D Voxel Grid    & Implicit     & Implicit             & Occ             & C-A             & 1               & CNN, LSTM                   & Self                 & Lyft, ArgoVerse, ApolloScape & \href{https://github.com/ai4ce/Occ4cast/}{\ccmark}                         \\ \hline
MSRM \cite{Wang2024semi}             & 2024          & BEV              & Implicit     & Implicit             & P-W             & C-A             & 1               & CNN                         & Semi                 & nuScenes                     & \href{https://github.com/kwwcv/SSMP}{\ccmark}                              \\
Fang et al. \cite{Fang2024}          & 2024          & BEV, MVI         & Implicit     & Implicit             & P-W             & C-A             & 1               & CNN                         & Self                 & nuScenes                     & \href{https://github.com/bshfang/self-supervised-motion}{\ccmark}          \\
Wang et al. \cite{Wang2024self}      & 2024          & BEV              & Implicit     & Implicit             & P-W             & C-A             & 1               & CNN                         & Self                 & nuScenes                     & \href{https://github.com/kwwcv/SelfMotion}{\ccmark}                        \\
DeTra \cite{Casas2024}               & 2024          & BEV              & HD Map       & A-A/A-S              & BBox            & V               & 5               & CNN, GNN, T, GRU       & Full                & Argoverse 2, WOD             & \xxmark                                                                    \\ \bottomrule
\end{tabular}%
}
% \todo[inline]{Table not yet cited in the text, but with lots of potentially useful information. After this table (whenever completed) you can even think of creating charts with statistics for each column so we can have an overview of the abundance of the several items, at least for the most relevant or populated in the table. Just a suggestion :-) }
\begin{tablenotes}
\tiny
\item[1] BEV (Bird's-Eye-View), MVI (Multi-View Images), FVI (Front-View Image), RV (Range-View), RI (Residual Image)
\item[2] HD Map (High-Definition Map), OSM (Online Semantic Map)
\item[3] A-A (Agent-Agent), A-S (Agent-Scene)
\item[4] BBox (Bounding Box), P-W (Pixel-Wise), Occ (Occupancy Map)
\item[5] V (Vehicles), P (Pedestrians), B (Bicycles), C-A (Class-Agnostic)
\item[6] PH (Prediction Horizon)
\item[7] CNN (Convolutional Neural Network), GNN (Graph Neural Network), GRU (Gated Recurrent Units), LSTM (Long Short-Term Memory), T (Transformer), VAE (Variational Autoencoder)
\item[8] WOD (Waymo Open Dataset)
\item[9] The symbols include hyperlinks to the respective repositories
\end{tablenotes}
\end{threeparttable}
\end{table*}

\subsection{Qualitative Analysis} \label{subsec:qualitative_analysis}

\autoref{tab:qualitative_analysis} provides a comprehensive overview of the 55 approaches for joint perception and prediction discussed in this survey. The table aims to give readers a detailed one-by-one summary of all the approaches in this field, encapsulating key aspects about the three levels of the proposed taxonomy, alongside miscellaneous information such as the year of publication, datasets or simulators employed, and public code availability. Specifically, each method is listed with the following information: year of publication, input representation, map modeling, interaction modeling, output representation (merged with trajectory modeling due to their close relationship), agents considered, prediction horizon in seconds, base neural network architectures, type of supervision learning applied, datasets and/or simulators used for training and evaluation, and public code availability.
% Finally, we consider to be a base architecture foundation neural networks that are repeatedly used to form more complex models, whether being used solely or in conjunction with other base architectures, such as CNNs, GNNs, GRUs, LSTMs and Transformers. 

Details on input representation, scene context modeling (map modeling, interaction modeling, and trajectory modeling), and output representation, including recent trends and commonly employed strategies, are discussed further in \autoref{sec:jpnp}. The most frequently considered dynamic agents are vehicles, as they are the primary category of interest in autonomous driving, sharing roads, behaviors, and rules with the ego vehicle. Pedestrians and bicycles are also commonly addressed due to their unique behaviors that pose distinct challenges for safe autonomous navigation. Additionally, class-agnostic (C-A) approaches are gaining popularity, driven by the rise of pixel-wise and occupancy-based outputs, along with the growing adoption of self-supervised learning techniques. Class-agnostic methods have the advantage of considering all dynamic objects in the scene, irrespective of their presence in the training data.

In terms of prediction horizons, SA-GNN stands out as the longest, predicting up to 7 seconds into the future using a BEV input representation. Several other methods predict up to 6 seconds ahead, often utilizing multi-view images. Predicting 7 seconds into the future can be qualified as long-term prediction, since human drivers, for example, typically assess the scene and adapt their decisions based on the immediate 0.5 to 2 seconds \cite{Triggs1982}. However, achieving reliable long-term predictions remains a significant research challenge, as uncertainties compound over time, making predictions less accurate. While supervised learning methods have advanced in extending prediction horizons, self-supervised learning approaches, which currently achieve predictions up to 3 seconds into the future, offer a promising research direction for reducing dependence on expensive manually labeled datasets.

Regarding base neural network architectures, CNNs are by far the most widely used, largely due to their established reliability as feature extractors. Their extensive research and versatility make them well-suited for a variety of representations, including images, BEV, and RV. However, transformer-based models have seen substantial growth in recent years, offering versatility in feature extraction, temporal and spatial information fusion, and interaction modeling. This flexibility has led many approaches to adopt transformers, demonstrating excellent results. Additionally, GNNs are frequently employed for interaction modeling.

As for datasets and simulators, nuScenes is by far the most popular. The nuScenes dataset \cite{Caesar2020} offers an excellent balance between size, complexity, diversity, and comprehensive labeled data. It features a full sensor suite (1 LiDAR, 5 RADAR, 6 cameras, IMU, GPS), 1000 20-second scenes across diverse conditions (e.g., heavy or light traffic, sunny, rainy, foggy, snowy, nighttime driving, construction zones, urban environments, among others), detailed HD map information, annotated 3D bounding boxes across 23 object classes, and LiDAR points labeled for 32 classes. These characteristics make it an ideal benchmark for joint perception and prediction research.

Finally, there has been an encouraging increase in recent years in the public release of implementations. These open-source contributions are invaluable as they provide foundational resources to accelerate advancements in joint perception and prediction research.

\subsection{Quantitative Comparison} \label{subsec:quantitative_comparison}

The quantitative comparison is structured according to the output representations, and evaluates both the perception and prediction performance of the approaches.

\subsubsection{Bounding Box} \label{subsubsec:quantitative_comparison_bbox}

\autoref{tab:quantitative_bbox} summarizes the quantitative comparison of approaches that use the bounding box output representation on the nuScenes validation set, focusing on vehicles, the most critical dynamic agents in the scene and the primary focus of many approaches. It is common for the methods to report performance metrics on the validation set, as nuScenes does not publicly provide ground-truth annotations for the testing set. The approaches evaluated in this subsection include IntentNet \cite{Casas2018}, SpAGNN \cite{Casas2020SpAGNN}, PnPNet \cite{Liang2020}, InteractionTransformer \cite{Li2020}, LiRaNet \cite{Shah2020}, LaserFlow \cite{Meyer2021}, MVFuseNet \cite{Laddha2021MVFuseNet}, MultiXNet \cite{Djuric2021}, and RV-FuseNet \cite{Laddha2021}.

\textbf{Perception performance} is evaluated using Average Precision (AP), a metric commonly used to assess object detection models. AP measures the area under the precision-recall curve, providing a single score that reflects both precision (the proportion of correctly identified positive samples) and recall (the proportion of actual positives correctly identified). A detection is considered a true positive if its Intersection over Union (IoU) exceeds 0.7 with a ground-truth label.

\textbf{Prediction performance} is measured using the Final Displacement Error (FDE) at 3 seconds, defined as:
\begin{equation}
\label{eq:FDE}
\text{FDE} = \frac{1}{N} \sum_{i=1}^{N} \lVert \hat{y}_i(T) - y_i(T) \rVert ~,
\end{equation}
where \( N \) is the total number of samples, \( \hat{y}_i(T) \) is the predicted position at the final time step \( T \) for the \( i \)-th sample, \( y_i(T) \) is the ground truth position at the final time step \( T \), and \( \lVert \cdot \rVert \) denotes the Euclidean distance. In other words, FDE is defined as the L2 (euclidean) distance between the predicted and ground-truth trajectories at the end of the prediction horizon. The accuracy of detections significantly impacts the FDE metric because FDE is calculated based on the predicted trajectories of detected objects. Inaccurate detections or missed objects result in trajectories that either misalign with the ground truth or are entirely absent. To ensure consistency in evaluation, this comparison uses a fixed recall of 60\% at a 0.5 IoU threshold. Additionally, multimodal approaches, such as MultiXNet and LiRaNet, predict multiple possible trajectories for a single agent. To ensure a fair comparison with unimodal approaches, the highest-probability mode of multimodal approaches is used, aligning the evaluation on their most confident predictions rather than leveraging multiple outputs to potentially lower error metrics.

% Please add the following required packages to your document preamble:
% \usepackage{multirow}
\begin{table}[!t]
\renewcommand{\arraystretch}{1.25}
\centering
\caption{Quantitative comparison between approaches using the \textbf{bounding box output representation} in nuScenes validation set. Works are sorted in ascending chronological order. Best results are highlighted in bold.}
\label{tab:quantitative_bbox}
\begin{threeparttable}
\begin{tabular}{lcr}
\multicolumn{1}{c}{\multirow{2}{*}{\textbf{Work}}} & \textbf{Perception}         & \textbf{Prediction (\SI{3}{s})}       \\ \cline{2-3} 
\multicolumn{1}{c}{}                               & \textbf{AP (\si{\percent}) $\uparrow$} & \textbf{FDE (\si{cm}) $\downarrow$} \\ \toprule
IntentNet\tnote{1} \cite{Casas2018}                         & 60.3                        & 118.0                          \\ 
SpAGNN \cite{Casas2020SpAGNN}                      & -                           & 145.0                          \\ 
PnPNet \cite{Liang2020}                            & -                           & \textbf{93.0}                  \\ 
InteractionTransformer \cite{Li2020}               & \textbf{70.3}                        & 112.4                          \\ 
LiRaNet \cite{Shah2020}                            & 63.7                        & 102.0                          \\ 
LaserFlow \cite{Meyer2021}                         & 56.1                        & 143.0                          \\ 
MVFuseNet \cite{Laddha2021MVFuseNet}               & 67.8                        & 99.0                           \\ 
MultiXNet \cite{Djuric2021}                        & 60.6                        & 105.0                          \\ 
RV-FuseNet \cite{Laddha2021}                       & 59.9                        & 120.0                          \\ \bottomrule
\end{tabular}
\begin{tablenotes}
\tiny
\item[1] Results retrieved from MVFuseNet \cite{Laddha2021MVFuseNet} and MultiXNet \cite{Djuric2021}
\end{tablenotes}
\end{threeparttable}
\end{table}

\textbf{Perception performance comparison:} The InteractionTransformer approach achieves the highest AP (70.3\%). This can be attributed to its multi-sensor fusion backbone, which integrates BEV LiDAR, front-view camera images, and HD maps. Each of these representations contributes to capturing fine-grained features essential for precise object detection. Furthermore, the interaction transformer module provides a slightly boost in detection AP while significantly improving motion forecasting metrics, demonstrating the benefits of interaction modeling for both perception and prediction tasks. It is worth noting that some approaches, such as SpAGNN and PnPNet, do not report perception metrics, as many joint perception and prediction methods prioritize improving prediction performance through joint optimization of multi-task learning.

\textbf{Prediction Performance Comparison:} PnPNet achieves the lowest FDE (\SI{93.0}{cm}). This superior performance is primarily due to its explicit retrieval of motion features from tracked trajectories at each time step, rather than inferring them from raw sensor data. This design ensures a more accurate and stable motion history, significantly reducing prediction errors.

% Please add the following required packages to your document preamble:
% \usepackage{multirow}
% \usepackage{graphicx}
\begin{table*}[!ht]
\renewcommand{\arraystretch}{1.25}
\centering
\caption{Quantitative comparison between approaches using the \textbf{pixel-wise output representation} in nuScenes validation set. Works are sorted in ascending chronological order. Best results are highlighted in bold.}
\label{tab:quantitative_pixelwise}
% \resizebox{\columnwidth}{!}{%
\begin{tabular}{lccrccrrr}
\multicolumn{1}{c}{\multirow{3}{*}{\textbf{Work}}} & \multicolumn{2}{c}{\textbf{Perception}}                                                      & \multicolumn{6}{c}{\textbf{Prediction (\SI{1}{s}) - L2 distances (\si{cm}) $\downarrow$}}                                                   \\ \cline{2-9} 
\multicolumn{1}{c}{}                               & \multirow{2}{*}{\textbf{MCA (\si{\percent}) $\uparrow$}} & \multirow{2}{*}{\textbf{OA (\si{\percent}) $\uparrow$}} & \multicolumn{2}{c}{\textbf{Static}} & \multicolumn{2}{c}{\textbf{Slow ($\leq \SI{5}{m/s}$})} & \multicolumn{2}{c}{\textbf{Fast ($> \SI{5}{m/s})$}} \\ \cline{4-9} 
\multicolumn{1}{c}{}                               &                                               &                                              & \textbf{Mean}   & \textbf{Median}   & \textbf{Mean}        & \textbf{Median}       & \textbf{Mean}       & \textbf{Median}      \\ \toprule
MotionNet \cite{Wu2020}                            & 70.3                                          & 95.8                                         & \textbf{2.01}   & 0                 & 22.92                & 9.52                  & 94.54               & 61.80                \\
SSPML \cite{Luo2021Pillar}                         & -                                             & -                                            & 16.20            & 1.0                 & 69.72                & 17.58                  & 355.04               & 208.44       \\ 
Khalil et al. \cite{Khalil2021}                    & 74.1                                          & \textbf{96.9}                                & 2.15            & 0                 & 25.10                & 9.61                  & 102.00              & 70.32                \\
LiCaNext \cite{Khalil2021LiCaNext}                 & 74.3                                          & \textbf{96.9}                                & 2.21            & 0                 & 24.25                & 9.61                  & 98.01               & 68.42                \\ 
LiCaNet \cite{Khalil2022LiCaNet}                   & 73.9                                          & \textbf{96.9}                                & 2.24            & 0                 & 25.04                & 9.64                  & 104.32              & 73.04                \\ 
STAN \cite{Wei2022}                                & 71.0                                          & 95.5                                         & 2.14            & 0                 & 24.26                & 9.57                  & 105.04              & 72.47                \\ 
BE-STI \cite{Wang2022}                             & \textbf{74.7}                                 & 93.8                                         & 2.20            & 0                 & \textbf{21.15}       & \textbf{9.29}         & \textbf{75.11}      & \textbf{54.13}                \\ 
LidNet \cite{Khalil2022}                           & 73.5                                          & 96.7                                         & 2.32            & 0                 & 23.19                & 9.50                  & 93.66               & 66.47                \\ 
WeakMotionNet \cite{Li2023}                        & -                                             & 94.4                                         & 2.43            & 0                 & 33.16                & 12.01                 & 164.22              & 103.19               \\ 
ContrastMotion \cite{Jia2023}                      & -                                             & -                                            & 8.29            & 0                 & 45.22                & 9.59                  & 352.66              & 132.33               \\ 
MSRM \cite{Wang2024semi}                           & -                                             & -                                            & 2.18            & 0                 & 27.46                & 9.96                  & 120.30              & 78.80                \\ 
Fang et al. \cite{Fang2024}                        & -                                             & -                                            & 5.14            & 0                 & 42.12                & 10.73                 & 207.66              & 132.26               \\ 
Wang et al. \cite{Wang2024self}                    & -                                             & -                                            & 4.19            & 0                 & 32.13                & 10.61                 & 229.43              & 105.08               \\ \bottomrule
\end{tabular}%
% }
\end{table*}

\subsubsection{Pixel-Wise} \label{subsubsec:quantitative_comparison_pixel-wise}

\autoref{tab:quantitative_pixelwise} provides a quantitative comparison of approaches employing pixel-wise output representations on the nuScenes validation set. The evaluated approaches include: MotionNet \cite{Wu2020}, SSPML \cite{Luo2021Pillar}, Khalil et al. \cite{Khalil2021}, LiCaNext \cite{Khalil2021LiCaNext}, LiCaNet \cite{Khalil2022LiCaNet}, STAN \cite{Wei2022}, BE-STI \cite{Wang2022}, LidNet \cite{Khalil2022}, WeakMotionNet \cite{Li2023}, ContrastMotion \cite{Jia2023}, MSRM \cite{Wang2024semi}, Fang et al. \cite{Fang2024}, and Wang et al. \cite{Wang2024self}. These methods consider five classes for evaluation: background, vehicle, pedestrian, bicycle, and others. The "others" category includes all remaining foreground objects from the nuScenes dataset and is intended to account for potentially unseen objects not included in the training data.

\textbf{Perception performance} is assessed using two metrics: Mean Classification Accuracy (MCA) and Overall Accuracy (OA). MCA computes the average accuracy across all five categories, while OA calculates the average accuracy across all cells in the BEV grid. MCA is defined as follows:

\begin{equation}
\label{eq:MCA}
\text{MCA} = \frac{1}{C} \sum_{c=1}^{C} \frac{\text{TP}_c}{\text{GT}_c} ~,
\end{equation}
where \( \text{TP}_c \) represents the total number of true positives (correct predictions) for class \( c \), \( \text{GT}_c \) represents the total number of ground truths for class \( c \), and \( C \) is the total number of classes. OA is defined as follows:

\begin{equation}
\label{eq:OA}
\text{OA} = \frac{1}{K} \sum_{k=1}^{K} \frac{\text{TP}_k}{N_k} ~,
\end{equation}
where \( K \) is the total number of samples in the dataset, \( \text{TP}_k \) represents the total number of true positives for sample \( k \), and \( N_k \) represents the total number of cells in sample \( k \).

\textbf{Prediction performance} is evaluated by computing the average and median L2 distances between the predicted and ground-truth displacement vectors associated with each cell at 1 second into the future. Cells are divided into three speed groups: static, slow ($\leq \SI{5}{m/s}$), and fast ($> \SI{5}{m/s}$). This quantitative evaluation methodology, was initially proposed by MotionNet, and has been widely adopted in subsequent research.

\textbf{Perception Performance Comparison:} BE-STI achieved the highest MCA (74.7\%), likely due to its semantic decoder, which integrates low-level and high-level features, enhancing the precision of semantic information. Meanwhile, Khalil et al., LiCaNext, and LiCaNet achieved the highest OA (96.9\%) by effectively fusing complementary features from multiple modalities and representations, enabling a holistic understanding of the scene. Other approaches that do not classify cells into specific categories, such as class-agnostic methods, do not report these perception metrics because their outputs are not designed for class-specific evaluation.

\textbf{Prediction Performance Comparison:} For static objects, the pioneering MotionNet achieved the best metric (mean: \SI{2.01}{cm}), attributed to its use of temporal losses, which facilitated learning for static objects, and the Multiple-Gradient Descent Algorithm (MGDA) \cite{Sener2018}, which adaptively balances the trade-off among its three prediction heads. BE-STI excelled in predicting the displacement of slow and fast-moving objects, achieving the best metrics in these categories. This success can be attributed to its two novel encoders: the Temporal-enhanced Spatial Encoder (TeSE) that extracts spatial features for each frame by leveraging temporal information from adjacent frames, compensating for the sparsity of LiDAR point clouds, and the Spatial-enhanced Temporal Encoder (SeTE) that captures motion cues by analyzing spatial variations across non-adjacent frames, ensuring accurate motion predictions even for fast-moving objects. Approaches using alternative forms of supervision, such as SSPML, WeakMotionNet, ContrastMotion, MSRM, Fang et al., and Wang et al., have yet to match the performance of fully supervised methods. However, these methods are consistently closing the gap over time. Their progress is significant, as they reduce or eliminate dependence on large-scale datasets requiring thousands of costly manual annotations, presenting a more scalable and sustainable solution for future research.

\subsubsection{Occupancy Map} \label{subsubsec:quantitative_comparison_occ}

\autoref{tab:quantitative_occ} presents a quantitative comparison of approaches using occupancy map output representations on the nuScenes validation set. The evaluated approaches include FIERY \cite{Hu2021}, BEVerse \cite{Zhang2022}, ST-P3 \cite{Hu2022}, StretchBEV \cite{Akan2022}, TBP-Former \cite{Fang2023}, UniAD \cite{Hu2023}, FusionAD \cite{Ye2023}, and PowerBEV \cite{Li2023PowerBEV}. This evaluation focuses solely on the vehicles class, considering only vehicle-occupied spaces within the BEV grid.

\textbf{Perception performance} is assessed using the traditional IoU metric, modeling the task as BEV segmentation at the present frame \cite{Hu2022}. \textbf{Prediction performance} is evaluated using two metrics: IoU for future semantic segmentation and Video Panoptic Quality (VPQ) for future instance segmentation. The IoU measures segmentation accuracy at present and future frames by evaluating the overlap between predicted and ground truth segmentations across each time frame. It is calculated as:

\begin{equation}
\label{eq:Future_IoU}
\text{IoU}(\hat{y}_{\text{seg}}^t, y_{\text{seg}}^t) = \frac{1}{T} \sum_{t=0}^{T} \frac{\sum_{h,w} \hat{y}_{\text{seg}}^t \cdot y_{\text{seg}}^t}{\sum_{h,w} \hat{y}_{\text{seg}}^t + y_{\text{seg}}^t - \hat{y}_{\text{seg}}^t \cdot y_{\text{seg}}^t} ~,
\end{equation}
where \( \hat{y}_{\text{seg}}^t \) and \( y_{\text{seg}}^t \) are the predicted and ground truth segmentations at time step \( t \), \( T \) is the total number of time steps for which predictions are made, and the summation over \( h, w \) denotes the spatial dimensions of the segmentation map. Perception IoU is essentially the same but limited to the present frame. The VPQ evaluates the combined recognition and segmentation quality of instance predictions over time. It assesses the consistency of instance IDs across time (recognition quality) and the accuracy of instance segmentations (segmentation quality). VPQ is defined as:
\begin{equation}
\label{eq:VPQ}
\text{VPQ} = \sum_{t=0}^{T} \frac{\sum_{(p_t, q_t) \in \text{TP}_t} \text{IoU}(p_t, q_t)}{|\text{TP}_t| + \frac{1}{2} |\text{FP}_t| + \frac{1}{2} |\text{FN}_t|} ~,
\end{equation}
where \( \text{TP}_t \) is the set of true positives at time step \( t \), representing correctly detected ground truth instances, \( \text{FP}_t \) is the set of false positives at time step \( t \), representing predicted instances that do not match any ground truth, \( \text{FN}_t \) is the set of false negatives at time step \( t \), representing ground truth instances that were not detected, \( \text{IoU}(p_t, q_t) \) is the intersection-over-union between a predicted instance \( p_t \) and a ground truth instance \( q_t \), and \( T \) is the prediction horizon. An instance is considered a true positive if the IoU exceeds 0.5 and the instance ID is consistently tracked across the 2-second prediction horizon. The results are reported in two ranges: \SI{30}{m} × \SI{30}{m} (Short) and \SI{100}{m} × \SI{100}{m} (Long) around the ego-vehicle. This quantitative evaluation methodology, proposed by FIERY, has been widely adopted in subsequent research.

% Please add the following required packages to your document preamble:
% \usepackage{multirow}
% \usepackage{graphicx}as due to the different super
\begin{table}[!t]
\renewcommand{\arraystretch}{1.25}
\centering
\caption{Quantitative comparison between approaches using the \textbf{occupancy map output representation} in nuScenes validation set. Works are sorted in ascending chronological order. Best results are highlighted in bold.}
\label{tab:quantitative_occ}
% \resizebox{\columnwidth}{!}{%
\begin{tabular}{lccccc}
\multicolumn{1}{c}{\multirow{3}{*}{\textbf{Work}}} & \textbf{Perception}                         & \multicolumn{4}{c}{\textbf{Prediction (\SI{2}{s})}}                                                    \\ \cline{2-6} 
\multicolumn{1}{c}{}                               & \multirow{2}{*}{\textbf{IoU (\si{\percent}) $\uparrow$}} & \multicolumn{2}{c}{\textbf{IoU (\si{\percent}) $\uparrow$}} & \multicolumn{2}{c}{\textbf{VPQ (\si{\percent}) $\uparrow$}} \\ \cline{3-6} 
\multicolumn{1}{c}{}                               &                                             & \textbf{Short}         & \textbf{Long}         & \textbf{Short}         & \textbf{Long}         \\ \hline
FIERY \cite{Hu2021}                                & 38.2                                        & 59.4                   & 36.7                  & 50.2                   & 29.9                  \\
BEVerse \cite{Zhang2022}                           & -                                           & 61.4                   & 40.9                  & 54.3                   & 36.1                  \\
ST-P3 \cite{Hu2022}                                & 40.1                                        & -                      & 38.9                  & -                      & 32.1                  \\
StretchBEV \cite{Akan2022}                         & -                                           & 58.1                   & \textbf{52.5}         & 53.0                   & 47.5                  \\
TBP-Former \cite{Fang2023}                         & \textbf{46.2}                               & 64.7                   & 41.9                  & 56.7                   & 36.9                  \\
UniAD \cite{Hu2023}                                & -                                           & 63.4                   & 40.2                  & 54.7                   & 33.5                  \\
FusionAD \cite{Ye2023}                             & -                                           & \textbf{71.2}          & 51.5                  & \textbf{65.5}          & \textbf{51.1}         \\
PowerBEV \cite{Li2023PowerBEV}                     & -                                           & 62.5                   & 39.3                  & 55.5                   & 33.8                  \\ \hline
\end{tabular}%
% }
\end{table}

\textbf{Perception Performance Comparison:} TBP-Former achieved the highest IoU (46.2\%) among the approaches that report this metric only at the present frame. This success can be attributed to its pose-synchronized BEV encoder, which integrates geometric priors with learning-based mechanisms to achieve one-step synchronization. By minimizing distortion and ensuring precise alignment of spatial and temporal features, TBP-Former significantly improves segmentation performance.

\textbf{Prediction Performance Comparison:} FusionAD outperformed other methods in IoU at short ranges and achieved the highest VPQ at both short and long ranges. This superior performance can be credited to its multi-modality fusion framework and the Fusion-Aided Modality-Aware Prediction (FMSPnP) module. FusionAD effectively combines complementary modalities (multi-view images and BEV LiDAR), generating robust BEV features that enhance motion prediction accuracy and reliability. Additionally, the FMSPnP module employs a modality self-attention mechanism to capture interactions between different modalities, enabling a comprehensive understanding of the multi-modal data and refining trajectory predictions. StretchBEV achieves state of the art performance in long range prediction, with the highest IoU (52.5\%) and the second highest VPQ (47.5\%). This success is largely due to its innovative approach of assigning a separate random variable to each spatial coordinate in the BEV grid, allowing the model to effectively capture and represent spatial uncertainty. This capability enables StretchBEV to maintain high prediction accuracy in long-range scenarios, where accurately preserving spatial details becomes significantly more challenging.

% TODO: Create table of abbreviations.

\section{Future Research Directions} \label{sec:future_research_directions}

Through an extensive analysis of the state-of-the-art on joint perception and prediction for autonomous driving, several research gaps were identified, despite the substantial advancements achieved so far. Based on this analysis, we outline potential future research directions to address these gaps and advance the field.

\begin{itemize}
    \item \textbf{Underutilization of certain modalities:} RADAR data remains underexplored in joint perception and prediction approaches. Integrating RADAR with other complementary modalities could enhance robustness, especially in poor visibility conditions. Additionally, the ability of RADAR sensors to infer object velocity can improve motion prediction accuracy, making it a valuable modality for future research;
    \item \textbf{Adapting to non-planar ground planes:} Joint perception and prediction approaches typically assume that all agents within a scene operate on the same ground plane. While this assumption is generally valid for local predictions over short distances, it may not hold true in geographically varied environments. For example, cities like San Francisco in the USA, characterized by steep hills and elevation changes, frequently have agents occupying different ground planes within the same timeframe. Such scenarios can substantially degrade the accuracy of motion predictions. Future models could incorporate 3D elevation data or dynamically adapt to varying ground planes using HD maps or LiDAR data to address this limitation;
    \item \textbf{Hybrid HD and online semantic maps:} HD maps provide detailed, static environmental information but are costly and difficult to maintain. In contrast, online semantic maps are adaptive and up-to-date but less detailed and prone to uncertainties. The advantages of these methods offer complementary benefits, even though they are often perceived as competing strategies. A hybrid approach combining HD maps with real-time sensor data could balance these advantages, offering dynamic map augmentation and predicting map changes on-the-fly with superior details;
    \item \textbf{Exploring inter-class interaction modeling:} Inter-class agent-agent interactions, which involve agents of different classes (e.g., vehicles, pedestrians, bicycles), remain underexplored. These interactions are crucial, as different object classes exhibit unique behaviors and interaction rules. Future research could explore GNNs and transformer-based attention mechanisms to explicitly model inter-class agent-agent interactions, building on their success in intra-class agent-agent and agent-scene interaction modeling;
    \item \textbf{Capturing uncertainty in pixel-wise outputs:} Pixel-wise output representations do not currently model the inherent uncertainty or multimodal nature of future trajectories. Expanding this representation to predict multiple motion hypotheses and represent them using probabilistic distributions, such as Gaussians or Laplacians, could significantly enhance its utility;
    \item \textbf{Instance-aware occupancy map methods:} Occupancy maps have gained popularity for their ability to capture the motion of the entire scene simultaneously in a class-agnostic manner, rather than predicting individual agents one at a time. However, object classes demonstrate unique motion behaviors and interaction dynamics that are critical for accurate predictions. Future research could focus on instance-aware occupancy methods that account for these class-specific behaviors while maintaining the holistic view of the scene;
    \item \textbf{Advancing weakly, semi, and self-supervised learning:} Supervised learning approaches currently dominate the field of joint perception and prediction. However, the emergence of weakly-supervised, semi-supervised, and self-supervised methods offers promising pathways to reduce reliance on large-scale, human-annotated datasets, which are resource-intensive and costly to produce. While these novel approaches have yet to fully match the performance of supervised learning, they are rapidly reducing the gap, as highlighted in \autoref{subsec:quantitative_comparison}. Future research should prioritize further advancements in these methods to minimize dependence on annotated data without compromising performance;
    \item \textbf{Unified metrics for joint perception and prediction:} Perception and prediction tasks are inherently complementary and influence each other, particularly in joint perception and prediction approaches where these tasks are optimized together. However, most existing methods continue to evaluate their performance using separate metrics for each task. This disjoint evaluation fails to account for the interdependence of perception and prediction. There is a need for a unified metric that penalizes errors such as trajectories with correct first-frame detections but incorrect forecasts (false forecasts) and trajectories with incorrect first-frame detections (missed forecasts). FutureDet \cite{Peri2022} introduced such unified metric, called Forecasting Average Precision ($AP_f$), designed for bounding box output representations, but its adoption remains limited. However, no unified metrics exist for pixel-wise or occupancy map output representations, highlighting a significant research gap that should be addressed by future research.
\end{itemize}

\section{Conclusion} \label{sec:conclusion}

This survey provides the first comprehensive review of state-of-the-art approaches in joint perception and prediction for autonomous driving. It introduces a detailed taxonomy that encompasses input representations, scene context modeling, and output representations, offering a structured framework to understand the field. Each section of the taxonomy highlights key methods, their contributions, and their limitations, delivering a thorough overview of current advancements.

The qualitative analysis underscores the increasing complexity of models, from well-established CNNs to the adoption of transformer architectures, alongside the growing interest in self-supervised learning as a complement or alternative to traditional supervised methods. The quantitative comparisons demonstrate substantial progress across the three primary output representations: bounding box, pixel-wise, and occupancy map. These advancements reflect the maturity of the field to achieve safer and more reliable autonomous systems.

This survey serves as a resource for researchers, summarizing key achievements while identifying research gaps and outlining possible future research directions. By bridging existing knowledge and future research possibilities, it aims to guide further exploration and innovation in joint perception and prediction for autonomous driving.
%%%%%%%%

% \section*{Acknowledgments}
% This work has been supported by FCT - Foundation for Science and Technology, in the context of Ph.D. scholarship 2023.02251.BD and under unit 00127-IEETA.
% This work has been supported by FCT - Foundation for Science and Technology, in the context of Ph.D. scholarship 2023.02251.BD and by National Funds through the FCT - Foundation for Science and Technology, in the context of the project UIDB/00127/2020.

\bibliographystyle{IEEEtran}
\bibliography{references}

% Generated by IEEEtran.bst, version: 1.14 (2015/08/26)
\begin{thebibliography}{10}
\providecommand{\url}[1]{#1}
\csname url@samestyle\endcsname
\providecommand{\newblock}{\relax}
\providecommand{\bibinfo}[2]{#2}
\providecommand{\BIBentrySTDinterwordspacing}{\spaceskip=0pt\relax}
\providecommand{\BIBentryALTinterwordstretchfactor}{4}
\providecommand{\BIBentryALTinterwordspacing}{\spaceskip=\fontdimen2\font plus
\BIBentryALTinterwordstretchfactor\fontdimen3\font minus \fontdimen4\font\relax}
\providecommand{\BIBforeignlanguage}[2]{{%
\expandafter\ifx\csname l@#1\endcsname\relax
\typeout{** WARNING: IEEEtran.bst: No hyphenation pattern has been}%
\typeout{** loaded for the language `#1'. Using the pattern for}%
\typeout{** the default language instead.}%
\else
\language=\csname l@#1\endcsname
\fi
#2}}
\providecommand{\BIBdecl}{\relax}
\BIBdecl

\bibitem{Yurtsever2020}
\BIBentryALTinterwordspacing
E.~Yurtsever, J.~Lambert, A.~Carballo, and K.~Takeda, ``A survey of autonomous driving: Common practices and emerging technologies,'' \emph{IEEE Access}, vol.~8, pp. 58\,443--58\,469, 6 2020. [Online]. Available: \url{https://ieeexplore.ieee.org/document/9046805/}
\BIBentrySTDinterwordspacing

\bibitem{Guo2020}
\BIBentryALTinterwordspacing
J.~Guo, U.~Kurup, and M.~Shah, ``Is it safe to drive? an overview of factors, metrics, and datasets for driveability assessment in autonomous driving,'' \emph{IEEE Transactions on Intelligent Transportation Systems}, vol.~21, no.~8, pp. 3135--3151, 2020. [Online]. Available: \url{https://ieeexplore.ieee.org/document/8760560}
\BIBentrySTDinterwordspacing

\bibitem{Levinson2011}
\BIBentryALTinterwordspacing
J.~Levinson, J.~Askeland, J.~Becker, J.~Dolson, D.~Held, S.~Kammel, J.~Z. Kolter, D.~Langer, O.~Pink, V.~Pratt, M.~Sokolsky, G.~Stanek, D.~Stavens, A.~Teichman, M.~Werling, and S.~Thrun, ``Towards fully autonomous driving: Systems and algorithms,'' in \emph{2011 IEEE Intelligent Vehicles Symposium (IV)}, 2011, pp. 163--168. [Online]. Available: \url{https://ieeexplore.ieee.org/document/5940562}
\BIBentrySTDinterwordspacing

\bibitem{Chen2023}
\BIBentryALTinterwordspacing
L.~Chen, Y.~Li, C.~Huang, B.~Li, Y.~Xing, D.~Tian, L.~Li, Z.~Hu, X.~Na, Z.~Li, S.~Teng, C.~Lv, J.~Wang, D.~Cao, N.~Zheng, and F.-Y. Wang, ``Milestones in autonomous driving and intelligent vehicles: Survey of surveys,'' \emph{IEEE Transactions on Intelligent Vehicles}, vol.~8, pp. 1046--1056, 2 2023. [Online]. Available: \url{https://ieeexplore.ieee.org/document/9963987/}
\BIBentrySTDinterwordspacing

\bibitem{Wang2023}
\BIBentryALTinterwordspacing
Y.~Wang, J.~Jiang, S.~Li, R.~Li, S.~Xu, J.~Wang, and K.~Li, ``Decision-making driven by driver intelligence and environment reasoning for high-level autonomous vehicles: A survey,'' \emph{IEEE Transactions on Intelligent Transportation Systems}, vol.~24, pp. 10\,362--10\,381, 10 2023. [Online]. Available: \url{https://ieeexplore.ieee.org/document/10133881/}
\BIBentrySTDinterwordspacing

\bibitem{Kiran2022}
\BIBentryALTinterwordspacing
B.~R. Kiran, I.~Sobh, V.~Talpaert, P.~Mannion, A.~A. Sallab, S.~Yogamani, and P.~Perez, ``Deep reinforcement learning for autonomous driving: A survey,'' \emph{IEEE Transactions on Intelligent Transportation Systems}, vol.~23, pp. 4909--4926, 6 2022. [Online]. Available: \url{https://ieeexplore.ieee.org/document/9351818}
\BIBentrySTDinterwordspacing

\bibitem{Coelho2022}
\BIBentryALTinterwordspacing
D.~Coelho and M.~Oliveira, ``A review of end-to-end autonomous driving in urban environments,'' \emph{IEEE Access}, vol.~10, pp. 75\,296--75\,311, 2022. [Online]. Available: \url{https://ieeexplore.ieee.org/document/9832636/}
\BIBentrySTDinterwordspacing

\bibitem{Chen2022RL}
\BIBentryALTinterwordspacing
J.~Chen, S.~E. Li, and M.~Tomizuka, ``Interpretable end-to-end urban autonomous driving with latent deep reinforcement learning,'' \emph{IEEE Transactions on Intelligent Transportation Systems}, vol.~23, no.~6, pp. 5068--5078, 2022. [Online]. Available: \url{https://ieeexplore.ieee.org/document/9346000}
\BIBentrySTDinterwordspacing

\bibitem{Coelho2024RLfOLD}
\BIBentryALTinterwordspacing
D.~Coelho, M.~Oliveira, and V.~Santos, ``Rlfold: Reinforcement learning from online demonstrations in urban autonomous driving,'' in \emph{Proceedings of the AAAI Conference on Artificial Intelligence}, vol.~38, 3 2024, pp. 11\,660--11\,668. [Online]. Available: \url{https://ojs.aaai.org/index.php/AAAI/article/view/29049}
\BIBentrySTDinterwordspacing

\bibitem{Coelho2024RLAD}
\BIBentryALTinterwordspacing
------, ``Rlad: Reinforcement learning from pixels for autonomous driving in urban environments,'' \emph{IEEE Transactions on Automation Science and Engineering}, vol.~21, pp. 7427--7435, 10 2024. [Online]. Available: \url{https://ieeexplore.ieee.org/document/10364974/}
\BIBentrySTDinterwordspacing

\bibitem{Feng2021}
\BIBentryALTinterwordspacing
D.~Feng, C.~Haase-Schutz, L.~Rosenbaum, H.~Hertlein, C.~Glaser, F.~Timm, W.~Wiesbeck, and K.~Dietmayer, ``Deep multi-modal object detection and semantic segmentation for autonomous driving: Datasets, methods, and challenges,'' \emph{IEEE Transactions on Intelligent Transportation Systems}, vol.~22, pp. 1341--1360, 3 2021. [Online]. Available: \url{https://ieeexplore.ieee.org/document/9000872}
\BIBentrySTDinterwordspacing

\bibitem{Ghorai2022}
\BIBentryALTinterwordspacing
P.~Ghorai, A.~Eskandarian, Y.-K. Kim, and G.~Mehr, ``State estimation and motion prediction of vehicles and vulnerable road users for cooperative autonomous driving: A survey,'' \emph{IEEE Transactions on Intelligent Transportation Systems}, vol.~23, pp. 16\,983--17\,002, 10 2022. [Online]. Available: \url{https://ieeexplore.ieee.org/document/9756845/}
\BIBentrySTDinterwordspacing

\bibitem{Karle2022}
\BIBentryALTinterwordspacing
P.~Karle, M.~Geisslinger, J.~Betz, and M.~Lienkamp, ``Scenario understanding and motion prediction for autonomous vehicles—review and comparison,'' \emph{IEEE Transactions on Intelligent Transportation Systems}, vol.~23, pp. 16\,962--16\,982, 10 2022. [Online]. Available: \url{https://ieeexplore.ieee.org/document/9733973/}
\BIBentrySTDinterwordspacing

\bibitem{Gonzalez2016}
\BIBentryALTinterwordspacing
D.~Gonzalez, J.~Perez, V.~Milanes, and F.~Nashashibi, ``A review of motion planning techniques for automated vehicles,'' \emph{IEEE Transactions on Intelligent Transportation Systems}, vol.~17, pp. 1135--1145, 4 2016. [Online]. Available: \url{http://ieeexplore.ieee.org/document/7339478/}
\BIBentrySTDinterwordspacing

\bibitem{Paden2016}
\BIBentryALTinterwordspacing
B.~Paden, M.~Cap, S.~Z. Yong, D.~Yershov, and E.~Frazzoli, ``A survey of motion planning and control techniques for self-driving urban vehicles,'' \emph{IEEE Transactions on Intelligent Vehicles}, vol.~1, pp. 33--55, 3 2016. [Online]. Available: \url{http://ieeexplore.ieee.org/document/7490340/}
\BIBentrySTDinterwordspacing

\bibitem{Qian2022}
\BIBentryALTinterwordspacing
R.~Qian, X.~Lai, and X.~Li, ``3d object detection for autonomous driving: A survey,'' \emph{Pattern Recognition}, vol. 130, p. 108796, 2022. [Online]. Available: \url{https://www.sciencedirect.com/science/article/pii/S0031320322002771}
\BIBentrySTDinterwordspacing

\bibitem{Liang2020}
\BIBentryALTinterwordspacing
M.~Liang, B.~Yang, W.~Zeng, Y.~Chen, R.~Hu, S.~Casas, and R.~Urtasun, ``Pnpnet: End-to-end perception and prediction with tracking in the loop,'' in \emph{2020 IEEE/CVF Conference on Computer Vision and Pattern Recognition (CVPR)}.\hskip 1em plus 0.5em minus 0.4em\relax IEEE, 6 2020, pp. 11\,550--11\,559. [Online]. Available: \url{https://ieeexplore.ieee.org/document/9157054/}
\BIBentrySTDinterwordspacing

\bibitem{Zhang2020}
\BIBentryALTinterwordspacing
Z.~Zhang, J.~Gao, J.~Mao, Y.~Liu, D.~Anguelov, and C.~Li, ``Stinet: Spatio-temporal-interactive network for pedestrian detection and trajectory prediction,'' in \emph{2020 IEEE/CVF Conference on Computer Vision and Pattern Recognition (CVPR)}.\hskip 1em plus 0.5em minus 0.4em\relax IEEE, 6 2020, pp. 11\,343--11\,352. [Online]. Available: \url{https://ieeexplore.ieee.org/document/9156474/}
\BIBentrySTDinterwordspacing

\bibitem{Luo2018}
\BIBentryALTinterwordspacing
W.~Luo, B.~Yang, and R.~Urtasun, ``Fast and furious: Real time end-to-end 3d detection, tracking and motion forecasting with a single convolutional net,'' in \emph{2018 IEEE/CVF Conference on Computer Vision and Pattern Recognition}.\hskip 1em plus 0.5em minus 0.4em\relax IEEE, 6 2018, pp. 3569--3577. [Online]. Available: \url{https://ieeexplore.ieee.org/document/8578474/}
\BIBentrySTDinterwordspacing

\bibitem{Zeng2019}
\BIBentryALTinterwordspacing
W.~Zeng, W.~Luo, S.~Suo, A.~Sadat, B.~Yang, S.~Casas, and R.~Urtasun, ``End-to-end interpretable neural motion planner,'' in \emph{2019 IEEE/CVF Conference on Computer Vision and Pattern Recognition (CVPR)}.\hskip 1em plus 0.5em minus 0.4em\relax IEEE, 6 2019, pp. 8652--8661. [Online]. Available: \url{https://ieeexplore.ieee.org/document/8954347/}
\BIBentrySTDinterwordspacing

\bibitem{Wu2020}
\BIBentryALTinterwordspacing
P.~Wu, S.~Chen, and D.~N. Metaxas, ``Motionnet: Joint perception and motion prediction for autonomous driving based on bird’s eye view maps,'' in \emph{2020 IEEE/CVF Conference on Computer Vision and Pattern Recognition (CVPR)}.\hskip 1em plus 0.5em minus 0.4em\relax IEEE, 6 2020, pp. 11\,382--11\,392. [Online]. Available: \url{https://ieeexplore.ieee.org/document/9157538/}
\BIBentrySTDinterwordspacing

\bibitem{Zhang2022}
\BIBentryALTinterwordspacing
Y.~Zhang, Z.~Zhu, W.~Zheng, J.~Huang, G.~Huang, J.~Zhou, and J.~Lu, ``Beverse: Unified perception and prediction in birds-eye-view for vision-centric autonomous driving,'' \emph{arXiv}, 5 2022. [Online]. Available: \url{http://arxiv.org/abs/2205.09743}
\BIBentrySTDinterwordspacing

\bibitem{Meyer2021}
\BIBentryALTinterwordspacing
G.~P. Meyer, J.~Charland, S.~Pandey, A.~Laddha, S.~Gautam, C.~Vallespi-Gonzalez, and C.~K. Wellington, ``Laserflow: Efficient and probabilistic object detection and motion forecasting,'' \emph{IEEE Robotics and Automation Letters}, vol.~6, pp. 526--533, 4 2021. [Online]. Available: \url{https://ieeexplore.ieee.org/document/9310205/}
\BIBentrySTDinterwordspacing

\bibitem{Tong2023}
\BIBentryALTinterwordspacing
W.~Tong, C.~Sima, T.~Wang, L.~Chen, S.~Wu, H.~Deng, Y.~Gu, L.~Lu, P.~Luo, D.~Lin, and H.~Li, ``Scene as occupancy,'' in \emph{2023 IEEE/CVF International Conference on Computer Vision (ICCV)}, 2023, pp. 8372--8381. [Online]. Available: \url{https://ieeexplore.ieee.org/document/10377239}
\BIBentrySTDinterwordspacing

\bibitem{Tian2023}
\BIBentryALTinterwordspacing
X.~Tian, T.~Jiang, L.~Yun, Y.~Mao, H.~Yang, Y.~Wang, Y.~Wang, and H.~Zhao, ``Occ3d: A large-scale 3d occupancy prediction benchmark for autonomous driving,'' in \emph{Advances in Neural Information Processing Systems}, A.~Oh, T.~Naumann, A.~Globerson, K.~Saenko, M.~Hardt, and S.~Levine, Eds., vol.~36.\hskip 1em plus 0.5em minus 0.4em\relax Curran Associates, Inc., 2023, pp. 64\,318--64\,330. [Online]. Available: \url{https://proceedings.neurips.cc/paper_files/paper/2023/file/cabfaeecaae7d6540ee797a66f0130b0-Paper-Datasets_and_Benchmarks.pdf}
\BIBentrySTDinterwordspacing

\bibitem{Khalil2021}
\BIBentryALTinterwordspacing
Y.~H. Khalil and H.~T. Mouftah, ``End-to-end multi-view fusion for enhanced perception and motion prediction,'' in \emph{2021 IEEE 94th Vehicular Technology Conference (VTC2021-Fall)}.\hskip 1em plus 0.5em minus 0.4em\relax IEEE, 9 2021, pp. 1--6. [Online]. Available: \url{https://ieeexplore.ieee.org/document/9625271/}
\BIBentrySTDinterwordspacing

\bibitem{Casas2018}
\BIBentryALTinterwordspacing
S.~Casas, W.~Luo, and R.~Urtasun, ``Intentnet: Learning to predict intention from raw sensor data,'' in \emph{Proceedings of The 2nd Conference on Robot Learning}, ser. Proceedings of Machine Learning Research, A.~Billard, A.~Dragan, J.~Peters, and J.~Morimoto, Eds., vol.~87.\hskip 1em plus 0.5em minus 0.4em\relax PMLR, 29--31 Oct 2018, pp. 947--956. [Online]. Available: \url{https://proceedings.mlr.press/v87/casas18a.html}
\BIBentrySTDinterwordspacing

\bibitem{Sadat2020}
\BIBentryALTinterwordspacing
A.~Sadat, S.~Casas, M.~Ren, X.~Wu, P.~Dhawan, and R.~Urtasun, \emph{Perceive, Predict, and Plan: Safe Motion Planning Through Interpretable Semantic Representations}.\hskip 1em plus 0.5em minus 0.4em\relax Springer, 2020, vol. 12368 LNCS, pp. 414--430. [Online]. Available: \url{https://link.springer.com/10.1007/978-3-030-58592-1_25}
\BIBentrySTDinterwordspacing

\bibitem{Casas2021}
\BIBentryALTinterwordspacing
S.~Casas, A.~Sadat, and R.~Urtasun, ``Mp3: A unified model to map, perceive, predict and plan,'' in \emph{2021 IEEE/CVF Conference on Computer Vision and Pattern Recognition (CVPR)}.\hskip 1em plus 0.5em minus 0.4em\relax IEEE, 6 2021, pp. 14\,398--14\,407. [Online]. Available: \url{https://ieeexplore.ieee.org/document/9577532/}
\BIBentrySTDinterwordspacing

\bibitem{Djuric2021}
\BIBentryALTinterwordspacing
N.~Djuric, H.~Cui, Z.~Su, S.~Wu, H.~Wang, F.-C. Chou, L.~S. Martin, S.~Feng, R.~Hu, Y.~Xu, A.~Dayan, S.~Zhang, B.~C. Becker, G.~P. Meyer, C.~Vallespi-Gonzalez, and C.~K. Wellington, ``Multixnet: Multiclass multistage multimodal motion prediction,'' in \emph{2021 IEEE Intelligent Vehicles Symposium (IV)}, vol. 2021-July.\hskip 1em plus 0.5em minus 0.4em\relax IEEE, 7 2021, pp. 435--442, add these references to the pre-thesis. [Online]. Available: \url{https://ieeexplore.ieee.org/document/9575718/}
\BIBentrySTDinterwordspacing

\bibitem{Luo2021}
\BIBentryALTinterwordspacing
K.~Luo, S.~Casas, R.~Liao, X.~Yan, Y.~Xiong, W.~Zeng, and R.~Urtasun, ``Safety-oriented pedestrian occupancy forecasting,'' in \emph{2021 IEEE/RSJ International Conference on Intelligent Robots and Systems (IROS)}.\hskip 1em plus 0.5em minus 0.4em\relax IEEE, 9 2021, pp. 1015--1022. [Online]. Available: \url{https://ieeexplore.ieee.org/document/9636691/}
\BIBentrySTDinterwordspacing

\bibitem{Khurana2022}
\BIBentryALTinterwordspacing
T.~Khurana, P.~Hu, A.~Dave, J.~Ziglar, D.~Held, and D.~Ramanan, ``Differentiable raycasting for self-supervised occupancy forecasting,'' in \emph{Computer Vision – ECCV 2022: 17th European Conference, Tel Aviv, Israel, October 23–27, 2022, Proceedings, Part XXXVIII}, 10 2022, pp. 353--369. [Online]. Available: \url{https://link.springer.com/10.1007/978-3-031-19839-7_21}
\BIBentrySTDinterwordspacing

\bibitem{Casas2020SpAGNN}
\BIBentryALTinterwordspacing
S.~Casas, C.~Gulino, R.~Liao, and R.~Urtasun, ``Spagnn: Spatially-aware graph neural networks for relational behavior forecasting from sensor data,'' in \emph{2020 IEEE International Conference on Robotics and Automation (ICRA)}.\hskip 1em plus 0.5em minus 0.4em\relax IEEE, 5 2020, pp. 9491--9497. [Online]. Available: \url{https://ieeexplore.ieee.org/document/9196697/}
\BIBentrySTDinterwordspacing

\bibitem{Casas2020ILVM}
\BIBentryALTinterwordspacing
S.~Casas, C.~Gulino, S.~Suo, K.~Luo, R.~Liao, and R.~Urtasun, ``Implicit latent variable model for scene-consistent motion forecasting,'' in \emph{Computer Vision – ECCV 2020: 16th European Conference, Glasgow, UK, August 23–28, 2020, Proceedings, Part XXIII}, 7 2020, pp. 624--641. [Online]. Available: \url{https://link.springer.com/10.1007/978-3-030-58592-1_37}
\BIBentrySTDinterwordspacing

\bibitem{Casas2020}
\BIBentryALTinterwordspacing
S.~Casas, C.~Gulino, S.~Suo, and R.~Urtasun, ``The importance of prior knowledge in precise multimodal prediction,'' in \emph{2020 IEEE/RSJ International Conference on Intelligent Robots and Systems (IROS)}.\hskip 1em plus 0.5em minus 0.4em\relax IEEE, 10 2020, pp. 2295--2302. [Online]. Available: \url{https://ieeexplore.ieee.org/document/9341199/}
\BIBentrySTDinterwordspacing

\bibitem{Phillips2021}
\BIBentryALTinterwordspacing
J.~Phillips, J.~Martinez, I.~A. Barsan, S.~Casas, A.~Sadat, and R.~Urtasun, ``Deep multi-task learning for joint localization, perception, and prediction,'' in \emph{2021 IEEE/CVF Conference on Computer Vision and Pattern Recognition (CVPR)}.\hskip 1em plus 0.5em minus 0.4em\relax IEEE, 6 2021, pp. 4677--4687. [Online]. Available: \url{https://ieeexplore.ieee.org/document/9577283/}
\BIBentrySTDinterwordspacing

\bibitem{Cui2021}
\BIBentryALTinterwordspacing
A.~Cui, S.~Casas, A.~Sadat, R.~Liao, and R.~Urtasun, ``Lookout: Diverse multi-future prediction and planning for self-driving,'' in \emph{2021 IEEE/CVF International Conference on Computer Vision (ICCV)}.\hskip 1em plus 0.5em minus 0.4em\relax IEEE, 10 2021, pp. 16\,087--16\,096. [Online]. Available: \url{https://ieeexplore.ieee.org/document/9710950/}
\BIBentrySTDinterwordspacing

\bibitem{Ma2018}
\BIBentryALTinterwordspacing
J.~Ma, W.~Shao, H.~Ye, L.~Wang, H.~Wang, Y.~Zheng, and X.~Xue, ``Arbitrary-oriented scene text detection via rotation proposals,'' \emph{IEEE Transactions on Multimedia}, vol.~20, no.~11, pp. 3111--3122, 2018. [Online]. Available: \url{Arbitrary-oriented scene text detection via rotation proposals}
\BIBentrySTDinterwordspacing

\bibitem{Wang2022}
\BIBentryALTinterwordspacing
Y.~Wang, H.~Pan, J.~Zhu, Y.-H. Wu, X.~Zhan, K.~Jiang, and D.~Yang, ``Be-sti: Spatial-temporal integrated network for class-agnostic motion prediction with bidirectional enhancement,'' in \emph{2022 IEEE/CVF Conference on Computer Vision and Pattern Recognition (CVPR)}.\hskip 1em plus 0.5em minus 0.4em\relax IEEE, 6 2022, pp. 17\,072--17\,081. [Online]. Available: \url{https://ieeexplore.ieee.org/document/9879542/}
\BIBentrySTDinterwordspacing

\bibitem{Li2023}
\BIBentryALTinterwordspacing
R.~Li, H.~Shi, Z.~Fu, Z.~Wang, and G.~Lin, ``Weakly supervised class-agnostic motion prediction for autonomous driving,'' in \emph{2023 IEEE/CVF Conference on Computer Vision and Pattern Recognition (CVPR)}.\hskip 1em plus 0.5em minus 0.4em\relax IEEE, 6 2023, pp. 17\,599--17\,608. [Online]. Available: \url{https://ieeexplore.ieee.org/document/10205104/}
\BIBentrySTDinterwordspacing

\bibitem{Wang2024semi}
\BIBentryALTinterwordspacing
K.~Wang, Y.~Wu, Z.~Pan, X.~Li, K.~Xian, Z.~Wang, Z.~Cao, and G.~Lin, ``Semi-supervised class-agnostic motion prediction with pseudo label regeneration and bevmix,'' in \emph{Proceedings of the AAAI Conference on Artificial Intelligence}, vol.~38, 3 2024, pp. 5490--5498. [Online]. Available: \url{https://ojs.aaai.org/index.php/AAAI/article/view/28358}
\BIBentrySTDinterwordspacing

\bibitem{Wang2024self}
\BIBentryALTinterwordspacing
K.~Wang, Y.~WU, J.~CEN, Z.~Pan, X.~Li, Z.~Wang, Z.~Cao, and G.~Lin, ``Self-supervised class-agnostic motion prediction with spatial and temporal consistency regularizations,'' in \emph{2024 IEEE/CVF Conference on Computer Vision and Pattern Recognition (CVPR)}, 2024, pp. 14\,638--14\,647. [Online]. Available: \url{https://ieeexplore.ieee.org/document/10658467}
\BIBentrySTDinterwordspacing

\bibitem{Khalil2022}
\BIBentryALTinterwordspacing
Y.~H. Khalil and H.~T. Mouftah, ``Lidnet: Boosting perception and motion prediction from a sequence of lidar point clouds for autonomous driving,'' in \emph{GLOBECOM 2022 - 2022 IEEE Global Communications Conference}.\hskip 1em plus 0.5em minus 0.4em\relax IEEE, 12 2022, pp. 3533--3538. [Online]. Available: \url{https://ieeexplore.ieee.org/document/10001152/}
\BIBentrySTDinterwordspacing

\bibitem{Zhang2021}
\BIBentryALTinterwordspacing
Y.~Zhang, Y.~Ye, Z.~Xiang, and J.~Gu, \emph{SDP-Net: Scene Flow Based Real-Time Object Detection and Prediction from Sequential 3D Point Clouds}.\hskip 1em plus 0.5em minus 0.4em\relax Springer, 2021, vol. 12622 LNCS, pp. 140--157, add these references to the pre-thesis. [Online]. Available: \url{http://link.springer.com/10.1007/978-3-030-69525-5_9}
\BIBentrySTDinterwordspacing

\bibitem{Ye2021}
\BIBentryALTinterwordspacing
S.~Ye, H.~Yao, W.~Wang, Y.~Fu, and Z.~Pan, ``Sdapnet: End-to-end multi-task simultaneous detection and prediction network,'' in \emph{2021 International Joint Conference on Neural Networks (IJCNN)}.\hskip 1em plus 0.5em minus 0.4em\relax IEEE, 7 2021, pp. 1--8, add these references to the pre-thesis. [Online]. Available: \url{https://ieeexplore.ieee.org/document/9533290/}
\BIBentrySTDinterwordspacing

\bibitem{Agro2023}
\BIBentryALTinterwordspacing
B.~Agro, Q.~Sykora, S.~Casas, and R.~Urtasun, ``Implicit occupancy flow fields for perception and prediction in self-driving,'' in \emph{2023 IEEE/CVF Conference on Computer Vision and Pattern Recognition (CVPR)}.\hskip 1em plus 0.5em minus 0.4em\relax IEEE, 6 2023, pp. 1379--1388. [Online]. Available: \url{https://waabi.ai/research/implicito. https://ieeexplore.ieee.org/document/10204893/}
\BIBentrySTDinterwordspacing

\bibitem{Lin2017}
\BIBentryALTinterwordspacing
T.-Y. Lin, P.~Dollár, R.~Girshick, K.~He, B.~Hariharan, and S.~Belongie, ``Feature pyramid networks for object detection,'' in \emph{2017 IEEE Conference on Computer Vision and Pattern Recognition (CVPR)}, 2017, pp. 936--944. [Online]. Available: \url{https://ieeexplore.ieee.org/document/8099589}
\BIBentrySTDinterwordspacing

\bibitem{Chen2022}
\BIBentryALTinterwordspacing
Z.~Chen, Y.~Wang, X.~Liu, and X.~Wang, ``Fs-gru: Continuous perception and prediction with inter frame feature sharing,'' in \emph{2022 IEEE 25th International Conference on Intelligent Transportation Systems (ITSC)}.\hskip 1em plus 0.5em minus 0.4em\relax IEEE, 10 2022, pp. 517--522, add these references to the pre-thesis. [Online]. Available: \url{https://ieeexplore.ieee.org/document/9922356/}
\BIBentrySTDinterwordspacing

\bibitem{Jia2023}
\BIBentryALTinterwordspacing
X.~Jia, H.~Zhou, X.~Zhu, Y.~Guo, J.~Zhang, and Y.~Ma, ``Contrastmotion: Self-supervised scene motion learning for large-scale lidar point clouds,'' in \emph{Proceedings of the Thirty-Second International Joint Conference on Artificial Intelligence}.\hskip 1em plus 0.5em minus 0.4em\relax International Joint Conferences on Artificial Intelligence Organization, 8 2023, pp. 929--937. [Online]. Available: \url{https://www.ijcai.org/proceedings/2023/103}
\BIBentrySTDinterwordspacing

\bibitem{Lang2019}
\BIBentryALTinterwordspacing
A.~H. Lang, S.~Vora, H.~Caesar, L.~Zhou, J.~Yang, and O.~Beijbom, ``Pointpillars: Fast encoders for object detection from point clouds,'' in \emph{2019 IEEE/CVF Conference on Computer Vision and Pattern Recognition (CVPR)}, 2019, pp. 12\,689--12\,697. [Online]. Available: \url{https://ieeexplore.ieee.org/document/8954311}
\BIBentrySTDinterwordspacing

\bibitem{Peri2022}
\BIBentryALTinterwordspacing
N.~Peri, J.~Luiten, M.~Li, A.~Osep, L.~Leal-Taixe, and D.~Ramanan, ``Forecasting from lidar via future object detection,'' in \emph{2022 IEEE/CVF Conference on Computer Vision and Pattern Recognition (CVPR)}.\hskip 1em plus 0.5em minus 0.4em\relax IEEE, 6 2022, pp. 17\,181--17\,190. [Online]. Available: \url{https://ieeexplore.ieee.org/document/9880029/}
\BIBentrySTDinterwordspacing

\bibitem{Casas2024}
\BIBentryALTinterwordspacing
S.~Casas, B.~Agro, J.~Mao, T.~Gilles, A.~Cui, T.~Li, and R.~Urtasun, ``Detra: A unified model for object detection and trajectory forecasting,'' in \emph{Computer Vision -- ECCV 2024}, A.~Leonardis, E.~Ricci, S.~Roth, O.~Russakovsky, T.~Sattler, and G.~Varol, Eds.\hskip 1em plus 0.5em minus 0.4em\relax Cham: Springer Nature Switzerland, 2025, pp. 326--342. [Online]. Available: \url{https://doi.org/10.1007/978-3-031-73223-2_19}
\BIBentrySTDinterwordspacing

\bibitem{Zhou2018}
\BIBentryALTinterwordspacing
Y.~Zhou and O.~Tuzel, ``Voxelnet: End-to-end learning for point cloud based 3d object detection,'' in \emph{2018 IEEE/CVF Conference on Computer Vision and Pattern Recognition (CVPR)}.\hskip 1em plus 0.5em minus 0.4em\relax Los Alamitos, CA, USA: IEEE Computer Society, Jun 2018, pp. 4490--4499. [Online]. Available: \url{https://doi.ieeecomputersociety.org/10.1109/CVPR.2018.00472}
\BIBentrySTDinterwordspacing

\bibitem{Wei2022}
\BIBentryALTinterwordspacing
Z.~Wei, X.~Qi, Z.~Bai, G.~Wu, S.~Nayak, P.~Hao, M.~Barth, Y.~Liu, and K.~Oguchi, ``Spatiotemporal transformer attention network for 3d voxel level joint segmentation and motion prediction in point cloud,'' in \emph{2022 IEEE Intelligent Vehicles Symposium (IV)}, vol. 2022-June.\hskip 1em plus 0.5em minus 0.4em\relax IEEE, 6 2022, pp. 1381--1386, add these references to the pre-thesis. [Online]. Available: \url{https://ieeexplore.ieee.org/document/9827310/}
\BIBentrySTDinterwordspacing

\bibitem{Hu2021}
\BIBentryALTinterwordspacing
A.~Hu, Z.~Murez, N.~Mohan, S.~Dudas, J.~Hawke, V.~Badrinarayanan, R.~Cipolla, and A.~Kendall, ``Fiery: Future instance prediction in bird’s-eye view from surround monocular cameras,'' in \emph{2021 IEEE/CVF International Conference on Computer Vision (ICCV)}.\hskip 1em plus 0.5em minus 0.4em\relax IEEE, 10 2021, pp. 15\,253--15\,262. [Online]. Available: \url{https://ieeexplore.ieee.org/document/9710288/}
\BIBentrySTDinterwordspacing

\bibitem{Jaderberg2015}
\BIBentryALTinterwordspacing
M.~Jaderberg, K.~Simonyan, A.~Zisserman, and k.~kavukcuoglu, ``Spatial transformer networks,'' in \emph{Advances in Neural Information Processing Systems}, C.~Cortes, N.~Lawrence, D.~Lee, M.~Sugiyama, and R.~Garnett, Eds., vol.~28.\hskip 1em plus 0.5em minus 0.4em\relax Curran Associates, Inc., 2015. [Online]. Available: \url{https://proceedings.neurips.cc/paper_files/paper/2015/file/33ceb07bf4eeb3da587e268d663aba1a-Paper.pdf}
\BIBentrySTDinterwordspacing

\bibitem{Li2023PowerBEV}
\BIBentryALTinterwordspacing
P.~Li, S.~Ding, X.~Chen, N.~Hanselmann, M.~Cordts, and J.~Gall, ``Powerbev: A powerful yet lightweight framework for instance prediction in bird’s-eye view,'' in \emph{Proceedings of the Thirty-Second International Joint Conference on Artificial Intelligence}.\hskip 1em plus 0.5em minus 0.4em\relax International Joint Conferences on Artificial Intelligence Organization, 8 2023, pp. 1080--1088. [Online]. Available: \url{https://www.ijcai.org/proceedings/2023/120}
\BIBentrySTDinterwordspacing

\bibitem{Hu2022ST-P3}
\BIBentryALTinterwordspacing
S.~Hu, L.~Chen, P.~Wu, H.~Li, J.~Yan, and D.~Tao, ``St-p3: End-to-end vision-based autonomous driving via spatial-temporal feature learning,'' in \emph{Computer Vision – ECCV 2022: 17th European Conference, Tel Aviv, Israel, October 23–27, 2022, Proceedings, Part XXXVIII}.\hskip 1em plus 0.5em minus 0.4em\relax Berlin, Heidelberg: Springer-Verlag, 2022, p. 533–549. [Online]. Available: \url{https://doi.org/10.1007/978-3-031-19839-7_31}
\BIBentrySTDinterwordspacing

\bibitem{Akan2022}
\BIBentryALTinterwordspacing
A.~K. Akan and F.~G{\"u}ney, ``Stretchbev: Stretching future instance prediction spatially and temporally,'' in \emph{Computer Vision -- ECCV 2022}, S.~Avidan, G.~Brostow, M.~Ciss{\'e}, G.~M. Farinella, and T.~Hassner, Eds.\hskip 1em plus 0.5em minus 0.4em\relax Cham: Springer Nature Switzerland, 2022, pp. 444--460. [Online]. Available: \url{https://doi.org/10.1007/978-3-031-19839-7_26}
\BIBentrySTDinterwordspacing

\bibitem{Jiang2022}
\BIBentryALTinterwordspacing
B.~Jiang, S.~Chen, X.~Wang, B.~Liao, T.~Cheng, J.~Chen, H.~Zhou, Q.~Zhang, W.~Liu, and C.~Huang, ``Perceive, interact, predict: Learning dynamic and static clues for end-to-end motion prediction,'' \emph{arXiv}, 12 2022. [Online]. Available: \url{http://arxiv.org/abs/2212.02181}
\BIBentrySTDinterwordspacing

\bibitem{Hu2023}
\BIBentryALTinterwordspacing
Y.~Hu, J.~Yang, L.~Chen, K.~Li, C.~Sima, X.~Zhu, S.~Chai, S.~Du, T.~Lin, W.~Wang, L.~Lu, X.~Jia, Q.~Liu, J.~Dai, Y.~Qiao, and H.~Li, ``Planning-oriented autonomous driving,'' in \emph{2023 IEEE/CVF Conference on Computer Vision and Pattern Recognition (CVPR)}.\hskip 1em plus 0.5em minus 0.4em\relax IEEE, 6 2023, pp. 17\,853--17\,862. [Online]. Available: \url{https://ieeexplore.ieee.org/document/10205112/}
\BIBentrySTDinterwordspacing

\bibitem{Fang2023}
\BIBentryALTinterwordspacing
S.~Fang, Z.~Wang, Y.~Zhong, J.~Ge, and S.~Chen, ``Tbp-former: Learning temporal bird's-eye-view pyramid for joint perception and prediction in vision-centric autonomous driving,'' in \emph{2023 IEEE/CVF Conference on Computer Vision and Pattern Recognition (CVPR)}.\hskip 1em plus 0.5em minus 0.4em\relax IEEE, 6 2023, pp. 1368--1378. [Online]. Available: \url{https://ieeexplore.ieee.org/document/10205450/}
\BIBentrySTDinterwordspacing

\bibitem{Gu2023}
\BIBentryALTinterwordspacing
J.~Gu, C.~Hu, T.~Zhang, X.~Chen, Y.~Wang, Y.~Wang, and H.~Zhao, ``Vip3d: End-to-end visual trajectory prediction via 3d agent queries,'' in \emph{2023 IEEE/CVF Conference on Computer Vision and Pattern Recognition (CVPR)}.\hskip 1em plus 0.5em minus 0.4em\relax IEEE, 6 2023, pp. 5496--5506. [Online]. Available: \url{https://ieeexplore.ieee.org/document/10205186/}
\BIBentrySTDinterwordspacing

\bibitem{Jiang2023}
\BIBentryALTinterwordspacing
B.~Jiang, S.~Chen, Q.~Xu, B.~Liao, J.~Chen, H.~Zhou, Q.~Zhang, W.~Liu, C.~Huang, and X.~Wang, ``Vad: Vectorized scene representation for efficient autonomous driving,'' in \emph{2023 IEEE/CVF International Conference on Computer Vision (ICCV)}.\hskip 1em plus 0.5em minus 0.4em\relax IEEE, 10 2023, pp. 8306--8316. [Online]. Available: \url{https://ieeexplore.ieee.org/document/10377337/}
\BIBentrySTDinterwordspacing

\bibitem{Laddha2021}
\BIBentryALTinterwordspacing
A.~Laddha, S.~Gautam, G.~P. Meyer, C.~Vallespi-Gonzalez, and C.~K. Wellington, ``Rv-fusenet: Range view based fusion of time-series lidar data for joint 3d object detection and motion forecasting,'' in \emph{2021 IEEE/RSJ International Conference on Intelligent Robots and Systems (IROS)}.\hskip 1em plus 0.5em minus 0.4em\relax IEEE, 9 2021, pp. 7060--7066, add these references to the pre-thesis. [Online]. Available: \url{https://ieeexplore.ieee.org/document/9636083/}
\BIBentrySTDinterwordspacing

\bibitem{Weng2020}
\BIBentryALTinterwordspacing
X.~Weng, J.~Wang, S.~Levine, K.~Kitani, and N.~Rhinehart, ``Inverting the pose forecasting pipeline with spf2: Sequential pointcloud forecasting for sequential pose forecasting,'' in \emph{Proceedings of the 2020 Conference on Robot Learning}, ser. Proceedings of Machine Learning Research, J.~Kober, F.~Ramos, and C.~Tomlin, Eds., vol. 155.\hskip 1em plus 0.5em minus 0.4em\relax PMLR, 16--18 Nov 2021, pp. 11--20. [Online]. Available: \url{https://proceedings.mlr.press/v155/weng21a.html}
\BIBentrySTDinterwordspacing

\bibitem{Fadadu2022}
\BIBentryALTinterwordspacing
S.~Fadadu, S.~Pandey, D.~Hegde, Y.~Shi, F.-C. Chou, N.~Djuric, and C.~Vallespi-Gonzalez, ``Multi-view fusion of sensor data for improved perception and prediction in autonomous driving,'' in \emph{2022 IEEE/CVF Winter Conference on Applications of Computer Vision (WACV)}.\hskip 1em plus 0.5em minus 0.4em\relax IEEE, 1 2022, pp. 3292--3300. [Online]. Available: \url{https://ieeexplore.ieee.org/document/9706809/}
\BIBentrySTDinterwordspacing

\bibitem{Hu2022}
\BIBentryALTinterwordspacing
S.~Hu, L.~Chen, P.~Wu, H.~Li, J.~Yan, and D.~Tao, ``St-p3: End-to-end vision-based autonomous driving via spatial-temporal feature learning,'' in \emph{Computer Vision – ECCV 2022: 17th European Conference, Tel Aviv, Israel, October 23–27, 2022, Proceedings, Part XXXVIII}, 2022, pp. 533--549. [Online]. Available: \url{https://link.springer.com/10.1007/978-3-031-19839-7_31}
\BIBentrySTDinterwordspacing

\bibitem{Khurana2023}
\BIBentryALTinterwordspacing
T.~Khurana, P.~Hu, D.~Held, and D.~Ramanan, ``Point cloud forecasting as a proxy for 4d occupancy forecasting,'' in \emph{2023 IEEE/CVF Conference on Computer Vision and Pattern Recognition (CVPR)}.\hskip 1em plus 0.5em minus 0.4em\relax IEEE, 6 2023, pp. 1116--1124. [Online]. Available: \url{https://ieeexplore.ieee.org/document/10204018/}
\BIBentrySTDinterwordspacing

\bibitem{Liu2023}
\BIBentryALTinterwordspacing
X.~Liu, M.~Gong, Q.~Fang, H.~Xie, Y.~Li, H.~Zhao, and C.~Feng, ``Lidar-based 4d occupancy completion and forecasting,'' \emph{arXiv}, 10 2023. [Online]. Available: \url{http://arxiv.org/abs/2310.11239}
\BIBentrySTDinterwordspacing

\bibitem{Hendy2020}
\BIBentryALTinterwordspacing
N.~Hendy, C.~Sloan, F.~Tian, P.~Duan, N.~Charchut, Y.~Xie, C.~Wang, and J.~Philbin, ``Fishing net: Future inference of semantic heatmaps in grids,'' \emph{arXiv}, 6 2020. [Online]. Available: \url{http://arxiv.org/abs/2006.09917}
\BIBentrySTDinterwordspacing

\bibitem{Li2020}
\BIBentryALTinterwordspacing
L.~L. Li, B.~Yang, M.~Liang, W.~Zeng, M.~Ren, S.~Segal, and R.~Urtasun, ``End-to-end contextual perception and prediction with interaction transformer,'' in \emph{2020 IEEE/RSJ International Conference on Intelligent Robots and Systems (IROS)}.\hskip 1em plus 0.5em minus 0.4em\relax IEEE, 10 2020, pp. 5784--5791. [Online]. Available: \url{https://ieeexplore.ieee.org/document/9341392/}
\BIBentrySTDinterwordspacing

\bibitem{Shah2020}
\BIBentryALTinterwordspacing
M.~Shah, Z.~Huang, A.~Laddha, M.~Langford, B.~Barber, s.~zhang, C.~Vallespi-Gonzalez, and R.~Urtasun, ``Liranet: End-to-end trajectory prediction using spatio-temporal radar fusion,'' in \emph{Proceedings of the 2020 Conference on Robot Learning}, ser. Proceedings of Machine Learning Research, J.~Kober, F.~Ramos, and C.~Tomlin, Eds., vol. 155.\hskip 1em plus 0.5em minus 0.4em\relax PMLR, 16--18 Nov 2021, pp. 31--48. [Online]. Available: \url{https://proceedings.mlr.press/v155/shah21a.html}
\BIBentrySTDinterwordspacing

\bibitem{Luo2021Pillar}
\BIBentryALTinterwordspacing
C.~Luo, X.~Yang, and A.~Yuille, ``Self-supervised pillar motion learning for autonomous driving,'' in \emph{2021 IEEE/CVF Conference on Computer Vision and Pattern Recognition (CVPR)}.\hskip 1em plus 0.5em minus 0.4em\relax IEEE, 6 2021, pp. 3182--3191. [Online]. Available: \url{https://ieeexplore.ieee.org/document/9578855/}
\BIBentrySTDinterwordspacing

\bibitem{Laddha2021MVFuseNet}
\BIBentryALTinterwordspacing
A.~Laddha, S.~Gautam, S.~Palombo, S.~Pandey, and C.~Vallespi-Gonzalez, ``Mvfusenet: Improving end-to-end object detection and motion forecasting through multi-view fusion of lidar data,'' in \emph{2021 IEEE/CVF Conference on Computer Vision and Pattern Recognition Workshops (CVPRW)}.\hskip 1em plus 0.5em minus 0.4em\relax IEEE, 6 2021, pp. 2859--2868, add these references to the pre-thesis. [Online]. Available: \url{https://ieeexplore.ieee.org/document/9522936/}
\BIBentrySTDinterwordspacing

\bibitem{Khalil2021LiCaNext}
\BIBentryALTinterwordspacing
Y.~H. Khalil and H.~T. Mouftah, ``Licanext: Incorporating sequential range residuals for additional advancement in joint perception and motion prediction,'' \emph{IEEE Access}, vol.~9, pp. 146\,244--146\,255, 2021. [Online]. Available: \url{https://ieeexplore.ieee.org/document/9585703/}
\BIBentrySTDinterwordspacing

\bibitem{Khalil2022LiCaNet}
\BIBentryALTinterwordspacing
------, ``Licanet: Further enhancement of joint perception and motion prediction based on multi-modal fusion,'' \emph{IEEE Open Journal of Intelligent Transportation Systems}, vol.~3, pp. 222--235, 2022. [Online]. Available: \url{https://ieeexplore.ieee.org/document/9738812/}
\BIBentrySTDinterwordspacing

\bibitem{Ye2023}
\BIBentryALTinterwordspacing
T.~Ye, W.~Jing, C.~Hu, S.~Huang, L.~Gao, F.~Li, J.~Wang, K.~Guo, W.~Xiao, W.~Mao, H.~Zheng, K.~Li, J.~Chen, and K.~Yu, ``Fusionad: Multi-modality fusion for prediction and planning tasks of autonomous driving,'' \emph{arXiv}, 8 2023. [Online]. Available: \url{http://arxiv.org/abs/2308.01006}
\BIBentrySTDinterwordspacing

\bibitem{Fang2024}
\BIBentryALTinterwordspacing
S.~Fang, Z.~Liu, M.~Wang, C.~Xu, Y.~Zhong, and S.~Chen, ``Self-supervised bird’s eye view motion prediction with cross-modality signals,'' in \emph{Proceedings of the AAAI Conference on Artificial Intelligence}, vol.~38, 3 2024, pp. 1726--1734. [Online]. Available: \url{https://ojs.aaai.org/index.php/AAAI/article/view/27940}
\BIBentrySTDinterwordspacing

\bibitem{Shi2015}
\BIBentryALTinterwordspacing
X.~SHI, Z.~Chen, H.~Wang, D.-Y. Yeung, W.-k. Wong, and W.-c. WOO, ``Convolutional lstm network: A machine learning approach for precipitation nowcasting,'' in \emph{Advances in Neural Information Processing Systems}, C.~Cortes, N.~Lawrence, D.~Lee, M.~Sugiyama, and R.~Garnett, Eds., vol.~28.\hskip 1em plus 0.5em minus 0.4em\relax Curran Associates, Inc., 2015. [Online]. Available: \url{https://proceedings.neurips.cc/paper_files/paper/2015/file/07563a3fe3bbe7e3ba84431ad9d055af-Paper.pdf}
\BIBentrySTDinterwordspacing

\bibitem{He2016}
\BIBentryALTinterwordspacing
K.~He, X.~Zhang, S.~Ren, and J.~Sun, ``Deep residual learning for image recognition,'' in \emph{2016 IEEE Conference on Computer Vision and Pattern Recognition (CVPR)}, 2016, pp. 770--778. [Online]. Available: \url{https://ieeexplore.ieee.org/document/7780459}
\BIBentrySTDinterwordspacing

\bibitem{Ronneberger2015}
\BIBentryALTinterwordspacing
O.~Ronneberger, P.~Fischer, and T.~Brox, ``U-net: Convolutional networks for biomedical image segmentation,'' in \emph{Medical Image Computing and Computer-Assisted Intervention -- MICCAI 2015}, N.~Navab, J.~Hornegger, W.~M. Wells, and A.~F. Frangi, Eds.\hskip 1em plus 0.5em minus 0.4em\relax Cham: Springer International Publishing, 2015, pp. 234--241. [Online]. Available: \url{https://doi.org/10.1007/978-3-319-24574-4_28}
\BIBentrySTDinterwordspacing

\bibitem{Caesar2020}
\BIBentryALTinterwordspacing
H.~Caesar, V.~Bankiti, A.~H. Lang, S.~Vora, V.~Liong, Q.~Xu, A.~Krishnan, Y.~Pan, G.~Baldan, and O.~Beijbom, ``nuscenes: A multimodal dataset for autonomous driving,'' in \emph{2020 IEEE/CVF Conference on Computer Vision and Pattern Recognition (CVPR)}.\hskip 1em plus 0.5em minus 0.4em\relax Los Alamitos, CA, USA: IEEE Computer Society, jun 2020, pp. 11\,618--11\,628. [Online]. Available: \url{https://doi.ieeecomputersociety.org/10.1109/CVPR42600.2020.01164}
\BIBentrySTDinterwordspacing

\bibitem{Liu2016}
\BIBentryALTinterwordspacing
W.~Liu, D.~Anguelov, D.~Erhan, C.~Szegedy, S.~Reed, C.-Y. Fu, and A.~C. Berg, ``Ssd: Single shot multibox detector,'' in \emph{Computer Vision -- ECCV 2016}, B.~Leibe, J.~Matas, N.~Sebe, and M.~Welling, Eds.\hskip 1em plus 0.5em minus 0.4em\relax Cham: Springer International Publishing, 2016, pp. 21--37. [Online]. Available: \url{https://doi.org/10.1007/978-3-319-46448-0_2}
\BIBentrySTDinterwordspacing

\bibitem{Cui2019}
\BIBentryALTinterwordspacing
H.~Cui, V.~Radosavljevic, F.-C. Chou, T.-H. Lin, T.~Nguyen, T.-K. Huang, J.~Schneider, and N.~Djuric, ``Multimodal trajectory predictions for autonomous driving using deep convolutional networks,'' in \emph{2019 International Conference on Robotics and Automation (ICRA)}.\hskip 1em plus 0.5em minus 0.4em\relax IEEE Press, 2019, p. 2090–2096. [Online]. Available: \url{https://doi.org/10.1109/ICRA.2019.8793868}
\BIBentrySTDinterwordspacing

\bibitem{Yin2021}
\BIBentryALTinterwordspacing
T.~Yin, X.~Zhou, and P.~Krahenbuhl, ``{ Center-based 3D Object Detection and Tracking },'' in \emph{2021 IEEE/CVF Conference on Computer Vision and Pattern Recognition (CVPR)}.\hskip 1em plus 0.5em minus 0.4em\relax Los Alamitos, CA, USA: IEEE Computer Society, Jun. 2021, pp. 11\,779--11\,788. [Online]. Available: \url{https://doi.ieeecomputersociety.org/10.1109/CVPR46437.2021.01161}
\BIBentrySTDinterwordspacing

\bibitem{Kuhn1955}
\BIBentryALTinterwordspacing
H.~W. Kuhn, ``The hungarian method for the assignment problem,'' \emph{Naval Research Logistics Quarterly}, vol.~2, no. 1-2, pp. 83--97, 1955. [Online]. Available: \url{https://onlinelibrary.wiley.com/doi/abs/10.1002/nav.3800020109}
\BIBentrySTDinterwordspacing

\bibitem{Triggs1982}
\BIBentryALTinterwordspacing
T.~J. Triggs and W.~G. Harris, ``Reaction time of drivers to road stimuli,'' Monash University Department of Psychology, Tech. Rep., 1982. [Online]. Available: \url{https://www.monash.edu/muarc/archive/our-publications/other/hfr12}
\BIBentrySTDinterwordspacing

\bibitem{Sener2018}
\BIBentryALTinterwordspacing
O.~Sener and V.~Koltun, ``Multi-task learning as multi-objective optimization,'' in \emph{Advances in Neural Information Processing Systems}, S.~Bengio, H.~Wallach, H.~Larochelle, K.~Grauman, N.~Cesa-Bianchi, and R.~Garnett, Eds., vol.~31.\hskip 1em plus 0.5em minus 0.4em\relax Curran Associates, Inc., 2018. [Online]. Available: \url{https://proceedings.neurips.cc/paper_files/paper/2018/file/432aca3a1e345e339f35a30c8f65edce-Paper.pdf}
\BIBentrySTDinterwordspacing

\end{thebibliography}
%% if using biblatex use this instead: % vsantos
%\printbibliography

%\newpage

\begin{IEEEbiography}[{\includegraphics[width=1in,height=1.25in,clip,keepaspectratio]{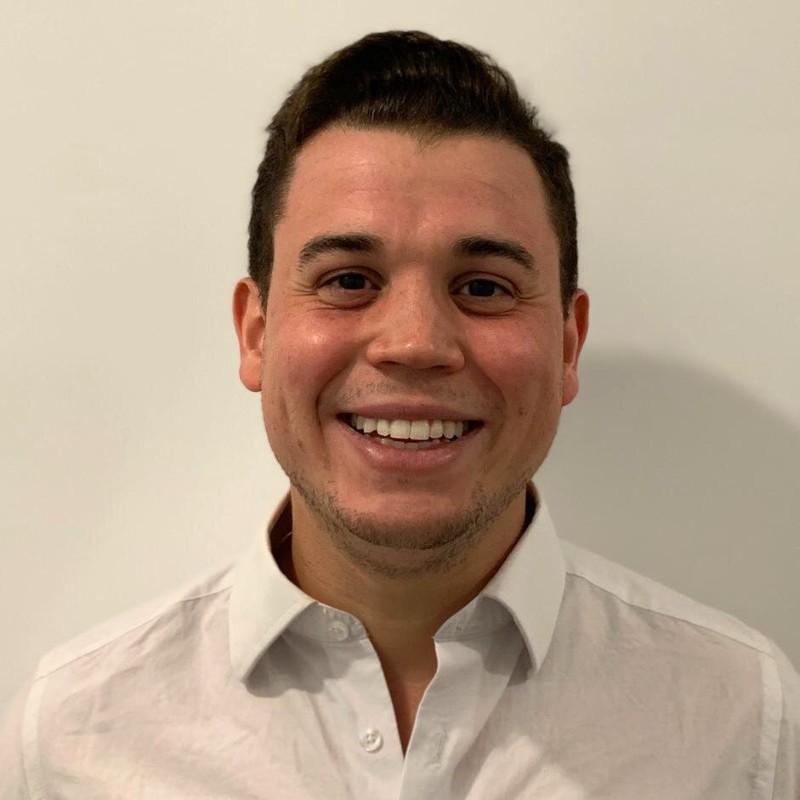}}]{Lucas Dal'Col}
received the bachelor’s and M.Sc. degrees from the University of Aveiro, Portugal, in 2020 and 2022, respectively. Currently, he is pursuing a Ph.D. at the same institution, focusing on deep learning algorithms for perception and prediction in autonomous driving. His research interests include artificial intelligence, deep learning, perception, object detection, object tracking, motion prediction, and autonomous driving.
\end{IEEEbiography}

\begin{IEEEbiography}[{\includegraphics[width=1in,height=1.25in,clip,keepaspectratio]{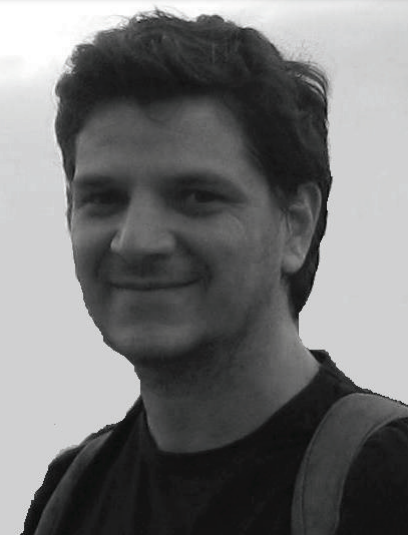}}]{Miguel Oliveira}
received the bachelor’s and M.Sc. degrees from the University of Aveiro, Aveiro, Portugal, in 2004 and 2007, respectively, and the Ph.D. degree in mechanical engineering (specialization in robotics, on the topic of autonomous driving systems) in 2013. From 2013 to 2017, he was an active Researcher with the Institute of Electronics and Telematics Engineering of Aveiro, Aveiro, and the Institute for Systems and Computer Engineering, Technology and Science, Porto, Portugal, where he participated in several EU-funded projects, as well as national projects. He is currently an Assistant Professor with the Department of Mechanical Engineering and the Director of the Master in Automation Engineering, University of Aveiro. He has supervised more than 20 M.Sc. students. He has authored more than 20 journal publications from 2015 to 2020. His research interests include autonomous driving, visual object recognition in open-ended domains, multimodal sensor fusion, computer vision, and the calibration of robotic systems.
\end{IEEEbiography}

\begin{IEEEbiography}[{\includegraphics[width=1in,height=1.25in,clip,keepaspectratio]{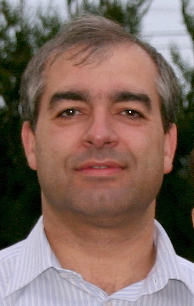}}]{Vítor Santos}
(Member, IEEE) received the degree (five-year) in electronics engineering and telecommunications, the Ph.D. degree in electrical engineering, and the Habilitation degree in mechanical engineering from the University of Aveiro, Portugal, in 1989, 1995, and 2018, respectively. He is currently an Associate Professor with the University of Aveiro, where he lectures several courses on robotics, autonomous vehicles, and computer vision. He has supervised and co-supervised more than 100 students in master’s, Ph.D., and post-doctoral programs, and coordinated the creation of two university degrees in the field of automation at the University of Aveiro. He founded the ATLAS project for mobile robot competition that achieved six first prizes in the annual Autonomous Driving Competition. He has coordinated the development of ATLASCAR, the first real car with autonomous navigation capabilities in Portugal. His research interests include mobile robotics, autonomous driving, advanced perception, and humanoid robotics, as well as public and privately funded projects. He was involved in the organization of several conferences, workshops, and special sessions in national and international events, including being the General Chair of the IEEE International Conference on Autonomous Robot Systems and Competitions, ICARSC2021. He is one of the founders of the Portuguese Robotics Open in 2001 and the Co-Founder of the Portuguese Society of Robotics in 2006.
\end{IEEEbiography}

\vfill

\end{document}